\PassOptionsToPackage{table}{xcolor}
\documentclass{article}
\usepackage{iclr2026_conference,times}

\usepackage[utf8]{inputenc}
\usepackage[T1]{fontenc}
\usepackage[pagebackref,breaklinks,colorlinks]{hyperref}
\usepackage{url}
\usepackage{booktabs}
\usepackage{makecell}
\usepackage{enumitem}
\usepackage{amsmath,amssymb,amsfonts,mathtools}
\usepackage{amsthm}
\usepackage{nicefrac}
\usepackage{microtype}
\usepackage{colortbl}
\usepackage{latexsym}
\usepackage{float}
\usepackage{graphicx}
\usepackage{multirow}
\usepackage{inconsolata}
\usepackage{tikz}
\usetikzlibrary{positioning,calc,arrows.meta,shadows}
\usepackage{caption}
\usepackage{subcaption}
\usepackage{tabularx}
\usepackage[most]{tcolorbox}
\tcbuselibrary{raster,skins,listings,breakable}
\usepackage{adjustbox}
\usepackage{placeins}
\usepackage{pifont}
\usepackage{array}
\usepackage{fontawesome5}
\usepackage{fancyhdr}
\usepackage{wrapfig}
\usepackage[capitalize,noabbrev]{cleveref}
\usepackage{etoc}
\usepackage{titletoc}
\usepackage[leftmargin=1em, rightmargin=1em, indentfirst=false]{quoting}

\theoremstyle{plain}

\theoremstyle{definition}

\theoremstyle{remark}

\usepackage{amsmath,amsfonts,bm}

\def\eqref#1{equation~\ref{#1}}

\def\1{\bm{1}}

\def\vh{{\bm{h}}}

\def\vz{{\bm{z}}}

\DeclareMathAlphabet{\mathsfit}{\encodingdefault}{\sfdefault}{m}{sl}
\SetMathAlphabet{\mathsfit}{bold}{\encodingdefault}{\sfdefault}{bx}{n}

\newcommand{\R}{\mathbb{R}}

\definecolor{valbest}{HTML}{d9ead3}
\newcommand{\valbest}[1]{\colorbox{valbest}{#1}}
\definecolor{valgood}{HTML}{cfe6ec}
\newcommand{\valgood}[1]{\colorbox{valgood}{#1}}
\definecolor{valmid}{HTML}{fce5cd}

\definecolor{valbad}{HTML}{ead1dc}

\definecolor{llamiabench}{HTML}{D8D4F2}

\definecolor{themegreen}{HTML}{365956}
\definecolor{themepurple}{HTML}{3c1b48}
\definecolor{themered}{HTML}{b43748}

\definecolor{StateColor}{HTML}{F94892}
\definecolor{LanguageColor}{HTML}{FF7F3F}
\definecolor{ActionColor}{HTML}{02B9F1}

\newcommand{\state}[1]{\textcolor{StateColor}{\textbf{#1}}}
\newcommand{\lang}[1]{\textcolor{LanguageColor}{\textbf{#1}}}
\newcommand{\action}[1]{\textcolor{ActionColor}{\textbf{#1}}}
\newcommand{\mstate}[1]{{\color{StateColor}#1}}
\newcommand{\mlang}[1]{{\color{LanguageColor}#1}}
\newcommand{\maction}[1]{{\color{ActionColor}#1}}
\newcommand*\circled[1]{\tikz[baseline=(char.base)]{
            \node[shape=circle,draw,inner sep=0.8pt] (char) {#1};}}

\newcommand{\somesh}[1]{}
\newcommand{\yaman}[1]{}
\newcommand{\harini}[1]{}

\providecommand{\Description}[1]{}

\DeclareRobustCommand{\coauth}{$^*$}
\DeclareRobustCommand{\adobelogo}{%
  \raisebox{0.15\height}{\includegraphics[height=2ex]{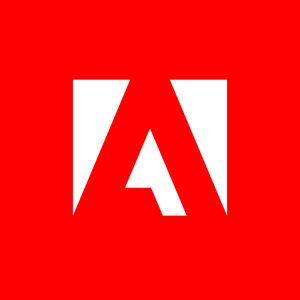}}}
\DeclareRobustCommand{\iiitdlogo}{%
  \raisebox{0.25\height}{\includegraphics[height=1.2ex]{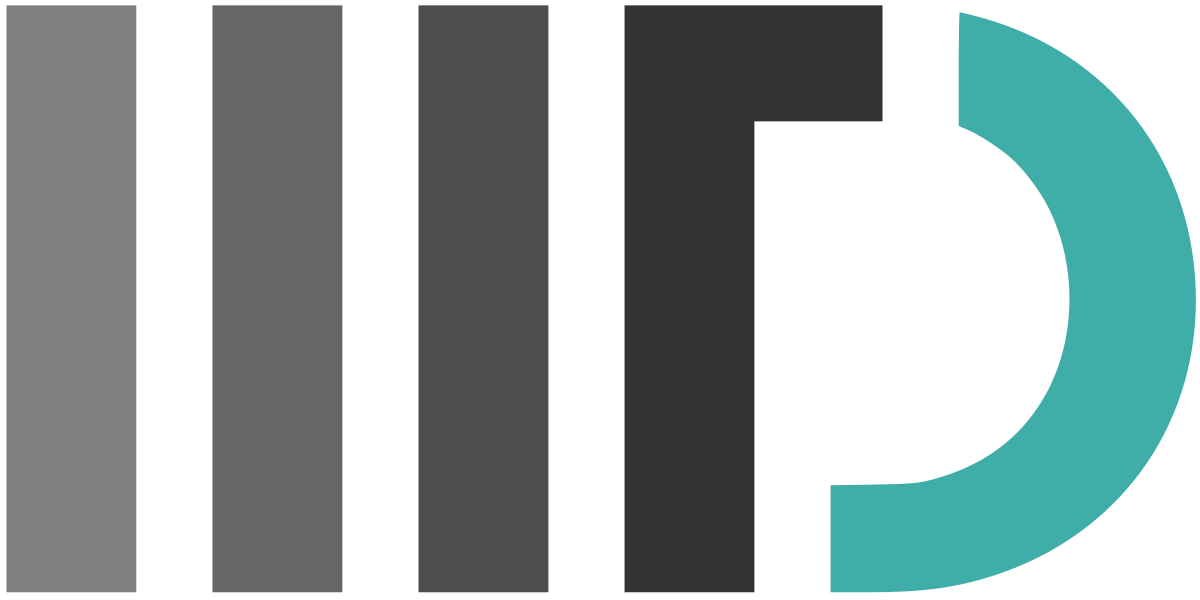}}}
\DeclareRobustCommand{\ublogo}{%
  \raisebox{0.25\height}{\includegraphics[height=1.2ex]{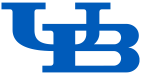}}}
\DeclareRobustCommand{\iitklogo}{%
  \raisebox{0.1\height}{\includegraphics[height=1.8ex]{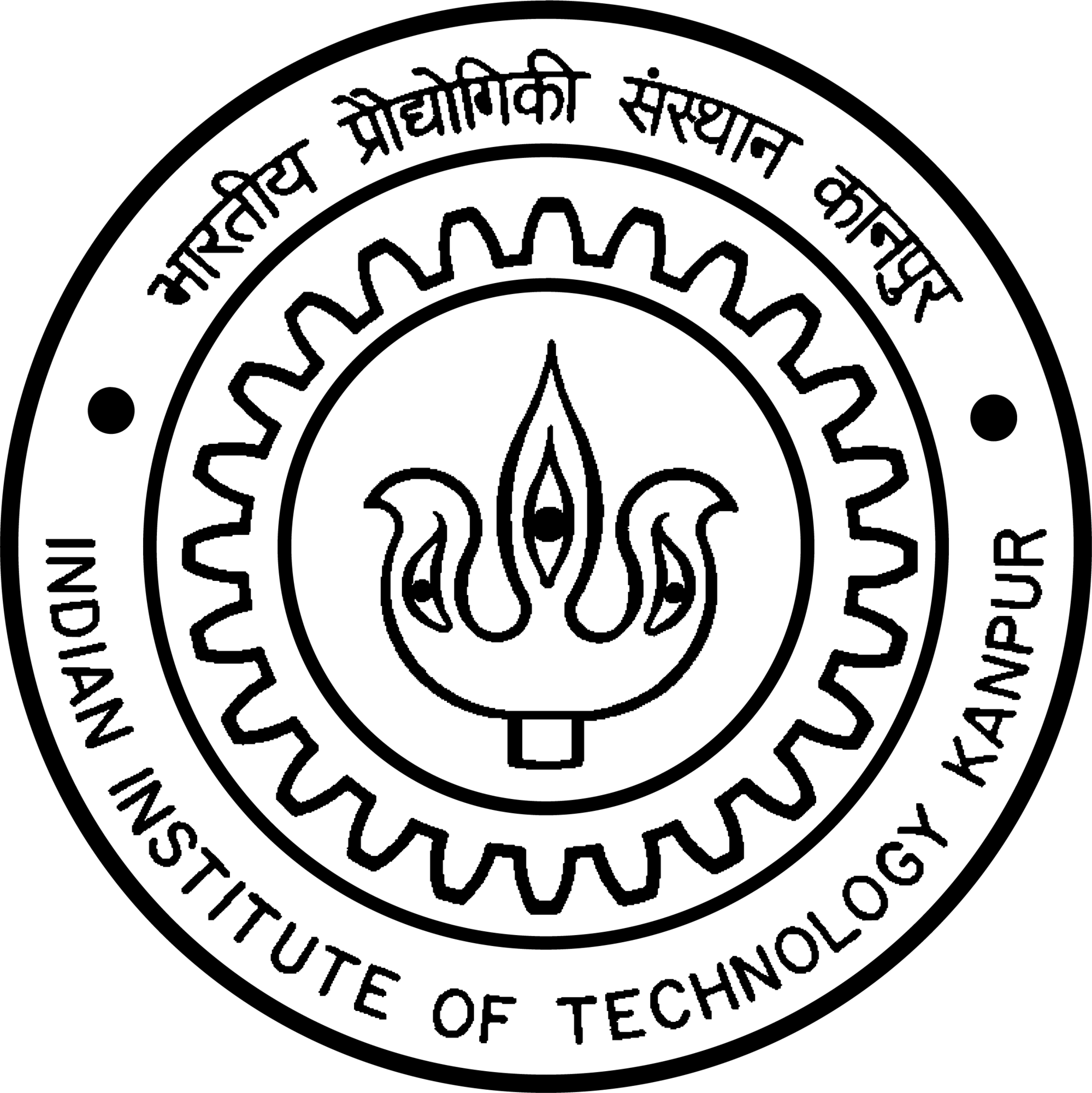}}}
\newcommand{\auth}[2]{\textbf{#1}~#2}

\newcommand\blfootnote[1]{%
  \begingroup
  \renewcommand\thefootnote{}\footnote{#1}%
  \addtocounter{footnote}{-1}%
  \endgroup
}
\newcommand{\cmark}{\textcolor{green!60!black}{$\checkmark$}}
\newcommand{\xmark}{\textcolor{red!70!black}{\texttimes}}

\title{Exploring Collaboration between a language and a non-language agent}

\author{%
\parbox{\textwidth}{\centering\vspace{4mm}
\renewcommand{\arraystretch}{1.35}
\setlength{\tabcolsep}{6pt}
\begin{tabular}{@{}ccc@{}}
  \auth{Harini S I\coauth}{\adobelogo} &
  \auth{Somesh Singh\coauth}{\adobelogo~\iiitdlogo~\ublogo} &
  \auth{Yaman K Singla}{\adobelogo} \\
\end{tabular}\\[1.5mm]
\begin{tabular}{@{}ccc@{}}
  \auth{Rajiv Ratn Shah}{\iiitdlogo~\iitklogo} &
  \auth{David Doermann}{\ublogo} &
  \auth{Balaji Krishnamurthy}{\adobelogo} \\[3mm]
\end{tabular}\\
{\adobelogo~Adobe Media and Data Science Research (MDSR)\\[0.5mm]
 \iiitdlogo~IIIT-Delhi,\quad \iitklogo~IIT Kanpur,\quad \ublogo~SUNY at Buffalo\\[1mm]}
\small\faEnvelope\ \texttt{\href{mailto:behavior-in-the-wild@googlegroups.com}{behavior-in-the-wild@googlegroups.com}}
}%
}

\renewcommand{\headrulewidth}{0.4pt}
\renewcommand{\footrulewidth}{0pt}

\iclrfinalcopy  %

\begin{document}

\maketitle
\renewcommand{\thefootnote}{}\footnotetext{Code, data, and examples can be found at \url{https://behavior-in-the-wild.github.io/llamia.html}}
\renewcommand{\thefootnote}{\arabic{footnote}}

\begin{abstract}
  LLMs are increasingly deployed as orchestrators that coordinate specialized subagents
  to solve complex tasks through natural language. However, in many important domains like game playing and robotics,
  the strongest available agents are not language models. Integrating non-language agents with LLMs would require \emph{verbalization}: compressing their rich continuous representations into sparse textual summaries at each interaction step.
  To study whether verbalization constitutes a bottleneck, we introduce \textsc{LLAMIA-Bench}, a suite of six diverse collaborative chess tasks spanning three facets: behavioral imitation, state assessment, and natural-language explanation. Each task instantiates a well-established chess problem that neither the LLM nor the chess engine can solve alone.
  To solve LLM collaboration with non-language agents, we introduce \emph{latent state internalization}, which projects the subagent's continuous representations directly into the LLM's token stream as learned state tokens, with dynamic re-encoding as actions advance the environment state. Comparing internalization to verbalized integration, our experiments reveal a consistent \emph{verbalization debt}: the performance gap widens throughout training and persists as the LLM scales from 4B to 14B parameters. A single 14B model,
  \textsc{LLAMIA}, trained with latent state internalization, matches or exceeds task specialists and frontier models including GPT-5.1 with tool access across all benchmark tasks, and generalizes out-of-distribution where task-specific finetunes collapse.
\end{abstract}

\begin{NoHyper}
    \blfootnote{\coauth \small Equal Contribution.}
\end{NoHyper}

\section{Introduction}
Large language models (LLMs) are increasingly deployed as general-purpose orchestrators that coordinate tools and agents to solve complex tasks \citep{hong_metagpt_2024, tran_multi-agent_2025}.
A central pattern in this paradigm is collaboration with \emph{subagents}: specialized agents trained to excel within a narrow domain \citep{anthropic_claude_subagents_2025,openai_codex_subagents_2025}.
This collaboration allows orchestrator LLMs to utilize the subagent's domain specific intelligence \citep{hong_metagpt_2024} and preserve its context length by task delegation \citep{zhang_chain_2024}, enabling effective multi-step reasoning and planning over long horizons.
Today, this collaboration is mediated entirely through natural language: the LLM invokes the subagent, receives a natural language description of its output, and reasons over that description to decide subsequent actions\citep{tran_multi-agent_2025}. 
Verbal collaboration is natural when both agents are language models as they share a vocabulary and can express their state in words.
However, in many important domains such as game playing, robotics, and autonomous driving the strongest available agents are not language models; their expertise is encoded in internal representations: latent states capturing policy, value estimates, and learned features like AlphaZero~\citep{silver2017mastering} and RT-1 \citep{brohan2022rt}.
This mismatch between LLMs and non-language agents' input space raises a fundamental question: 
\begin{quoting}
\centering\emph{How can LLMs effectively collaborate and jointly reason with non-language subagents?}
\end{quoting}

The depth of collaboration between LLMs and subagents can solve many useful tasks that neither can solve alone.
Consider chess: LLMs have been trained on more chess literature than most experts will study in a lifetime, yet they cannot leverage it to play the game competently, trailing far behind modern engines and experts \citep{kolasani2025llmchessbenchmarkingreasoning}.
Conversely, pretrained engines surpassed human grandmasters decades ago\citep{campbell2002deep}, yet they remain narrow specialists that are unable to explain the rationale behind a move or strategize under different contexts~\citep{jhamtani-etal-2018-learning, lee2022improving}.
And there exist tasks like game commentary, preparing against an opponent, and designing interesting puzzles which require both the chess engine's deep positional understanding and the LLM's ability to reason over human intent. Effective collaboration between LLMs and the subagent can unlock these applications. 

Existing approaches to LLM-subagent collaboration attempt to bridge this gap symbolically, by \emph{verbalizing} the subagent's outputs into natural language before passing them to the LLM.
Early work finetunes language models on textual descriptions of agent actions and value estimates \citep{zang2019automated, lee2022improving}, while more recent systems rely on in-context learning and tool-calling interfaces to surface subagent outputs at inference time \citep{schick2023toolformerlanguagemodelsteach, yao2022react, kim2025bridginggapexpertlanguage}.
However, these approaches share a common assumption: that the subagent's expertise can be faithfully verbalized.
In this work we demonstrate that this assumption is fundamentally limiting.
A chess engine's latent representation encodes positional structures, long-range tactical motifs, and learned look-ahead~\citep{jenner2024evidence}-- semantic features that cannot be translated to text faithfully.
Verbalization therefore forces the subagent's representations through a lossy bottleneck, collapsing rich latent structure into surface-level descriptions. We call this concept the \textbf{Verbalization Debt}~\cite{zhu2025surveylatentreasoning} and quantify its downstream cost under controlled, heterogeneous LLM--agent collaboration. Moreover, we show that this error compounds across interactions: in multi-step settings, each exchange between the agents strips away details and accumulates errors over the reasoning horizon, diminishing the benefits that motivated this collaboration in the first place.

\begin{figure*}[!h]
    \centering
    \includegraphics[width=0.98\linewidth]{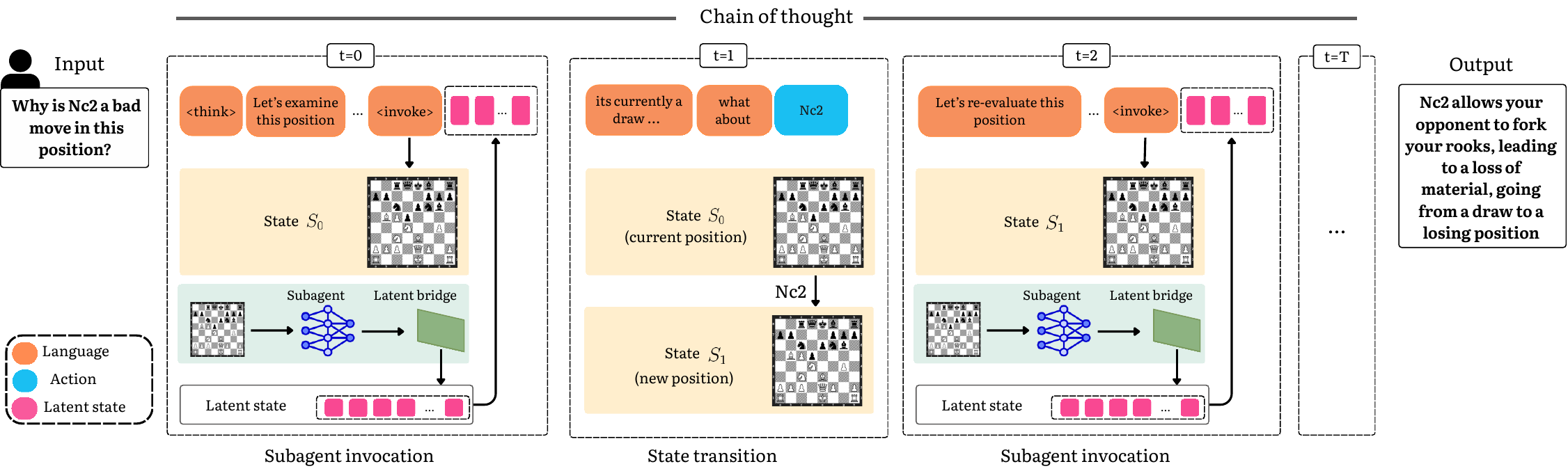}
    \caption{\textbf{Latent State Internalization.}
  A single chain of thought rolled out over time.
  \emph{$t{=}0$ (subagent invocation):} the LLM begins reasoning in
  \mlang{language} and emits \texttt{<invoke>}; the current board $S_0$ is
  passed through the frozen subagent (Lc0-BT4) and the \textbf{LatentBridge}
  maps its hidden activations into $k{=}32$ continuous \mstate{latent state}
  tokens that are appended to the context.
  \emph{$t{=}1$ (state transition):} conditioned on $S_0$ and its latent
  state, the LLM continues reasoning and emits an \maction{action} (Nc2),
  which advances the environment to $S_1$.
  \emph{$t{=}2$ (subagent invocation):} the LLM chooses to re-invoke,
  encoding $S_1$ into a fresh \mstate{latent state} so subsequent reasoning
  is grounded in the updated board. The LLM decides when to re-invoke
  rather than re-encoding on every step. Stage~1 trains only the
  LatentBridge on state--policy pairs; Stage~2 jointly fine-tunes it with
  the LLM via DAPO. The subagent is frozen throughout.}
    \vspace{-3mm}
    \label{fig:LLAMIA}

\end{figure*}

\textbf{Internalization:} To bridge this gap we introduce \emph{latent state internalization}, a paradigm in which an LLM reasons over a non-language agent's internal state over a single trace of three interleaved token types: \lang{language} tokens (the LLM's chain-of-thought), \action{action} tokens (moves that advance the environment), and \state{latent state} tokens (the agent's penultimate-layer activations projected into the LLM's embedding space). As shown in \cref{fig:LLAMIA}, the paradigm consists of three distinct steps: \circled{1} the LLM reasons in language and generates actions that evolve the environment state, such as counterfactual states within its CoT; \circled{2} The LLM on demand requests the agent to evaluate a state, either the current position or a counterfactual reached by a candidate move; \circled{3} the agent encodes the resulting state and its activations are projected into $k{=}32$ latent tokens appended to the reasoning trace, dynamically re-encoded after each state transition. We train a lightweight three-layer MLP, \textbf{LatentBridge} (\cref{sec:architecture}), that learns the projection from the agent's internal state to the LLM's token space.

\textbf{LLAMIA:} We instantiate latent state internalization by training an LLM backbone in two stages: supervised projector alignment followed by reinforcement learning (DAPO), detailed in \cref{sec:training}. We call the resulting model LLAMIA (Large Language and Action Models with Internal Agents). Beyond the technical contribution, internalization opens new avenues for real world applications. Because internalization requires access to model weights, it cannot be applied directly to closed-source models. LLAMIA resolves this tension by acting as a bridge: it interacts with LLMs in natural language and internalizes the subagent, giving closed-weight models indirect but faithful access to the non-language agent's expertise. This makes real-world creative applications like grounded game commentary and opponent-specific preparation deployable, without retraining closed-weight models.

\textbf{Verbalization Debt:} To establish and analyze the effect of internalization when compared to verbalization, we train LLAMIA-Verb, identical to LLAMIA except that the subagent's outputs reach the LLM as text rather than latent tokens. LLAMIA consistently achieves higher reward across all tasks throughout training (\cref{fig:combined}). The gap widens on tasks requiring deeper multi-step integration. These results show that verbalization is a fundamental bottleneck when non-language agents are treated as tools.

\textbf{LLAMIA-Bench, Chess as a testbed:} Despite abundant applications, LLM collaboration with non-language agents remains underexplored. A central reason is the lack of environments that support studying this collaboration at scale: diverse tasks with verifiable metrics and open pretrained agents. Chess is a perfect testbed: decades of human-engine collaboration on commentary, preparation, and puzzles have produced diverse tasks with verifiable metrics, strong open pretrained agents, established evaluation protocols, and large public corpora like Lichess~\cite{feng2024chessgpt}. We therefore use chess as our primary testbed and introduce \textsc{LLAMIA-Bench} (\cref{sec:app_bench}), a curated suite of six tasks spanning behavior cloning across skill levels, puzzle interest and difficulty estimation, move annotation and game-level commentary. We introduce a new dataset curated from \href{https://www.youtube.com/channel/UCL5YbN5WLFD8dLIegT5QAbA}{Agadmator's YouTube Channel} for game level commentary.

\begin{figure}
\centering
\includegraphics[width=0.95\linewidth]{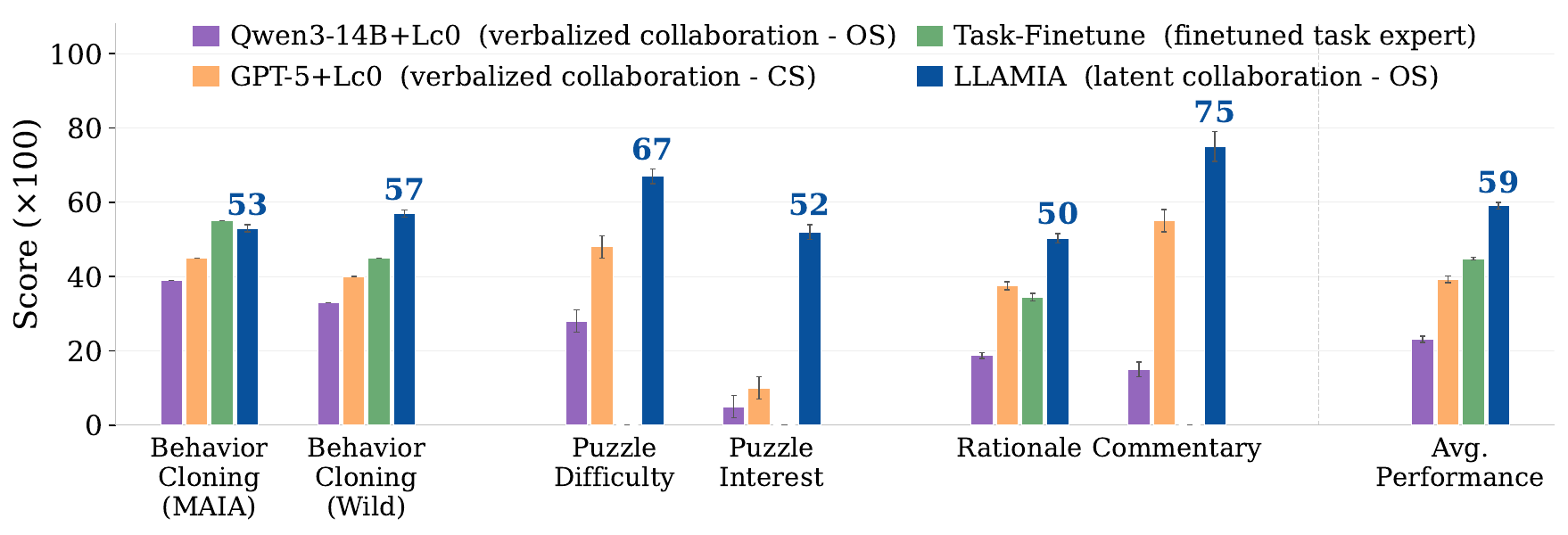}
    \caption{\textbf{LLAMIA-Bench performance across collaboration interfaces.}
    Scores ($\times 100$) on six chess--LLM tasks. Methods are labeled by
    access type and backbone (OS = open-source, CS = closed-source):
    verbalized tool use (Qwen3-14B+Lc0, GPT-5+Lc0), per-task finetuned
    experts (Task-Finetune), and LLAMIA's latent integration. LLAMIA
    matches or exceeds every baseline on every task and is the only system
    to score on Puzzle Interest, where the engine signal has no text
    surrogate. ``Avg.\ Performance'' averages each method over its
    reported tasks.}
    \vspace{-3mm}
    \label{fig:radar_intro}
\end{figure}

\textbf{Results.} A single LLAMIA-14B model matches or exceeds every task-specific specialist and frontier model across all six \textsc{LLAMIA-Bench} tasks (\cref{fig:radar_intro}). The verbalization debt is sharpest on tasks requiring multi-step state tracking or non-verbalizable signals: LLAMIA-Verb’s reward stays flat on commentary and puzzle interest despite identical compute, indicating that verbalization discards task-relevant information needed for effective multi-step collaboration . Internalization also shapes the \emph{kind} of collaboration that emerges: LLAMIA develops counterfactual queries and multi-step lookahead strategies absent from LLAMIA-Verb, which collapses to engine-follow regardless of task, suggesting that access to the full latent state is what makes richer collaboration learnable. Beyond benchmark numbers, LLAMIA reproduces human behavioral signatures: under time pressure it commits the same blunders humans do, and at different skill levels its attention concentrates on the same pieces human players prioritize. A human study confirms this: LLAMIA's gameplay passes as human in the majority of trials, and its commentary is preferred on both strategic insight and explanatory depth compared to the verbalized baseline.

Our contributions are fourfold:
\begin{enumerate}[leftmargin=*, labelsep=0.5em, topsep=2pt, itemsep=2pt]
\item \textbf{Latent state internalization} (Section~\ref{sec:methodology}): A paradigm for LLM--non-language agent collaboration, enabling LLMs to reason over interleaved chain of $\langle
\lang{language}, \action{action}, \state{latent state}\rangle$ tokens.
\item \textbf{LLAMIA} (Section~\ref{sec:methodology}): A two-stage training framework of self-supervised projector alignment followed by end-to-end RL (DAPO) that yields a single model, LLAMIA, achieving state-of-the-art performance across diverse collaborative tasks. LLAMIA enables real world application by giving close-weight models faithful access to subagents.
\item \textbf{Verbalization Debt} (Section~\ref{sec:experiments}): Through controlled ablations against LLAMIA-Verb, we give the first empirical quantification of the \emph{Verbalization Debt} (the performance gap between internalized and verbalized integration) in heterogeneous LLM-agent collaboration, suboptimal in both performance and compute and widening on tasks requiring deeper multi-step collaboration.
\item \textbf{\textsc{LLAMIA-Bench}} (Section~\ref{sec:experiments}): A curated benchmark of six chess tasks spanning behavior cloning, puzzle understanding, commentary, and planning.
\end{enumerate}

\vspace{-2mm}

\newcommand{\latenttokens}{%
\tikz[baseline=-0.6ex]{
    \node[
        fill=StateColor!20, draw=StateColor,
        minimum size=1.5ex,
        inner sep=1ex
    ] {};
}}

\section{Methodology}
\label{sec:methodology}
\subsection{Formulation}
\label{sec:formulation}

We formalize latent state internalization through a running example. An LLM plays chess with access to a pretrained engine exposed through a \textbf{tool API}: functions to read the board state and legal moves, advance the game by making moves, and---\emph{critically}---query the engine's assessment of any position via \texttt{get\_policy} (full schema in \Cref{sec:app_toolcall_tools}). The first two categories handle environment interaction; \texttt{get\_policy} is the interface to the subagent, and \emph{what it returns} is the variable this paper studies.

\smallskip
\noindent``\texttt{\mlang{The knight should develop.} \maction{Nf6}\mlang{.} \texttt{get\_policy($\mstate{s_1}$)} $\to$ (\hspace{0.3em}\latenttokens\hspace{0.3em}\latenttokens\hspace{0.3em}\latenttokens\hspace{0.3em})\hspace{0.3em}\mlang{Nc2: P\!=\!0.34, +0.12;\; e4: P\!=\!0.21, +0.08;\;\ldots}}"
\smallskip

\noindent The LLM reasons in language, plays a knight move to f6 (advancing the game to position $\mstate{s_1}$), then calls \texttt{get\_policy}. The engine returns a text summary---per-move prior probabilities and value estimates---that both model variants receive. In LLAMIA, the call additionally injects $k$ continuous \mstate{state tokens} (\hspace{0.3em}\latenttokens\hspace{0.3em}) projected from the engine's internal activations. In LLAMIA-Verb, only the text is returned. Three token types thus interleave in a trace: \mlang{language tokens} (the LLM's chain-of-thought), \maction{action tokens} (moves that advance the game), and \mstate{state tokens} (the engine's projected latent representation, present only under internalization).

More formally: a policy $\pi_\theta$ (the LLM) interacts with an environment whose states $\mstate{s} \in \mathcal{\mstate{S}}$ evolve through actions $\maction{a} \in \mathcal{\maction{A}}$. A pretrained \textbf{subagent} with encoder $G_\psi\!: \mathcal{\mstate{S}} \to \mstate{\R^d}$ maps each state to a latent representation $\mstate{\vh_s} = G_\psi(\mstate{s})$. The LLM queries the subagent at self-chosen moments via \texttt{get\_policy}, specifying either the current position $\mstate{s_t}$ or a hypothetical state reached by a candidate move. Over a $T$-step interaction the policy produces a trace $\tau$:
\vspace{-2mm}
\begin{equation}
\tau = \bigl(\,
  \mstate{\vz_0^{1:k}},\;
  \mlang{w_{1:j_1}},\;
  \maction{a_1},\;
  \mstate{\vz_1^{1:k}},\;
  \ldots,\;
  \mlang{w_{\mathrm{final}}}
\,\bigr)
\label{eq:trace}
\end{equation}
\vspace{-1mm}
where $\mlang{w_t} \in \mathcal{\mlang{V}}$ are language tokens (including the text returned by \texttt{get\_policy}), $\maction{a_t}$ are actions, and $\mstate{\vz_t^{1:k}}$ are $k$ contiguous state tokens injected alongside the text response. In LLAMIA-Verb, $k{=}0$: no state tokens appear, and the LLM reasons over text alone. In LLAMIA, $k{=}32$. How these state tokens are produced defines the internalization method.

\paragraph{Verbalization.} Every \texttt{get\_policy} call returns a text serialization of the subagent's output:
\begin{equation}
\mstate{\hat{w}_t} = \mathrm{Verb}\bigl(G_\psi(s_t)\bigr) \in \mathcal{V}^*
\label{eq:verb_channel}
\end{equation}
Both LLAMIA and LLAMIA-Verb receive $\hat{w}_t$. A typical return lists the engine's top moves with prior probabilities and value estimates. This captures the subagent's headline assessment but discards the remaining structure in $G_\psi(s_t)$: the full distribution over all legal moves, the value landscape across candidate continuations, and positional features like piece coordination and king safety that interpretability work has identified in engine hidden layers~\cite{jenner2024evidence}.

\paragraph{Internalization.} In LLAMIA, each \texttt{get\_policy} call additionally produces $k$ continuous tokens by projecting the subagent's full latent state into the LLM's embedding space via a learned projection, \textbf{LatentBridge}:
\begin{equation}
\mstate{\vz_t} = H_\varphi\bigl(G_\psi(s_t)\bigr) \in \R^{k \times e}
\label{eq:latent_channel}
\end{equation}
$H_\varphi\!: \R^d \to \R^{k \times e}$ produces $k$ continuous embeddings of dimension $e$ matching the LLM's hidden size. These state tokens sit alongside language and action tokens in the trace (Figure~\ref{fig:LLAMIA}), and the LLM attends over all three types jointly. Where the text serialization imposes a fixed summary regardless of what the current reasoning step requires, latent tokens let the LLM's attention selectively read different aspects of the representation at each step. Gradients flow from the training objective through the LLM back to $H_\varphi$, so the projection adapts to the task.
\subsection{Architecture}
\label{sec:architecture}

\textbf{Subagent.}
We instantiate $G_\psi$ with Lc0-BT4~\cite{monroe2024mastering}, the strongest open-source chess engine, a 15-layer Transformer encoder (240\,M parameters) whose representations encode positional features, piece-value geometry, and lookahead-related structure~\cite{jenner2024evidence}, producing $\vh_s \in \R^{1024}$. We ablate this choice across five Lc0 variants of varying strength in Appendix~\ref{sec:app_abl_agent_size_playing_strength}.

\textbf{Large Language Model.}
We use Qwen3~\cite{yang2025qwen3}, the strongest open-weight LLM at this scale at the time of training, as the backbone $\pi_\theta$. The tool API (\Cref{sec:app_toolcall_tools}) is provided in the system prompt via Hermes-format function calling; Qwen3 natively supports structured tool calls without additional training. To accept \mstate{state tokens}, $k$ contiguous positions in the input sequence serve as placeholders whose embeddings are replaced by the projected state $\vz_s$.

\textbf{LatentBridge.}
The projection $H_\varphi\!: \R^{d} \to \R^{k \times e}$ (Eq.~\ref{eq:latent_channel}) is a three-layer MLP with GeLU activations, motivated by the projector design in vision-language models~\cite{liu2023visual}. It maps Lc0's $1024$-dimensional latent state into $k{=}32$ embeddings of dimension $e$ matching the LLM's hidden size. The resulting input to the LLM is a mixed sequence $[\mlang{w_1}, \mstate{\vz_s^{1:k}}, \maction{a_1}, \mlang{w_2}]$ (Figure~\ref{fig:LLAMIA}). We use the penultimate block (layer 14 of 15) as we observe empirically that its held-out Stage-1 alignment loss is lowest across all blocks (\cref{sec:abl_layer}). Prior interpretability research ~\cite{jenner2024evidence,lin2026bt4features} has shown that this layer in BT4 network locates value, square, and look-ahead-to-action features as well. We set $k{=}32$, where downstream performance saturates in a projector-and-policy sweep giving us an optimal token cost to performance tradeoff \cref{sec:app_abl_projection_token_count}.

\subsection{Training}
\label{sec:training}
Training proceeds in two stages. Stage~1 aligns the subagent's representations with the LLM's embedding space while keeping the LLM frozen. Stage~2 trains the LLM and LatentBridge jointly via reinforcement learning. The subagent is frozen throughout.

\subsubsection{Stage~1: Projector Alignment}
\label{sec:stage1}

We train only $H_\varphi$ while keeping $\pi_\theta$ frozen, on a dataset $\mathcal{D}_{\text{pre}} = \{(s, \pi(s))\}$ of state--policy pairs from the subagent's self-play. Each example pairs the projected state tokens $\vz_s$ with a language prompt (e.g., ``Analyze position: top move?''; format in \cref{sec:app_prompts_stage1}), and the model learns to generate the correct action token via cross-entropy:
\vspace{-2mm}
\[
\mathcal{L}_{\text{Stage\,1}} = -\mathbb{E}_{(s, \pi(s)) \sim \mathcal{D}_{\text{pre}}} \log \pi_\theta\!\left( \pi(s) \mid \vz_s, \text{prompt} \right)
\]
\vspace{-1mm}
Because the LLM is frozen, the projector trains on abundant agent-generated data without risking catastrophic forgetting of language capabilities.
\vspace{-2mm}
\subsubsection{Stage~2: Reinforcement Learning}
\label{sec:stage2}

Stage~2 unfreezes both $\pi_\theta$ and $H_\varphi$ and optimizes them jointly via DAPO~\cite{yu2025dapo}, a group-relative policy optimization variant with asymmetric clipping. Rollouts produce complete traces $\tau$ (Eq.~\ref{eq:trace}). The policy gradient is computed over positions where the LLM generates: \mlang{language tokens} (index set $\mathcal{I}_L$) and \maction{action tokens} (index set $\mathcal{I}_A$), collectively $\mathcal{I}_{\mathrm{gen}} = \mathcal{I}_L \cup \mathcal{I}_A$. \mstate{State token} positions are agent-injected and excluded via gradient masking. Writing $y_t$ for the token at position $t$, the importance-sampling ratio between the current policy $\pi_\theta$ and the reference policy $\pi_{\theta_{\mathrm{old}}}$ from the previous iteration depends on the integration channel:
\begin{align}
r_t^{\,\mathrm{verb}} &=
  \frac{\pi_\theta\bigl(y_t \mid y_{<t},\; \mstate{\hat{w}_{<t}}\bigr)}
       {\pi_{\theta_{\mathrm{old}}}\bigl(y_t \mid y_{<t},\; \mstate{\hat{w}_{<t}}\bigr)},
  \label{eq:ratio_verb} \\[4pt]
r_t^{\,\mathrm{latent}} &=
  \frac{\pi_\theta\bigl(y_t \mid y_{<t},\; \mstate{\vz_{<t}}\bigr)}
       {\pi_{\theta_{\mathrm{old}}}\bigl(y_t \mid y_{<t},\; \mstate{\vz_{<t}}\bigr)}.
  \label{eq:ratio_latent}
\end{align}
\noindent The DAPO objective maximizes:
\vspace{-2mm}
\ifdim\columnwidth<\textwidth
\begin{equation}
\begin{aligned}
J(\theta, \varphi) =\; & \mathbb{E}_{\tau \sim \pi_{\theta_{\mathrm{old}}}}
  \Bigg[ \frac{1}{|\mathcal{I}_{\mathrm{gen}}|}
  \!\!\sum_{t \,\in\, \mlang{\mathcal{I}_{L}} \,\cup\, \maction{\mathcal{I}_{A}}}\!\!\!\!
  \min\!\Big( r_t\,\hat{A}_t, \\
  & \mathrm{clip}(r_t,\, 1{-}\varepsilon_l,\, 1{+}\varepsilon_h)\,\hat{A}_t
  \Big) \Bigg]
\end{aligned}
\label{eq:dapo}
\end{equation}
\else
\begin{equation}
J(\theta, \varphi) = \mathbb{E}_{\tau \sim \pi_{\theta_{\mathrm{old}}}}
  \Bigg[ \frac{1}{|\mathcal{I}_{\mathrm{gen}}|}
  \!\!\sum_{t \,\in\, \mlang{\mathcal{I}_{L}} \,\cup\, \maction{\mathcal{I}_{A}}}\!\!\!\!
  \min\!\Big( r_t\,\hat{A}_t,\;
  \mathrm{clip}(r_t,\, 1{-}\varepsilon_l,\, 1{+}\varepsilon_h)\,\hat{A}_t
  \Big) \Bigg]
\label{eq:dapo}
\end{equation}
\fi
where $r_t$ is the importance ratio (Eq.~\ref{eq:ratio_latent}), $\hat{A}_t = (\mathcal{R}(\tau) - \mu_G)/\sigma_G$ is the group-normalized advantage computed over $G$ rollouts sharing the same prompt~\cite{yu2025dapo}, and the asymmetric clip bounds $\varepsilon_l \leq \varepsilon_h$ encourage exploration. Each \textsc{LLAMIA-Bench} task defines a scalar outcome reward $\mathcal{R}(\tau)$ based on its evaluation metric (Appendix~\ref{sec:app_bench}).

Because $H_\varphi$ is jointly optimized, the RL objective shapes the integration interface end-to-end: the projector learns what representation to present, while the LLM learns how to reason over it. The tool call is part of the policy's action space, so RL also learns \emph{when} and \emph{whether} to query it.

\section{Experiments and Results}
\label{sec:experiments}

We evaluate \textsc{LLAMIA} on \textsc{LLAMIA-Bench}, a suite of six tasks spanning three collaboration facets: behavioral imitation, state assessment, and natural-language explanation (\cref{sec:app_bench}). For comparing LLAMIA with closed-source models where training is not possible, we pair them with Lc0 as a verbalized tool. To evaluate against open-source models we create three settings: an untrained tool-use baseline (Qwen3-14B\,+\,Lc0), a
training-matched verbalization system (LLAMIA-Verb), and our internalized model (LLAMIA), isolating the integration interface as the sole variable.  Within the 
open-source trained systems, we report both SFT and SFT + DAPO checkpoints to disentangle the contributions of supervised pretraining and reinforcement 
learning.

\subsection{LLAMIA-Bench}
\textsc{LLAMIA-Bench} spans six tasks drawn from prominent problems in chess literature and industry, each unsolvable by either component alone: the subagent produces no language, while the LLM lacks the positional signals that make chess-specific judgments possible. The suite covers behavior cloning, puzzle understanding, move annotation, and game-level commentary (details in \cref{sec:app_bench}).
  Behavior cloning, difficulty estimation, and move annotation are drawn from established benchmarks~\cite{mcilroy2020aligning,lichess2024puzzles,jhamtani-etal-2018-learning}. We introduce three new evaluation targets: \textbf{Wild BC}, three OOD
  splits (GM-25, Low-Time, $\Delta$Elo) probing generalization to grandmaster play, time pressure, and asymmetric skill-gap; \textbf{interest estimation}, ranking puzzles by community-derived interestingness scores, a signal with no
   verbal proxy in any engine output; and \textbf{Agadmator-2K}, the first large-scale game-level commentary dataset of 1,900 narrated games with move-aligned
  transcripts.
  Full dataset descriptions, metrics, and per-task prompts appear in Appendices~\ref{sec:app_bench}--\ref{sec:app_prompts}.
\subsection{Setup}
\label{sec:exp_setup}

\paragraph{Baselines}
We compare four systems. \textbf{(1) GPT-5.1\,+\,Lc0}: the strongest frontier model with verbalized Lc0-BT4 tool access. \textbf{(2) Qwen3-14B\,+\,Lc0}: the LLAMIA backbone with the same verbalized tool and 5-shot prompting, without training. \textbf{(3) LLAMIA-Verb}: the matched-recipe ablation with the latent channel replaced by the verbalized tool, isolating the integration interface as the sole variable. \textbf{(4) LLAMIA}: our full system with latent state-token integration. Extended comparisons including LLM-only baselines, SFT and DAPO checkpoints at all three model sizes (4B, 8B, and 14B), frontier model comparisons, and task-specific experts are in \cref{sec:app_results}. Training details and hyperparameters are in Appendix~\ref{sec:app_implementation}.

\paragraph{Metrics}
Each task has a primary metric detailed in \cref{sec:app_bench}: move-matching accuracy (behavior cloning), Spearman $\rho$ (difficulty and interest), and G-eval and BLEU-2 (annotation and commentary). Primary metrics include 95\% bootstrap confidence intervals where sample sizes warrant; per-task breakdowns with CIs appear in \cref{sec:app_results}.

\subsection{Main Results}
\label{sec:results_main}
\begin{figure*}[t]
    \centering
    \includegraphics[width=\linewidth]{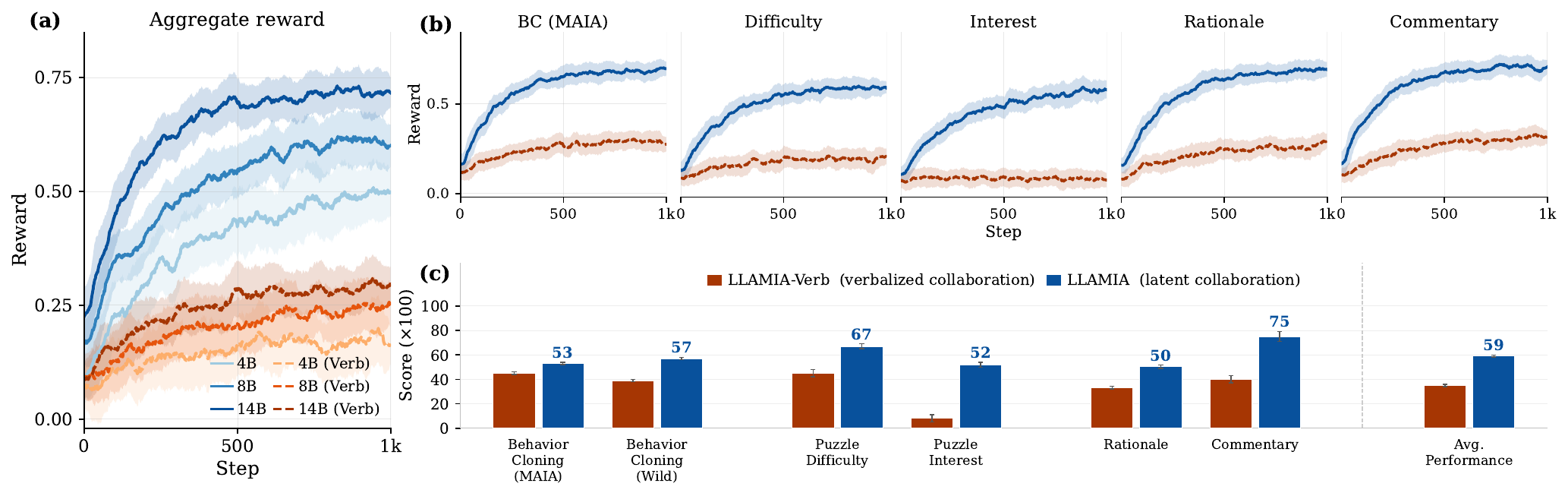}
    \caption{%
    \textbf{DAPO training dynamics and LLAMIA-Bench evaluation.}
    \textbf{(a)}~Aggregate reward vs.\ training step at three backbone scales.
    Solid: LLAMIA (latent); dashed: LLAMIA-Verb (verbalized, identical recipe and backbone).
    The debt widens throughout and reaches $2$--$3\times$ by convergence; scaling the backbone does not close it for LLAMIA-Verb.
    \textbf{(b)}~Per-task reward curves at 14B.
    LLAMIA-Verb gains partial signal on behavior cloning and difficulty (tasks with verbalizable proxies) but stays near-flat on interest and commentary, where the reward requires non-verbalizable features or multi-step integration.
    \textbf{(c)}~LLAMIA-Bench scores ($0$--$100$).
    Puzzle Interest is diagnostic: every verbalized system scores ${\leq}12$ regardless of model scale or frontier capability, while LLAMIA-14B reaches $52$.
    Full per-task tables with confidence intervals in \cref{sec:app_results}.
    }
    \label{fig:combined}
\end{figure*}
LLAMIA-14B achieves the highest score on all six \textsc{LLAMIA-Bench} tasks (\cref{fig:combined}c), surpassing frontier verbalized systems an order of magnitude larger and remaining competitive with dedicated task-specific finetunes that are trained on substantially more in-domain data.
On behavior cloning, Maia and Allie are trained on tens of millions of chess-specific games, against LLAMIA's general-purpose backbone. LLAMIA-14B remains inside the expert band on the in-distribution Maia split and surpasses the strongest expert by a wide margin on the OOD Wild splits (GM-25, Low-Time, $\Delta$Elo), which probe regimes absent from the experts' blitz-only training mixture. Latent access to the engine's policy and value structure thus generalizes more reliably than direct supervision on a narrower distribution. Interest and commentary have no dedicated task-specific baseline at all, no published system predicts puzzle interestingness or generates grounded move commentary from engine state, which is itself evidence that these tasks require the joint reasoning LLAMIA provides rather than a narrower specialist.
The advantage over GPT-5\,+\,Lc0 does not require the 14B backbone: LLAMIA-8B already leads on all six tasks, and LLAMIA-4B on four of six (\cref{tab:full_bench}).
Per-task evaluations with additional metrics, baselines, and confidence intervals are in \cref{sec:app_results}.

\paragraph{Latent tokens enable new evaluation targets.}
Puzzle Interest requires ranking positions by community-derived interestingness, a signal that depends on the engine's policy distribution and value gradients across candidate moves.
No verbalized engine output carries these features.
Every verbalized system scores ${\leq}12$ on Interest regardless of model scale or frontier capability; LLAMIA-14B reaches $52$ (\cref{fig:combined}c).
Verbalization has zero useful signal for this task, while latent tokens give the LLM direct access to the distributional structure that defines interestingness.

\subsection{Verbalization Debt}
\label{sec:results_vd}

LLAMIA and LLAMIA-Verb share the same 14B backbone, Lc0-BT4 subagent, and DAPO recipe; the only difference is whether the subagent's state reaches the LLM as latent tokens or as verbalized text. We define the resulting performance gap as the \textbf{verbalization debt}.

\paragraph{Verbalization debt is significant across all tasks}
Figure \ref{fig:combined} shows that the verbalization debt is consistent across all six tasks. The gap is largest on Interest and Commentary, where the target signal lives in the engine's full policy distribution or value landscape and has no faithful text equivalent, and smallest on in-distribution behavior cloning, where the engine's top-$k$ moves already approximate the answer and the verbal summary loses little.

\paragraph{The debt persists across backbone scale.}
Increasing the LLM from 4B to 14B improves both systems, but the debt persists at every scale (\cref{tab:full_bench}). On Interest, LLAMIA-Verb-14B scores $8$ while LLAMIA-4B already reaches $38$.

\paragraph{The debt widens throughout training.}
The debt grows throughout DAPO, reaching $2$--$3\times$ by the end of training (\cref{fig:combined}a). Per-task reward curves (\cref{fig:combined}b) reveal where the verbal channel saturates: LLAMIA-Verb gains partial signal on behavior cloning and difficulty, where the verbalized output carries a useful proxy (top-$k$ moves, solution length), but stays near-flat on interest and commentary, where no such proxy exists.

\subsection{Ablations}
\label{sec:results_ablations}
To further understand the verbalization debt and isolate the contribution of internalization and reinforcement learning we conduct the following ablations and compare in \cref{tab:ablation_main}. \textbf{LLM-Only} trains with RL but no engine, so any gain has to come from the weights. \textbf{LLM-ChessCLIP} trains with RL and the same $32$ latent slots as LLAMIA, but filled by a raw board encoder (ChessCLIP, a PaLM-E-style injection) rather than Lc0's state, so any gain has to come from capacity rather than content. \textbf{Qwen3+Lc0 (untr.)} is untrained tool use. \textbf{LLAMIA-Verb} is RL on top of verbalized text. \textbf{LLAMIA-SFT (latent)} removes RL, the reasoning trace, and the invocation policy, leaving only the latent channel. \textbf{LLAMIA} is the full system. Extended controls, latent-only, shuffled tokens, per-task probes, and templates, are in \cref{sec:abl_interface,sec:app_baselines_probe}.

\begin{table}[t]
\centering
\caption{\textbf{Interface controls} (14B). BC in \% move-match; Difficulty and Interest in Spearman $\rho$; Rationale in BLEU-2; Commentary in G-eval. All trained systems share backbone, data, and recipe; only the interface differs.}
\label{tab:ablation_main}
\begin{adjustbox}{max width=\columnwidth}
\renewcommand{\arraystretch}{1.15}
\small
\begin{tabular}{lcccccc}
\toprule[1.2pt]
\textbf{System} & \textbf{BC-M} & \textbf{BC-W} & \textbf{Diff.} & \textbf{Int.} & \textbf{Rat.} & \textbf{Comm.} \\
\midrule
LLM-Only (RL, no engine)          & $34$ & $19$ & $0.22$ & $0.07$ & $16.1$ & $0.13$ \\
LLM-ChessCLIP (RL, raw encoder)   & $39$ & $28$ & $0.24$ & $0.08$ & $23.1$ & $0.29$ \\
Qwen3\,+\,Lc0 (untr.\ tool use)   & $39$ & $33$ & $0.28$ & $0.05$ & $18.8$ & $0.15$ \\
LLAMIA-Verb (RL, text only)       & $45$ & $39$ & $0.45$ & $0.08$ & $33.2$ & $0.40$ \\
LLAMIA-SFT (latent, no RL)        & $51$ & $46$ & $0.66$ & $0.48$ & $38.5$ & $0.58$ \\
LLAMIA (latent, RL)               & \valbest{$53$} & \valbest{$49$} & \valbest{$0.71$} & \valbest{$0.52$} & \valbest{$45.8$} & \valbest{$0.75$} \\
\bottomrule[1.2pt]
\end{tabular}
\end{adjustbox}
\vspace{-3mm}
\end{table}

\paragraph{The gain comes from Lc0's latent state, not from weights or capacity.}
LLM-Only and LLM-ChessCLIP get the same RL recipe as LLAMIA and still land near or below untrained tool use, so DAPO cannot manufacture the missing expertise on its own, whether it is asked to bake it into the weights or to make sense of $32$ slots filled with the wrong content. Shuffling LLAMIA's own latent tokens produces the same collapse toward LLAMIA-Verb even though the token count never changes (\cref{sec:abl_interface}). The pattern only breaks when those slots carry Lc0's own policy and value representations. Further, LLAMIA-SFT, with no RL recovers most of the verbalization debt. However, it compounds the effect of internalization by bringing the improvements in multi step strategies, where the model has to plan across latent states i.e. Game Commentary and Rationale generation. We further show this through the emergent latent collaboration strategies in \cref{sec:results_agency}.

\subsection{Collaboration Strategies}
\label{sec:results_agency}
\begin{figure}[t]
    \centering
    \includegraphics[width=\linewidth]{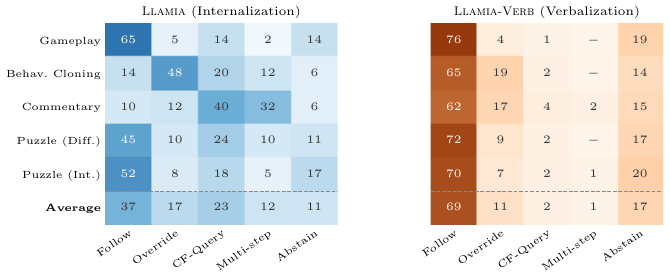}
    \caption{\textbf{Collaboration strategy distribution (\%) per task at convergence.}
  Fraction of 500 episodes assigned to each strategy by a GPT-4o judge
  ($\kappa{=}0.78$). \emph{Left:} LLAMIA (latent). \emph{Right:} LLAMIA-Verb
  (verbalized). LLAMIA's dominant strategy shifts with the task
  (engine-follow for gameplay, consult-then-override for BC, counterfactual
  query for commentary); LLAMIA-Verb collapses to engine-follow on every
  row (62--76\%). Detailed discussion can be found in \cref{sec:abl_agency_strategies}}
    \label{fig:agency_heatmaps}

    \vspace{-2mm}
\end{figure}

Does the integration interface determine how the model learns to use the subagent, or only how well it performs? We classify subagent invocations during evaluation into five recurring strategies and trace their evolution during DAPO training (\cref{fig:agency_task_specificity}; strategy definitions and per-task heatmaps in \cref{fig:agency_heatmaps}). The five strategies are \emph{engine-follow} (adopt the top recommendation), \emph{consult-then-override} (query then diverge), \emph{counterfactual query} (play a hypothetical move, re-invoke, compare states), \emph{multi-step lookahead} (chain two or three counterfactual sequences), and \emph{abstention} (act from language knowledge alone). A GPT-4o judge classifies 500 episodes per task per system ($\kappa{=}0.78$ vs.\ human raters).

\paragraph{Internalization produces task-specific collaboration; verbalization collapses it.}
LLAMIA adapts its strategy to the task: engine-follow dominates gameplay (65\%), consult-then-override dominates behavior cloning (48\%), and counterfactual query dominates commentary (40\%). LLAMIA-Verb collapses to engine-follow on every task (62--76\%), regardless of what the task requires (\cref{fig:agency_heatmaps}). The verbalized channel returns the same compressed summary no matter how the model queries it, so RL converges on a single use pattern.

\paragraph{The learned strategy makes internalization inference-cost neutral.}
Because it reasons over the full latent state, LLAMIA learns to invoke the subagent less often than LLAMIA-Verb ($1.9$ vs.\ $2.9$ calls per query at 14B). Each latent invocation adds a fixed $32$ tokens ($182$ vs.\ $150$), but the lower call count offsets this, so average tokens-per-query and wall-clock latency are comparable or lower than the verbalized interface; training cost stays within ${\sim}6\%$ of the verbalized pipeline at every scale (\cref{sec:app_cost}). Latent internalization therefore does not trade accuracy for inference cost.

\vspace{-2mm}
\subsection{Human Evaluation}
\label{sec:results_human}

\begin{figure*}[t]
    \centering
    \begin{subfigure}[t]{0.235\linewidth}
        \centering
        \includegraphics[width=\linewidth]{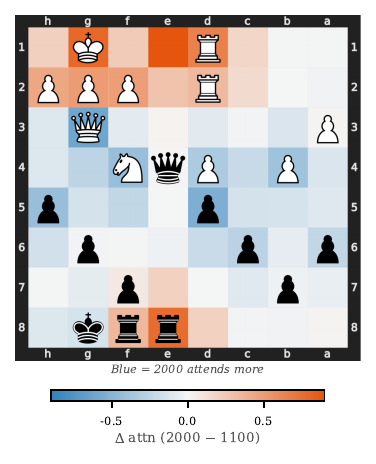}
        \label{fig:human_heatmap}
    \end{subfigure}\hfill
    \begin{subfigure}[t]{0.245\linewidth}
        \centering
        \includegraphics[width=\linewidth]{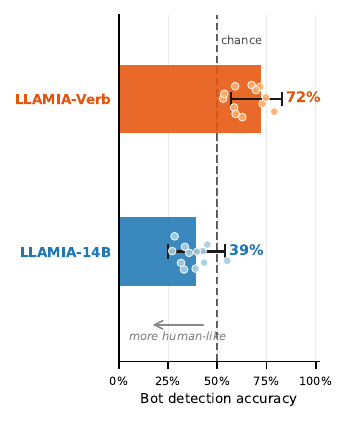}
        \label{fig:human_gameplay_main}
    \end{subfigure}\hfill
    \begin{subfigure}[t]{0.245\linewidth}
        \centering
        \includegraphics[width=\linewidth]{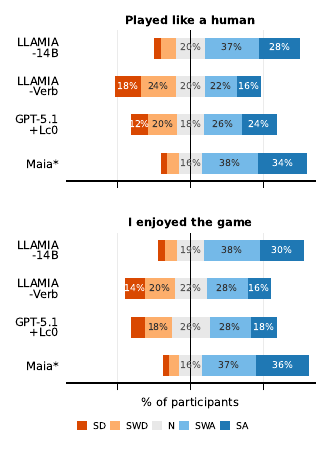}
        \label{fig:human_survey_main}
    \end{subfigure}\hfill
    \begin{subfigure}[t]{0.245\linewidth}
        \centering
        \includegraphics[width=\linewidth]{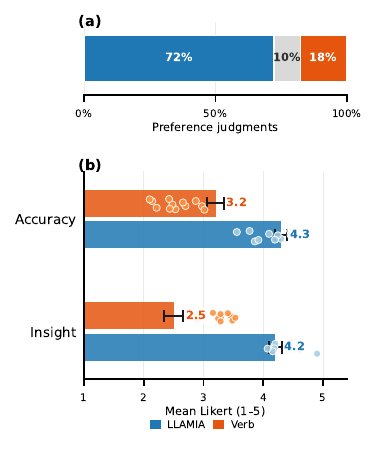}
        \label{fig:human_commentary_main}
    \end{subfigure}
    \vspace{-4mm}
    \caption{%
    \textbf{Human evaluation ($n{=}12$ skilled players, all rated ${\geq}1700$).}
    \textbf{(a)}~Attention difference over latent tokens on a back-rank mate position: conditioning on 2000~Elo concentrates attention on mating geometry (red); 1100~Elo disperses to material (blue). The same latent state is read differently depending on the language instruction.
    \textbf{(b)}~Bot-detection rate (Study~1): LLAMIA-14B passes as human in 61\% of trials (detection 39\%, below chance); LLAMIA-Verb detected in 72\%.
    \textbf{(c)}~Post-game Likert (Strongly Disagree $\rightarrow$ Strongly Agree): LLAMIA matches Maia* (best Maia variant per Elo bucket) on human-likeness without training on human-move distributions.
    \textbf{(d)}~Commentary preference (Study~2): 72.2\% of 180 judgments favour LLAMIA; Insight gap (1.7\,pts) exceeds Accuracy gap (1.1\,pts).
    Study design and qualitative results in \cref{sec:app_human}.
    }
    \vspace{-3mm}
    \label{fig:human_eval}
\end{figure*}

\paragraph{Gameplay.}
Skilled players ($n{=}12$, all ${\geq}1700$ Elo) cannot reliably distinguish LLAMIA from a human opponent: detection falls below chance (\cref{fig:human_gameplay_main}), and post-game ratings place LLAMIA alongside Maia* on perceived human-likeness (\cref{fig:human_survey_main}). Maia is trained directly on millions of move distributions; LLAMIA receives no human-move supervision. Instead, behavioral signatures such as time-pressure blunders and skill-appropriate piece saliency emerge from latent-state conditioning alone. LLAMIA-Verb, trained with the same backbone and DAPO budget, is detected at rates well above chance, consistent with the stylistic regularities that verbal summaries impose on move selection.

\paragraph{Commentary.}
Both systems achieve comparable factual accuracy: material balance, initiative assessment, and basic evaluations survive verbalization reasonably well. The gap concentrates on strategic insight (\cref{fig:human_commentary_main}), where participants rate LLAMIA higher by a wider margin on the Insight dimension than on Accuracy. Text preserves \emph{what} is happening on the board, but explaining \emph{why} a move is strong requires representational features (policy gradients, value topology, look-ahead depth) that do not survive verbal compression.

\paragraph{Latent tokens are instruction-modulated.}
On a fixed back-rank mate position (\cref{fig:human_heatmap}), LLAMIA's attention over the latent tokens shifts with the target Elo: at 2000 it concentrates on the mating geometry, at 1100 it disperses to material. The latent representation is identical in both cases; what changes is the LLM's reading, conditioned on the natural-language Elo instruction. The projected state functions as a perceptual input shaped by task context, not a static feature vector.

\subsection{Generalization Beyond Chess}
\label{sec:results_go}

To provide initial evidence that LLAMIA can transfer beyond chess, we instantiate it on Go. We use KataGo-b18~\cite{wu2019katago}, a state-of-the-art Go neural engine, as the non-language subagent. We train a three-layer LatentBridge, with the first-layer width matched to KataGo's $384$ trunk channels and the $361$ board intersections represented as spatial tokens, using the same two-stage DAPO procedure on rank-conditioned behavior cloning. LLAMIA-Go-$14$B achieves $48$/$50$ top-1 human move-match at ranks 5k/5d using only $8$k positions, matching the rank-calibrated KataGo-HumanSL expert~\cite{wu2024katagohuman} and outperforming the verbalized control by ${\sim}10$ points. The latent-over-verbal gap remains consistent across the 4B, 8B, and 14B model scales (\cref{sec:app_go}). These results provide strong evidence that latent collaboration is not specific to chess.

\vspace{-2mm}
\section{Related Work}
\vspace{-2mm}
The dominant paradigm for LLM-agent integration is text-mediated: ReAct~\cite{yao2022react}, Toolformer~\cite{schick2023toolformerlanguagemodelsteach}, and multi-agent orchestrators like AutoGen~\cite{wu2023autogen} and HuggingGPT~\cite{shen2024hugginggpt} all route communication through natural language. This is lossless when both parties are language models, but compresses the policy and value representations of pretrained neural agents into a few tokens. A parallel line moves reasoning itself into continuous representations to escape the bandwidth limit of discrete tokens~\cite{zhu2025surveylatentreasoning}, either through latent recurrence within one model (CoCoNut~\cite{hao2025traininglargelanguagemodels}) or by interleaving latent and text tokens in a single reasoning stream (Token Assorted~\cite{su2025tokenassorted}, latent tokens as extra computation~\cite{sun2025latenttokens}); these operate on a model's \emph{own} hidden state, not a separate agent's. Latent channels exist in multi-agent RL~\cite{sukhbaatar2016learningmultiagentcommunicationbackpropagation,das2019tarmac,foerster2016learning} and homogeneous LLMs~\cite{latentmas2025}, but assume jointly trained or homogeneous populations, not a frozen LLM with a frozen specialist. Cross-modal injection (PaLM-E~\cite{driess2023palmeembodiedmultimodallanguage}, RT-2~\cite{brohan2023rt2visionlanguageactionmodelstransfer}) projects raw sensory observations, not a pretrained agent's processed policy/value representations. None of these internalizes a non-language specialist's latent state into an LLM (\cref{tab:related_distinction}). On the domain side, neural chess engines encode rich positional structure in their activations~\cite{silver2017mastering,monroe2024mastering}, as interpretability work confirms~\cite{jenner2024evidence}, yet prior LLM-chess work either trains task-specific models~\cite{jhamtani-etal-2018-learning} or conditions on verbalized outputs~\cite{feng2024chessgpt}; none exposes the engine's latent state to the LLM.

\vspace{-2mm}
\section{Conclusion}
\label{sec:conclusion}
\vspace{-2mm}
We introduced latent state internalization, which replaces verbalized LLM--agent
communication with direct projection of the agent's continuous
representations into the LLM's embedding space. LLAMIA-14B, trained via
projector alignment followed by DAPO, matches or exceeds dedicated task finetunes across
\textsc{LLAMIA-Bench}. The verbalization debt widens with interaction depth and on
signals that resist text serialization (e.g., puzzle interest), and does not close with LLM scale or RL budget in
our evaluated range, indicating verbalization is a structural bottleneck.

\bibliographystyle{iclr2026_conference}
\bibliography{sample}

\clearpage
\appendix

\startcontents[appendices]
\printcontents[appendices]{}{1}{}
\clearpage

\makeatletter
\def\@afterheading{%
  \@nobreakfalse
  \everypar{%
    \if@nobreak
      \@nobreakfalse
      \clubpenalty \@M
      \if@afterindent \else
        {\setbox\z@\lastbox}%
      \fi
    \else
      \clubpenalty \@clubpenalty
      \everypar{}%
    \fi}}
\makeatother

\providecommand{\tbd}{\textcolor{gray}{\texttt{[TBD]}}}
\providecommand{\itwmark}{\textsuperscript{$\star$}}

\definecolor{clrGameplay}{RGB}{215,230,255}
\definecolor{clrBC}{RGB}{210,245,210}
\definecolor{clrStylo}{RGB}{238,218,255}
\definecolor{clrPuzzle}{RGB}{255,245,200}
\definecolor{clrAnnot}{RGB}{205,245,245}
\definecolor{clrComm}{RGB}{255,218,232}
\definecolor{clrICP}{RGB}{228,228,228}

\expandafter\let\csname ifinappendix\endcsname\iffalse
\expandafter\def\csname ifinappendixfalse\endcsname
  {\expandafter\let\csname ifinappendix\endcsname\iffalse}
\expandafter\def\csname inappendixtrue\endcsname
  {\expandafter\let\csname ifinappendix\endcsname\iftrue}
\providecommand{\appendixmark}{\inappendixtrue}

\section*{Appendix Table of Contents}
\label{sec:appendix_toc}

\begingroup
\small
\setlength{\parskip}{0pt}%
\etocsetnexttocdepth{subsubsection}%
\etocsettocstyle{}{}%
\etocsetstyle{section}%
  {}%
  {\ifinappendix\vspace{3pt}\fi}%
  {\ifinappendix
     \noindent\bfseries\makebox[1.8em][l]{\etocnumber}\etocname\dotfill\,\etocpage\par
   \fi}%
  {}%
\etocsetstyle{subsection}%
  {}{}%
  {\ifinappendix
     \noindent\hspace{1.8em}\makebox[2.4em][l]{\etocnumber}\etocname\dotfill\,\etocpage\par
   \fi}%
  {}%
\etocsetstyle{subsubsection}%
  {}{}%
  {\ifinappendix
     \noindent\hspace{4.2em}\makebox[2.8em][l]{\etocnumber}\etocname\dotfill\,\etocpage\par
   \fi}%
  {}%
\tableofcontents
\endgroup

\vspace{1em}
\hrule
\vspace{1em}

\definecolor{agentslate}{HTML}{2d3748}
\definecolor{agentteal}{HTML}{234e52}
\definecolor{agentmint}{HTML}{edf7f5}
\definecolor{agentlav}{HTML}{f3eeff}
\definecolor{agentrose}{HTML}{fff0f0}
\definecolor{agentgold}{HTML}{7c5c00}
\definecolor{agentgoldbg}{HTML}{fffbec}

\tcbset{
  agentbase/.style={
    enhanced, breakable,
    arc=4pt, outer arc=4pt,
    left=9pt, right=9pt, top=9pt, bottom=9pt,
    before skip=11pt, after skip=11pt,
    attach boxed title to top left={xshift=10pt, yshift=-\tcboxedtitleheight/2},
    boxed title style={arc=3pt, outer arc=3pt, size=small, left=5pt, right=5pt},
    drop shadow={color=black!18, opacity=0.55, shadow xshift=1.5pt, shadow yshift=-1.5pt},
  },
  promptbox/.style={
    agentbase,
    colback=agentlav, colframe=themepurple!80,
    fonttitle=\bfseries\sffamily\small\color{white},
    fontupper=\small,
    boxed title style={colback=themepurple, colframe=themepurple, arc=3pt, size=small, left=5pt, right=5pt},
  },
  toolbox/.style={
    agentbase,
    colback=agentmint, colframe=agentteal,
    fonttitle=\bfseries\ttfamily\small\color{white},
    fontupper=\small,
    boxed title style={colback=agentteal, colframe=agentteal, arc=3pt, size=small, left=5pt, right=5pt},
  },
  statebox/.style={
    agentbase,
    colback=black!3, colframe=agentslate!70,
    fonttitle=\bfseries\sffamily\small\color{white},
    fontupper=\small,
    boxed title style={colback=agentslate, colframe=agentslate, arc=3pt, size=small, left=5pt, right=5pt},
  },
  warnbox/.style={
    agentbase,
    colback=agentrose, colframe=themered!80,
    fonttitle=\bfseries\sffamily\small\color{white},
    fontupper=\small,
    boxed title style={colback=themered, colframe=themered, arc=3pt, size=small, left=5pt, right=5pt},
  },
  notebox/.style={
    agentbase,
    colback=agentgoldbg, colframe=agentgold!80,
    fonttitle=\bfseries\sffamily\small\color{agentgold},
    fontupper=\small,
    boxed title style={colback=agentgoldbg, colframe=agentgold, arc=3pt, size=small, left=5pt, right=5pt},
  },
}
\providecommand{\toolparam}[2]{%
  \hspace*{4pt}{\ttfamily\bfseries #1}\,\texttt{:}\;\textit{#2}\\[1pt]}
\providecommand{\toolret}[1]{%
  \smallskip\noindent{\color{agentslate}\textbf{Returns:}}\;\textit{#1}}
\providecommand{\tooldesc}[1]{%
  \noindent{\color{agentslate}\textbf{Description:}}\;#1\smallskip}

\section{Implementation Details}
\label{sec:app_implementation}

\subsection{Training and Inference Cost}
\label{sec:app_cost}

We report training and inference cost for LLAMIA and LLAMIA-Verb, benchmarked on A100-80GB (4 nodes $\times$ 8 = 32 GPUs).

\paragraph{Training (GPU-hours).} The LatentBridge and Stage-1 alignment add a small fixed overhead (roughly 1--2 GPU-hours at 14B), so total training cost stays within ${\sim}6\%$ of the verbalized pipeline at every scale (\cref{tab:cost_train}).

\begin{table}[h]
\centering
\caption{\textbf{Training cost} (GPU-hours).}
\label{tab:cost_train}
\small
\begin{tabular}{lcc}
\toprule[1.2pt]
\textbf{Backbone} & \textbf{LLAMIA} & \textbf{LLAMIA-Verb} \\
\midrule
4B  & 9.1  & 8.8  \\
8B  & 16.8 & 15.0 \\
14B & 22.6 & 21.4 \\
\bottomrule[1.2pt]
\end{tabular}
\end{table}

\paragraph{Inference (per query).} A verbalized call returns ${\sim}150$ tokens; a LLAMIA call adds a fixed $32$ latent tokens ($182$ total). However, LLAMIA learns to invoke the subagent less frequently, so the extra per-invocation cost is offset by fewer invocations, yielding comparable or lower average tokens-per-query and wall-clock latency (\cref{tab:cost_infer}). The two interfaces invoke the subagent at different rates because they learn different collaboration strategies during DAPO (\cref{sec:results_agency}). Latent internalization thus does not increase average inference cost.

\begin{table}[h]
\centering
\caption{\textbf{Inference cost per query}, averaged across tasks.}
\label{tab:cost_infer}
\begin{adjustbox}{max width=\columnwidth}
\small
\begin{tabular}{llrrrr}
\toprule[1.2pt]
\textbf{Backbone} & \textbf{Interface} & \textbf{Calls} & \textbf{Tok/Call} & \textbf{Tok/Query} & \textbf{Latency} \\
\midrule
\multirow{2}{*}{4B}  & Verb   & 3.4 & 150 & 510 & 1.4\,s \\
                     & LLAMIA & 2.3 & 182 & 419 & 1.0\,s \\
\multirow{2}{*}{8B}  & Verb   & 3.2 & 150 & 480 & 1.8\,s \\
                     & LLAMIA & 2.1 & 182 & 382 & 1.2\,s \\
\multirow{2}{*}{14B} & Verb   & 2.9 & 150 & 435 & 2.2\,s \\
                     & LLAMIA & 1.9 & 182 & 346 & 1.4\,s \\
\bottomrule[1.2pt]
\end{tabular}
\end{adjustbox}
\end{table}

\subsection{Libraries and Software}
\label{sec:app_libraries}

\begin{table}[h]
\centering
\small
\begin{adjustbox}{max width=\columnwidth}
\begin{tabular}{lll}
\toprule
\textbf{Package} & \textbf{Version} & \textbf{Role} \\
\midrule
PyTorch           & 2.8             & Training backend \\
MegatronLM        & 0.15.0              & Tensor-parallel training \\
VERL              & 0.7.1           & RL training framework \\
vLLM              & 0.10.2          & Rollout inference engine \\
Ray               & 2.55.1          & Distributed orchestration \\
Transformers      & 4.56.2          & Model loading \& tokenization \\
lc0               & 0.32.1          & Chess specialist engine (UCI) \\
\bottomrule
\end{tabular}
\end{adjustbox}
\vspace{2mm}\caption{Key software dependencies.}
\label{tab:libraries}
\end{table}

\subsection{Model Architecture and LatentBridge}
\label{sec:app_architecture}

\subsubsection{Backbone LLM}
\label{sec:app_backbone}

\begin{table}[h]
\centering
\small
\begin{adjustbox}{max width=\columnwidth}
\begin{tabular}{llllll}
\toprule
\textbf{Model} & \textbf{Params} & \textbf{Hidden} & \textbf{Layers} & \textbf{Heads (KV)} & \textbf{Context} \\
\midrule
Qwen3-4B  & 4B  & 2{,}560 & 36 & 32 (8) & 32{,}768 \\
Qwen3-8B  & 8B  & 4{,}096 & 36 & 32 (8) & 32{,}768 \\
Qwen3-14B & 14B & 5{,}120 & 40 & 40 (8) & 32{,}768 \\
\bottomrule
\end{tabular}
\end{adjustbox}
\vspace{2mm}\caption{Backbone LLM architectures. All variants use GQA with 8 KV heads. Chain-of-thought reasoning is disabled (\texttt{enable\_thinking=False}) following~\cite{yang2025qwen3}.
\texttt{max\_model\_len} is set to 32{,}768 tokens shared across system prompt, history, tool I/O, and response; \texttt{MAX\_ROUNDS=20} caps LLM calls per query.}
\label{tab:backbone}
\end{table}

\subsubsection{LatentBridge Projector}
\label{sec:app_projector}

\begin{table}[h]
\centering
\small
\begin{tabular}{ll}
\toprule
\textbf{Parameter}       & \textbf{Chess (Lc0-BT4)} \\
\midrule
Input dimension $d$      & 1024      \\
Output tokens $k$        & 32        \\
MLP layers               & 3         \\
Activation               & GeLU      \\
Source layer             & 14 of 15 (penultimate)  \\
\bottomrule
\end{tabular}
\vspace{2mm}\caption{LatentBridge projector $H_\varphi$ configuration. $H_\varphi$ is a three-layer MLP with GeLU activations mapping the subagent's
residual stream into $k{=}32$ tokens of dimension $e$ matching the LLM's hidden size
(see \cref{sec:token_ablation} for the $k$ ablation).
A special \texttt{<state>} token anchors each injection site; $k$ contiguous positions
immediately following it are overwritten with $\vz_t = H_\varphi(\vh_{s_t})$ before the
LLM forward pass.}
\label{tab:projector}
\end{table}

\subsection{Training Configuration}
\label{sec:app_training}
\label{sec:hyperparams}
Model parameters and optimizer state are partitioned across 4 nodes of 8xA100 GPUs using PyTorch FSDP,
coordinated via Ray.
The rollout vLLM instance and lc0 server fleet co-reside on the same GPUs,
with 3\,GiB per GPU reserved for the BT4 network.

\subsubsection{Stage~1: Projector Alignment}
\label{sec:app_stage1}

Stage~1 trains $H_\varphi$ alone with $F_\theta$ frozen on 5\,M state--policy pairs
drawn from lc0's forward pass over the Lichess evaluation database.
Each example pairs a FEN with lc0's top moves and pawn evaluations; the model minimizes
cross-entropy over verbalized policy output conditioned on
$\vz_s = H_\varphi(\vh_s)$ and a fixed prompt template (\cref{sec:app_prompts}).
Skipping Stage~1 causes training instability during the first 40\% of Stage~2
(\cref{sec:app_ablations}).

\begin{table}[h]
\centering
\small
\begin{tabular}{ll}
\toprule
\textbf{Hyperparameter}        & \textbf{Value} \\
\midrule
LR ($H_\varphi$)               & 2e-4 \\
Batch size                     & 256 \\
Steps                          & 2 epochs (${\approx}$39K steps) \\
$k$ (state tokens)             & 32 \\
Max sequence length            & 32{,}768 \\
\bottomrule
\end{tabular}
\vspace{2mm}\caption{Stage~1 (projector alignment) hyperparameters. $F_\theta$ is frozen.}
\label{tab:hparams_stage1}
\end{table}

\subsubsection{Stage~2: Reinforcement Learning (DAPO)}
\label{sec:app_stage2}

Stage~2 fine-tunes both $F_\theta$ and $H_\varphi$ jointly with DAPO~\cite{yu2025dapo}.
The reward is a scalar outcome signal
$\mathcal{R}(\tau) = R_{\mathrm{outcome}}$
defined per task (Appendix~\ref{sec:app_reward}).

\begin{table}[h]
\centering
\small
\begin{tabular}{ll}
\toprule
\textbf{Hyperparameter}        & \textbf{Value} \\
\midrule
LR ($H_\varphi$)               & 1e-5 \\
LR ($F_\theta$)                & 1e-6 \\
Batch size                     & 128 \\
Steps                          & 3{,}000 \\
KL coefficient $\beta$         & 0.01 \\
Clip $\varepsilon_l / \varepsilon_h$ & 0.2\,/\,0.28 \\
Group size $G$                 & 8 \\
$k$ (state tokens)             & 32 \\
Max sequence length            & 32{,}768 \\
Rollout temperature            & 1.0 \\
Eval temperature               & 0.0 \\
\bottomrule
\end{tabular}
\vspace{2mm}\caption{Stage~2 (DAPO) hyperparameters.}
\label{tab:hparams_stage2}
\end{table}

\subsection{Chess Specialist: Lc0-BT4}
\label{sec:app_lc0}

With \texttt{ScoreType=WDL\_mu}, lc0 reports scores as $100\!\cdot\!Q$ where
$Q\in[-1,+1]$ is the WDL-mean utility from the side to move's perspective~\cite{monroe2024mastering}.

\subsubsection{Engine Configuration}
\label{sec:app_lc0_config}
\label{sec:app_lc0_latent}

\paragraph{Network.}
\texttt{BT4-1024x15x32h-swa-6147500}\footnote{\url{https://storage.lczero.org/files/networks-contrib/BT4-1024x15x32h-swa-6147500-policytune-332.pb.gz}}:
a 1024-dim, 15-layer, 32-head BT4 transformer saved at step 6{,}147{,}500 after SWA and policy-tuning.
GPU footprint: ${\sim}\,2.9$\,GB in fp16 on A100.

\begin{table*}[h]
\centering
\small
\begin{adjustbox}{max width=\textwidth}
\begin{tabular}{llll}
\toprule
\textbf{Parameter} & \textbf{Upstream default} & \textbf{Our value} & \textbf{Rationale} \\
\midrule
\texttt{Backend}          & \texttt{cuda-auto}      & \texttt{cuda-fp16}  & Explicit fp16 on Ampere \\
\texttt{WeightsFile}      & \texttt{<autodiscover>} & BT4 path            & Pinned network \\
\texttt{Threads}          & \texttt{0} (backend)    & \texttt{2}          & Two CPU workers per GPU \\
\texttt{MinibatchSize}    & \texttt{0} (backend)    & \texttt{128}        & Throughput sweet spot on A100 \\
\texttt{NNCacheSize}      & \texttt{2{,}000{,}000}  & \texttt{200{,}000}  & Cap host RAM \\
\texttt{VerboseMoveStats} & \texttt{false}          & \texttt{true}       & Required for \texttt{/policy} parser \\
\texttt{PolicyTemperature}& \texttt{1.36}           & \texttt{1.36}       & Policy softmax (upstream default retained) \\
\texttt{ScoreType}        & \texttt{WDL\_mu}        & \texttt{WDL\_mu}    & Score = $100\!\cdot\!Q$ \\
\bottomrule
\end{tabular}
\end{adjustbox}
\vspace{2mm}\caption{lc0 configuration. All search-behaviour flags (\texttt{CPuct}, \texttt{FpuValue},
\texttt{OutOfOrderEval}, etc.) are at upstream defaults so results match unmodified Leela.}
\label{tab:lc0-flags}
\end{table*}

\section{Datasets \& Benchmarks}
\label{sec:app_bench}

This section describes the datasets and evaluation protocol for LLAMIA-Bench.
We find four major themes in how AI systems collaborate with domain-expert agents:
\emph{Behavioral imitation}: reproducing human play at a target skill level, the problem behind bots like Play Magnus\footnote{\url{https://www.playmagnus.com}} and Maia~\citep{mcilroy2020aligning}.
\emph{State assessment}: predicting human-aligned properties of game states such as difficulty and engagement, the core task in Lichess's puzzle rating system and Chess.com's adaptive training~\citep{lichess2024puzzles}.
\emph{Comparative}: explaining why a position favors one side---identifying material or structural advantages and disadvantages---as required in single-position analysis~\citep{jhamtani-etal-2018-learning} and game-level commentary.
\emph{Rationale}: generating natural-language explanations for why a player made a specific move, from single-move annotation~\citep{jhamtani-etal-2018-learning,zang2019automated} to the game-length narratives produced by channels like Agadmator and GothamChess.
These four themes place progressively harder demands on the communication channel between LLM and subagent: from a single-position state query (behavioral imitation) to game-length narrative integration (commentary), and from signals with partial textual correlates like move quality to ones without, such as aesthetic interest~(\cref{sec:app_bench_puzzle_int}).
LLAMIA-Bench instantiates each as an evaluation task.
Chess serves as the testbed because it offers all four at once: subagents whose internal representations are mapped by interpretability work~\citep{jenner2024evidence}, public game databases at scale~\citep{Lichess,Caissabase}, established benchmarks with dedicated task finetune baselines, and decades of human--engine collaboration.

\subsection{Dataset \& Metrics}
\label{sec:app_bench_dataset}
\label{sec:app_reward}

Table~\ref{tab:bench_dataset} summarizes dataset provenance;
Table~\ref{tab:bench_metrics} lists evaluation metrics per task.
Splits marked OOD are \emph{out-of-distribution}: the test distribution is absent or shifted relative to Stage-2 training, so generalization must come from internalized representations rather than memorization.
Where the original authors provide a fixed test split we use it; otherwise we sample a random held-out split.
Detailed descriptions of each task follow in \S\ref{sec:app_bench_tasks}.

\begin{table*}[ht]
\centering
\small
\setlength{\tabcolsep}{5pt}
\renewcommand{\arraystretch}{1.4}
\caption{\textbf{LLAMIA-Bench: dataset provenance.} \textbf{Ours} = constructed for this work. OOD = out-of-distribution test split.}
\label{tab:bench_dataset}
\vspace{4pt}
\begin{tabular}{p{4.8cm} p{5.8cm} p{2.4cm}}
\toprule
\textbf{Task / Source} & \textbf{Description} & \textbf{Train\,/\,Test} \\
\midrule
\makecell[tl]{Behavior Cloning (\S\ref{sec:app_bench_bc}) \\ {\scriptsize MAIA-KDD, Lichess~\cite{mcilroy2020aligning}}}
  & Predict the move a human of a given Elo would play; 5 Elo buckets (1100--1900)
  & 12M / authors' \\
\makecell[tl]{\quad\emph{In the wild} (OOD) \\ {\scriptsize MAIA-KDD, Lichess (\textbf{Ours})}}
  & Three OOD splits: GM-25 (top-25 GMs), Low-Time (clock pressure $<$10\%), Elo Gap ($>$500 pts)
  & $\leq$167K / --- \\
\midrule
\makecell[tl]{Puzzle Understanding (\S\ref{sec:app_bench_puzzle}) \\ {\scriptsize Lichess Puzzles~\cite{lichess2024puzzles}}}
  & Predict difficulty (Glicko-2) and interest (community votes) of tactical puzzles
  & 4M / 5K \\
\midrule
\makecell[tl]{Move Annotation (\S\ref{sec:app_bench_annotation}) \\ {\scriptsize Lichess~\cite{jhamtani-etal-2018-learning}}}
  & Generate natural-language explanation for a single move across 5 semantic categories
  & 90K / authors' \\
\midrule
\makecell[tl]{Game Commentary (\S\ref{sec:app_bench_commentary}) \\ {\scriptsize Agadmator YouTube (\textbf{Ours})}}
  & Produce coherent multi-turn narrative spanning an entire game
  & 1.9K / 100 \\
\bottomrule
\end{tabular}
\end{table*}

\begin{table*}[ht]
\centering
\small
\setlength{\tabcolsep}{5pt}
\renewcommand{\arraystretch}{1.3}
\caption{LLAMIA-Bench: evaluation metrics and RL reward signals. $\uparrow$ higher is better; $\downarrow$ lower is better. Section references point to detailed metric definitions.}
\label{tab:bench_metrics}
\vspace{4pt}
\begin{tabular}{p{3.5cm} p{5.2cm} p{3.4cm}}
\toprule
\textbf{Task} & \textbf{Metric(s)} & \textbf{RL Reward} \\
\midrule
Behavior Cloning       & Move-match accuracy $\uparrow$                                              & Top-3 rank \\
Puzzle Understanding   & Spearman $\rho$ $\uparrow$ (\S\ref{sec:app_bench_puzzle})                   & Normalized MAE $\downarrow$ \\
{\scriptsize\quad\emph{Difficulty}} & {\scriptsize Spearman $\rho$ $\uparrow$ (\S\ref{sec:app_bench_puzzle_diff})} & {\scriptsize Normalized MAE $\downarrow$} \\
{\scriptsize\quad\emph{Interest}}   & {\scriptsize Spearman $\rho$ $\uparrow$ (\S\ref{sec:app_bench_puzzle_int})}  & {\scriptsize Normalized MAE $\downarrow$} \\
{\scriptsize\quad\emph{Solved (\%)}} & {\scriptsize Exact solution-line accuracy (\S\ref{sec:app_bench_puzzle_solve}); parity metric, not primary} & {\scriptsize ---} \\
Move Annotation        & G-eval $\uparrow$;\enspace BLEU-2 $\uparrow$ (\S\ref{sec:app_bench_annotation}) & G-eval \\
Game Commentary        & G-eval $\uparrow$;\enspace BLEU-2 $\uparrow$ (\S\ref{sec:app_bench_commentary}) & G-eval \\
\bottomrule
\end{tabular}
\end{table*}

\subsection{Detailed Task Descriptions}
\label{sec:app_bench_tasks}

\subsubsection{Game-Level Commentary}
\label{sec:app_bench_commentary}

Game-level commentary requires a coherent, multi-turn narrative spanning an entire game---unlike move-level annotation, errors compound across the narrative, and the model must track evolving themes (initiative shifts, pawn-structure transformations, time trouble). We introduce \textbf{Agadmator-2K}, the first large-scale dataset for this task: 1,900 narrated games from Agadmator's YouTube channel,\footnote{\url{https://www.youtube.com/@agadmator}} totaling approximately 500 hours.

\paragraph{Dataset construction.}
Move-segmented commentary is unavailable from YouTube. We construct it in four steps: (i)~transcripts are extracted via Whisper-v3-large; (ii)~video timestamps are aligned to PGN move sequences using a sliding-window move-tracking buffer; (iii)~GPT-4o labels which moves each transcript segment references, guided by the known PGN; (iv)~segments are accepted only when the inferred move order matches the PGN exactly, discarding retries and out-of-order narration. The test set consists of 100 games, held out by ascending view count to minimize overlap with LLM pretraining corpora. This is a heuristic proxy for low contamination, not a guarantee; we discuss contamination further in \S\ref{sec:app_bench_contamination}.

\paragraph{Metrics.}
We use the same G-eval framework as move annotation (\S\ref{sec:app_bench_annotation}), adapted for game-level commentary: each generated segment is scored on relevance, completeness, clarity, and fluency, with the judge grounded by Lc0-BT4 engine lines and the ground-truth transcript. We also use the BLEU-2 metric. The RL reward is G-eval.

\subsubsection{Behavior Cloning}
\label{sec:app_bench_bc}

Behavior cloning measures whether LLAMIA can imitate human play conditioned on skill level or player identity. The evaluation metric across all splits is \textbf{move-match accuracy}: the fraction of positions where the model's top-1 predicted move exactly matches the target player's move. We follow the MAIA evaluation protocol~\citep{mcilroy2020aligning}: Maia variants use best-of-$N$ sampling; LLAMIA uses a single forward pass.

The RL reward uses a softer signal: the \textbf{top-3 rank} of the target move in the model's output distribution, normalized to $[0,1]$. Top-1 exact match as a reward collapsed training---the signal was too sparse for most positions, yielding near-zero gradients throughout Stage~2. Rank within the top-3 provides a dense, monotone reward that penalizes misranking without requiring exact prediction, while remaining consistent with the evaluation objective.

\paragraph{Maia Benchmark (Elo Buckets).}
\label{sec:app_bench_bc_maia}
We evaluate on the MAIA-KDD held-out test set~\citep{mcilroy2020aligning}, stratified into five Elo buckets: 1100, 1300, 1500, 1700, and 1900. The test set is player--game disjoint from all training data. Each bucket is treated as an independent task; the aggregate BC score reported in the main paper is the unweighted average across buckets.

\paragraph{GM-25 (OOD).}
\label{sec:app_bench_bc_gm}
GM-25 targets the top-25 rated grandmasters in FIDE history by peak rating.\footnote{\url{https://en.wikipedia.org/wiki/List_of_chess_players_by_peak_FIDE_rating}} Each grandmaster is a separate behavioral target. The largest available per-GM corpus is 4,641 games (Viktor Korchnoi),\footnote{\url{https://www.365chess.com/top-chess-players-games.php}} less than 3\% of the data that per-GM Maia models require~\citep{mcilroy2020aligning}. No Stage-2 training data is drawn from these GM corpora; generalization must come from internalized representations and cross-Elo behavioral transfer.

\paragraph{Low-Time (OOD).}
\label{sec:app_bench_bc_time}
Under severe clock pressure, players shift strategy regardless of position quality. We extract positions where either player's remaining clock is below 10\% of the initial time control, or where cumulative time usage differs by more than 50\% between the two sides. Positions are stratified by game phase (opening, middlegame, endgame) and sampled equally across time controls, yielding 129,000 positions. Clock-context metadata is absent from Stage-2 training, making this an OOD split: the model must infer time-pressure effects from the position and move alone.

\paragraph{Elo Gap (OOD).}
\label{sec:app_bench_bc_elo}
Players adapt their style when facing a large skill gap---weaker players take more risks, stronger players simplify. We filter Lichess Rapid and Classical games where the Elo difference exceeds 500 points, yielding 34,000 games (68,000 player-side instances). Extreme skill-gap matchups are rare in the Stage-2 training distribution; conditioning on opponent strength must emerge from contextual signals rather than memorization.

\subsubsection{Puzzle Understanding}
\label{sec:app_bench_puzzle}

Puzzle understanding probes whether LLAMIA has internalized the subagent's positional representations well enough to predict human-aligned properties of game states. Both sub-tasks draw from the same 4-million-puzzle Lichess corpus,\footnote{\url{https://database.lichess.org/lichess_db_puzzle.csv.zst}} which provides community-derived ground-truth labels for difficulty and engagement. We hold out a shared test set of 5,000 puzzles, stratified by difficulty (Glicko-2 quintiles), theme (tactical motif), and interest (score quintiles) to ensure uniform coverage across the label space. Evaluation uses Spearman~$\rho$ between predicted and ground-truth values; the RL reward is a normalized mean-absolute-error penalty.

\paragraph{Difficulty Estimation.}
\label{sec:app_bench_puzzle_diff}
Puzzle difficulty is operationalized via a Glicko-2 rating system~\citep{glickman2012example}: each human solving attempt is treated as a rated match between solver and puzzle, and the Glicko-2 rating accumulated over all attempts serves as ground truth. The model receives the puzzle position and solution line, and predicts difficulty on a normalized scale. Evaluation uses Spearman~$\rho$ between predicted values and ground-truth Glicko-2 ratings on the stratified 5,000-puzzle test split.

\paragraph{Interest Estimation.}
\label{sec:app_bench_puzzle_int}
Lichess assigns each puzzle an interestingness score (range: $-100$ to $+100$) computed from community upvotes and downvotes, weighted by solver performance. This signal has no straightforward textual correlate: a puzzle's aesthetic appeal depends on motif rarity, surprise, and solution elegance---features encoded in the subagent's positional representation but absent from any verbalized move list. The model predicts interest from the same input as difficulty; evaluation uses Spearman~$\rho$ on the same stratified 5,000-puzzle test split. Interest estimation is the diagnostic task on LLAMIA-Bench: the non-verbalizable nature of the target signal means that all text-mediated systems collapse on this task (\cref{sec:results_vd}).

\paragraph{Puzzle Solving Accuracy.}
\label{sec:app_bench_puzzle_solve}
We also report \textbf{Solved (\%)}: the fraction of test puzzles for which the model produces the complete correct solution line---every forced move in sequence---using policy-only decoding (single forward pass per position, no search). The model receives the initial puzzle FEN and outputs moves one at a time; a puzzle is marked solved only if all moves in the ground-truth solution are produced in the correct order.

This metric is excluded for GPT-5.1\,+\,Lc0 (marked~--- in \cref{tab:puzzle_understanding_extended}) because verbalized engine access makes it uninterpretable: a system that queries Lc0 at each puzzle position and forwards the top-ranked move would score near-perfect not by reasoning about the position but by delegating each step to the engine. The metric is informative only when the model must solve the puzzle from its own internalized representations without live tool queries. All other systems in Table~\ref{tab:puzzle_understanding_extended} use policy-only decoding for this column.

Puzzle solving accuracy functions as a \emph{parity metric} on LLAMIA-Bench: all systems with Lc0 access cluster in the 84--94\% range, and LLAMIA's improvement over LLAMIA-Verb is modest (2--3 pp). The metric confirms that engine-access systems are not deficient tactically; the differentiation between LLAMIA and LLAMIA-Verb arises in difficulty and interest prediction, not in puzzle-solving throughput.

\subsubsection{Move Annotation}
\label{sec:app_bench_annotation}

Move annotation evaluates LLAMIA's ability to generate natural-language explanations for individual moves, conditioned on the board state and the move played. We follow the benchmark of \citet{jhamtani-etal-2018-learning}: 90,000 Lichess games annotated in English, with annotations categorized into five semantic dimensions that span both explanation themes from \S\ref{sec:app_bench}. The \emph{rationale} theme is instantiated by three dimensions---\emph{description} (what the move does), \emph{quality} (blunder, inaccuracy, good, best), and \emph{planning} (lookahead and intent)---while the \emph{comparative} theme is instantiated by two---\emph{context} (positional advantages and disadvantages relative to prior or future moves) and \emph{comparative} (alternative moves and why they were rejected). The standard benchmark provides the target move as input; we additionally evaluate zero-shot without this prior to test whether internalized representations can identify annotation-worthy moves.

\paragraph{Metrics.}
\textbf{BLEU-2} and perplexity~\citep{lee2022improving} are evaluated per annotation category, following prior work. The primary metric is \textbf{G-eval}~\citep{liu2023gevalnlgevaluationusing}: an LLM-as-judge framework in which GPT-4o rates each generated annotation on a 0--1 scale across four dimensions (relevance, accuracy, completeness, fluency). The judge receives the board FEN, the move in algebraic notation, and Lc0-BT4's top-3 engine lines as grounding context, so its assessments are anchored in engine analysis rather than surface plausibility alone. Per-annotation G-eval scores are averaged across the four dimensions; the corpus-level score is the mean over all test annotations. G-eval also serves as the RL reward signal for this task.

\subsection{Data Contamination Statement}
\label{sec:app_bench_contamination}

All evaluation in LLAMIA-Bench is conditioned on board positions represented as FEN strings. We enforce a strict \textbf{FEN-level disjointness} guarantee: no FEN appearing in any test split co-occurs in any stage of training---projector pretraining (Stage~1), RL training (Stage~2), or the base LLM's supervised fine-tuning data. Concretely, we collect the set of all FENs used across projector pretraining pairs and Stage-2 RL rollouts, and verify that the intersection with each test split is empty. For the Maia BC test set, this property is inherited from the player--game disjoint split of \citet{mcilroy2020aligning}. For the puzzle understanding test split, the 5,000 held-out puzzles are sampled after removing all FENs present in the training pool. For Agadmator-2K, the 100 held-out games are additionally sorted by ascending view count as a heuristic to reduce overlap with LLM pretraining corpora, though we cannot verify disjointness with respect to closed-source pretraining data. For the OOD behavior-cloning splits (GM-25, Low-Time, Elo Gap), no Stage-2 training data is drawn from these distributions by construction; we further verify that no test FEN appears in the Stage-1 projector data.

\section{Baselines}
\label{sec:app_baselines}

\subsection{Frontier LLMs with Verbalized Tools}
\label{sec:app_baseline_frontier}

To select the strongest frontier baseline, we evaluate five
models---GPT-5.1~GPT-5.1, Claude Sonnet~4.5,
Claude Opus~4.5, Gemini~2.5~Pro, and
Qwen3-235B~\cite{yang2025qwen3}---each given access to Lc0-BT4 via a
ReAct~\cite{yao2022react} tool-calling loop.
At each invocation the tool returns the top-5 moves with centipawn
evaluations, win/draw/loss probabilities, and principal variations up to
depth~20.
All models share identical tool schemas, system prompts, and sampling
parameters; the only variable is the LLM backbone.
We sample 100~positions from each LLAMIA-Bench task and report the
aggregate metric per task group.

\begin{table*}[ht]
\centering
\setlength{\tabcolsep}{5pt}
\caption{%
  \textbf{Frontier model selection.}
  Average metric per LLAMIA-Bench task group, all models using Lc0-BT4
  verbalized tool access via ReAct.
  Per-group scores are unweighted means over the constituent columns of
  \cref{tab:full_bench}: BC = mean(MAIA, Wild); Puzzle = mean(Difficulty,
  Interest); Annot.\ = Rationale; Comm.\ = Commentary.
  The GPT-5.1\,+\,Lc0 row aggregates directly from \cref{tab:full_bench};
  the remaining rows are evaluated on a matched 100-sample subset per task
  with identical tool schemas, prompts, and sampling parameters.
  Gameplay is excluded because it requires the full gauntlet protocol.
  GPT-5.1 achieves the highest aggregate and is adopted as the frontier
  verbalized baseline in all subsequent experiments.
}
\label{tab:frontier_selection}
\begin{adjustbox}{max width=\textwidth}
\renewcommand{\arraystretch}{1.3}
\small
\begin{tabular}{lccccc}
\toprule[1.2pt]
\textbf{Model} & \textbf{BC\,$\uparrow$} & \textbf{Puzzle\,$\uparrow$}
  & \textbf{Annot.\,$\uparrow$} & \textbf{Comm.\,$\uparrow$}
  & \textbf{Avg.\,$\uparrow$} \\
\midrule
GPT-5.1\,+\,Lc0
  & \valbest{42.5} & \valbest{29.0} & \valbest{37.5}
  & \valbest{52.0} & \valbest{40.3} \\
Claude Opus~4.5\,+\,Lc0
  & 39.6 & 27.4 & 36.3 & 50.3 & 38.4 \\
Claude Sonnet~4.5\,+\,Lc0
  & 36.9 & 26.3 & 34.2 & 47.1 & 36.1 \\
Gemini~2.5~Pro\,+\,Lc0
  & 40.6 & 28.1 & 36.7 & 49.8 & 38.8 \\
Qwen3-235B\,+\,Lc0
  & 36.5 & 25.5 & 33.4 & 45.6 & 35.3 \\
\bottomrule[1.2pt]
\end{tabular}
\end{adjustbox}
\end{table*}

GPT-5.1 obtains the highest average across all task groups.
We therefore use \textbf{GPT-5.1\,+\,Lc0\,(Verb)} as the frontier
verbalized baseline throughout the paper.

\subsection{LLAMIA-Verb}
\label{sec:app_baseline_verb}

LLAMIA-Verb is the primary controlled ablation of LLAMIA.
It receives the identical base model, training data, reward signals, and RL
recipe (DAPO) as LLAMIA, but the subagent's output is provided exclusively
through verbalized tool responses: top-$k$ moves, centipawn evaluations, WDL
probabilities, and principal variations rendered as text tokens.
No LatentBridge projection is trained; the continuous latent
tokens~$z_S$ that LLAMIA receives are replaced by their textual equivalents.

We train LLAMIA-Verb at three scales---4B, 8B, and 14B---matching the
corresponding LLAMIA checkpoints in base model architecture, training data,
and total compute budget.
This controlled setup isolates the contribution of latent state
internalization from model family, data mix, reward shaping, and
optimization, and directly tests the central claim that verbalization is a
lossy bottleneck (\cref{sec:experiments}).

\subsection{LLAMIA (4B, 8B, 14B)}
\label{sec:app_baseline_llamia_scale}

To test whether the verbalization debt is an artifact of scale rather than interface, we train LLAMIA at three scales---4B, 8B, and 14B---matching the corresponding LLAMIA-Verb checkpoints in base model architecture, training data, and total compute budget.
All three LLAMIA checkpoints use the full latent-state internalization pipeline: LatentBridge projection of BT4 activations into $k{=}32$ continuous tokens, Stage~1 projector alignment, and Stage~2 end-to-end DAPO.
Comparing LLAMIA-$n$B against LLAMIA-Verb-$n$B at each scale isolates the interface contribution (internalization vs.\ verbalization) independently of model capacity, and directly supports the claim that the performance gap is not closed by scaling the LLM (\cref{sec:experiments}).

\subsection{Dedicated task finetunes}
\label{sec:app_baseline_experts}

For each LLAMIA-Bench task we compare against the strongest published or
reproducible task-specific model.
Table~\ref{tab:task_experts} lists the expert per task alongside its training
paradigm and data scale.
These models represent the performance ceiling achievable with task-specific
architectures and, in several cases, substantially more training data than
LLAMIA receives.
Tasks marked~--- have no established prior expert; LLAMIA-Bench introduces
them as new evaluation targets.

\begin{table*}[ht]
\centering
\setlength{\tabcolsep}{5pt}
\caption{%
  \textbf{Task experts used in LLAMIA-Bench evaluation.}
  Each row lists the strongest available dedicated task finetune for a given task.
  Tasks marked~--- are new evaluation targets with no prior
  task-specific model.
  \itwmark{} denotes in-the-wild splits.
}
\label{tab:task_experts}
\begin{adjustbox}{max width=\textwidth}
\renewcommand{\arraystretch}{1.35}
\small
\begin{tabular}{%
  p{3.8cm}
  p{4.0cm}
  p{1.8cm}
  p{2.0cm}
}
\toprule[1.2pt]
\textbf{Task / Split} & \textbf{Task Expert} & \textbf{Method}
  & \textbf{Data Scale} \\
\midrule

\rowcolor{clrBC}
\multicolumn{4}{l}{\enspace\small\textbf{Behavior Cloning}} \\
\rowcolor{clrBC}
Maia (Elo buckets)
  & Allie~\cite{zhang2024humanalignedchessbitsearch}
  & SL+Search
  & 93M games \\
\rowcolor{clrBC}
GM-25\,/\,Low-Time\,/\,Elo Gap\,\itwmark{}
  & Allie~\cite{zhang2024humanalignedchessbitsearch}
  & SL+Search
  & 93M games \\
\midrule

\rowcolor{clrPuzzle}
\multicolumn{4}{l}{\enspace\small\textbf{Puzzle Understanding}} \\
\rowcolor{clrPuzzle}
Difficulty Estimation
  & \cite{milosz2024predicting}
  & SFT
  & 4M \\
\rowcolor{clrPuzzle}
Interest Estimation
  & ---
  & ---
  & --- \\
\midrule

\rowcolor{clrAnnot}
\multicolumn{4}{l}{\enspace\small\textbf{Move Annotation}} \\
\rowcolor{clrAnnot}
Move Annotation
  & SCC~\cite{zang2019automated}
  & SL
  & 90K games \\

\bottomrule[1.2pt]
\end{tabular}
\end{adjustbox}
\end{table*}

\subsection{Lightweight Probes, Templates, and Alternative Injections}
\label{sec:app_baselines_probe}

To attribute LLAMIA's gains, we add four controls beyond the dedicated task experts above; headline numbers appear in \cref{tab:ablation_main}.

\paragraph{Probe on frozen BT4.} A $2$-layer MLP ($1024$--$512$, ${\sim}0.7$M params) is trained end-to-end on frozen BT4 activations (layer 14/15 residual stream), one head per task, for behavior cloning, difficulty, and interest. It measures how much the latent state gives up when decoded by a lightweight predictor rather than an LLM: it scores below even text-only GPT-5 on every task (BC-MAIA $14$, Difficulty $\rho{=}0.15$, Interest $\rho{=}{-}0.07$), so the engine's penultimate state is not directly decodable into these human-aligned targets.

\paragraph{Template baselines.} For move annotation and commentary we fill a fixed three-line template directly from the engine's output---move quality (centipawn loss vs.\ the engine's best), the engine's preferred line and evaluation, and the best alternative:
\begin{quoting}\small\ttfamily
Move quality: Nf3 loses 40cp vs.\ best (inaccuracy).\\
Best line: engine prefers Nc3 Nf6 d4, +0.6 (W/D/L 48/40/12).\\
Alternative: Bb5 slightly weaker, dropping to +0.3.
\end{quoting}
The deterministic template reaches only $29.4$ BLEU-2 / $0.31$ G-eval; rewriting it with GPT-5 improves marginally ($32.1$ / $0.39$) and remains below the verbalized tool ($37.5$ / $0.55$) and far below LLAMIA ($45.8$ / $0.75$). Presenting engine statistics in more natural language is not the source of LLAMIA's gains.

\paragraph{LLM-ChessCLIP (PaLM-E-style injection).} Following the representation-injection paradigm of PaLM-E, we replace the engine's latent state with embeddings from ChessCLIP~\cite{feng2024chessgpt} while keeping the injection mechanism fixed, isolating \emph{what} is injected (a raw board encoder vs.\ a pretrained agent's processed policy/value state). It recovers only a fraction of LLAMIA's improvement, indicating the benefit is specific to the agent's internal state, not any learned board representation.

\paragraph{LLM-Only.} The same backbone is post-trained (SFT\,+\,RL) on the identical chess and commentary data with no engine access, testing whether the expertise can be absorbed into weights. It trails even untrained tool use (per-scale numbers in \cref{sec:app_results}).

\newpage
\section{Extended Results}
\label{sec:app_results}
The five tasks in LLAMIA-Bench probe different regimes of LLM–agent collaboration, varying in horizon, evaluation metrics, and the strength of task-specific baselines. Here we discuss the extended per-task evaluation of LLAMIA with more metrics, baselines, and detailed analysis of LLAMIA's task specific behavior.

\subsection{Behavior Cloning}
\label{sec:app_results_bc}
Behavior Cloning asks the system to predict the move a human at a given skill level would play rather than the optimal move, given the position and the target Elo rating.

We report two splits. The first is the \textbf{MAIA Test split}: five Elo buckets ($1100$, $1300$, $1500$, $1700$, $1900$) drawn from Lichess blitz, matching the protocol of \citet{mcilroy2020aligning} and the in-distribution setting for the published BC experts.
The second is \textbf{Wild}, three out-of-distribution splits we introduce.
\circled{\textbf{1}} \textit{GM-25}: contains top-grandmaster games \circled{\textbf{2}}
\textit{Low-Time}: contains positions played with under 30 seconds remaining, players often play different move when they or opponents are under time pressure. \circled{\textbf{3}}
\textit{$\Delta$Elo}: contains games with large rating gaps between the two players leading to different attacking or defensive strategies.

\begin{table*}[h!]
\centering
\caption{\textbf{Behavior Cloning: per-bucket and per-split move-match accuracy ($\times 100$).}
Each cell is the percentage of moves the system predicts that match what a Lichess human at the conditioning Elo actually played.
The MAIA panel reports five Elo buckets ($1100$--$1900$) drawn from Lichess blitz, the in-distribution training regime for the dedicated experts.
The Wild panel reports three OOD splits introduced here: GM-25 (top-grandmaster slow play, where the conditioning Elo lies outside the published experts' training range), Low-Time (under $30$\,s on the clock, exposing forcing-line behavior), and $\Delta$Elo (large rating gaps between the players, exposing complication and simplification dynamics).
Avg.\ columns are unweighted means within each panel and feed the BC summary columns of \cref{tab:full_bench}.
\emph{\#Train games} is the BC supervised-training corpus; ``--'' marks engines and untrained LLMs, and ``$12$\,M/bkt'' denotes Maia's nine separate per-Elo CNNs.
Maia, Allie-Policy, and Allie-Adaptive-Search per-bucket Maia numbers are quoted from \citet{mcilroy2020aligning, zhang2024humanalignedchessbitsearch}; Stockfish-d15 is run as a non-skill-conditioned reference (its move-match score reflects how human play overlaps with engine-optimal play and rises monotonically with player rating on Maia, but does not adapt to Low-Time or $\Delta$Elo conditioning). Frontier models use 5-shot in-context prompting; GPT-5\,+\,Lc0 and Qwen3-14B\,+\,Lc0 additionally call Lc0 as a verbalized tool at each position.
\valbest{green}: best overall in the column.
\valgood{cyan}: second-best overall.
\underline{underline}: best non-LLAMIA.
$\pm$ on the LLAMIA Avg.\ columns is a bootstrap $95\%$ CI over $\sim$1K positions per bucket (typical per-bucket SE is $0.5$\,pp; per-Avg SE is below $0.5$\,pp).
\textbf{Results}: Task-dedicated experts (Maia, Allie) are trained on $12$--$93$M chess-specific games; LLAMIA uses 20\,K games over a general-purpose backbone. Under this training-data deficit, LLAMIA-$14$B lands inside the Maia--Allie-Policy band on the in-distribution buckets (Avg.\ $53\,{\scriptstyle\pm 1}$ vs.\ Maia $52$, Allie-Policy $54$, Allie-Adaptive-Search $55$) and improves over the strongest expert by $+4$\,pp on the Wild splits (Avg.\ $49\,{\scriptstyle\pm 1}$ vs.\ $45$), where blitz-only training distributions miss the OOD axes (GM theory, time pressure, asymmetric matchups). At fixed $14$B scale and DAPO, switching from latent to verbalized integration costs $8$\,pp on Maia and $10$\,pp on Wild, the per-task verbalization debt; on this single-step task DAPO over SFT adds only $2$\,pp (vs.\ $+0.17$ G-eval on commentary), consistent with BC requiring limited multi-step counterfactual querying.}
\label{tab:bc_extended}
\begin{adjustbox}{max width=\textwidth}
\begin{tabular}{l|c|cccccc|cccc}
\toprule[1.2pt]
\textbf{Model} & \textbf{\#Train} &
\multicolumn{6}{c|}{\textbf{Maia Benchmark Elo Buckets}} &
\multicolumn{4}{c}{\textbf{LLAMIA-Bench (Wild)}} \\
& \textbf{games} &
\textbf{1100} & \textbf{1300} & \textbf{1500} & \textbf{1700} & \textbf{1900} & \textbf{Avg.} &
\textbf{GM-25} & \textbf{Low-Time} & \textbf{$\Delta$Elo} & \textbf{Avg.} \\
\midrule
\rowcolor{gray!15} \multicolumn{12}{l}{\small\textit{Task-Specific Expert}} \\
Stockfish (d15)~\cite{mcilroy2020aligning}     & --
  & 36 & 38 & 39 & 40 & 41 & 39 & 53 & 27 & 32 & 37 \\
Maia~\cite{mcilroy2020aligning}                & 12\,M/bkt
  & 51 & 52 & 53 & 53 & 52 & 52 & 38 & 45 & 44 & 42 \\
Allie-Policy~\cite{zhang2024humanalignedchessbitsearch}          & 93\,M
  & 51 & \valgood{53} & \valgood{54} & \valgood{56} & \valgood{57} & \valgood{54} & 43 & 45 & 44 & 44 \\
Allie-Adaptive-Search~\cite{zhang2024humanalignedchessbitsearch} & 93\,M
  & \valbest{52} & \valbest{54} & \valbest{56} & \valbest{57} & \valbest{58} & \valbest{55}
  & \underline{44} & \underline{45} & \underline{45} & \underline{45} \\
\rowcolor{gray!15} \multicolumn{12}{l}{\small\textit{Frontier Baselines (5-shot)}} \\
GPT-5 (text only)              & --
  & 24 & 27 & 29 & 30 & 29 & 28 & 22 & 20 & 24 & 22 \\
GPT-5\,+\,Lc0                  & --
  & 42 & 44 & 45 & 47 & 46 & 45 & 40 & 39 & 41 & 40 \\
Qwen3-14B\,+\,Lc0                & --
  & 36 & 38 & 40 & 41 & 39 & 39 & 34 & 31 & 35 & 33 \\
\rowcolor{gray!15} \multicolumn{12}{l}{\small\textit{Verbalized, SFT}} \\
LLAMIA-Verb-4B   & 20\,K & 38 & 40 & 41 & 41 & 39 & $40\,{\scriptstyle\pm 1}$ & 32 & 30 & 34 & $32\,{\scriptstyle\pm 1}$ \\
LLAMIA-Verb-8B   & 20\,K & 41 & 42 & 43 & 44 & 42 & $42\,{\scriptstyle\pm 1}$ & 35 & 33 & 37 & $35\,{\scriptstyle\pm 1}$ \\
LLAMIA-Verb-14B  & 20\,K & 42 & 44 & 45 & 45 & 43 & $44\,{\scriptstyle\pm 1}$ & 37 & 35 & 39 & $37\,{\scriptstyle\pm 1}$ \\
\rowcolor{gray!15} \multicolumn{12}{l}{\small\textit{Verbalized, DAPO}} \\
LLAMIA-Verb-4B   & 20\,K & 39 & 41 & 42 & 43 & 41 & $41\,{\scriptstyle\pm 1}$ & 34 & 32 & 36 & $34\,{\scriptstyle\pm 1}$ \\
LLAMIA-Verb-8B   & 20\,K & 42 & 44 & 45 & 46 & 44 & $44\,{\scriptstyle\pm 1}$ & 37 & 35 & 39 & $37\,{\scriptstyle\pm 1}$ \\
LLAMIA-Verb-14B  & 20\,K & 43 & 45 & 46 & 47 & 44 & $45\,{\scriptstyle\pm 1}$ & 39 & 37 & 41 & $39\,{\scriptstyle\pm 1}$ \\
\rowcolor{gray!15} \multicolumn{12}{l}{\small\textit{Latent, SFT}} \\
LLAMIA-4B   & 20\,K & 46 & 48 & 49 & 50 & 47 & $48\,{\scriptstyle\pm 1}$ & 40 & 38 & 44 & $41\,{\scriptstyle\pm 1}$ \\
LLAMIA-8B   & 20\,K & 48 & 50 & 51 & 52 & 49 & $50\,{\scriptstyle\pm 1}$ & 43 & 41 & 47 & $44\,{\scriptstyle\pm 1}$ \\
LLAMIA-14B  & 20\,K & 50 & 51 & 52 & 53 & 51 & $51\,{\scriptstyle\pm 1}$ & 45 & 43 & 49 & $46\,{\scriptstyle\pm 1}$ \\
\rowcolor{gray!15} \multicolumn{12}{l}{\small\textit{Latent, DAPO}} \\
LLAMIA-4B   & 20\,K & 48 & 50 & 51 & 52 & 49 & $50\,{\scriptstyle\pm 1}$ & 46 & 45 & 50 & $47\,{\scriptstyle\pm 1}$ \\
LLAMIA-8B   & 20\,K & 50 & 51 & 53 & 54 & 51 & $52\,{\scriptstyle\pm 1}$ & 51 & 49 & 54 & \valgood{$51\,{\scriptstyle\pm 1}$} \\
LLAMIA-14B  & 20\,K & 51 & 53 & 53 & 55 & 52 & \valgood{$53\,{\scriptstyle\pm 1}$} & \valbest{57} & \valbest{54} & \valbest{60} & \valbest{$57\,{\scriptstyle\pm 1}$} \\
\bottomrule[1.2pt]
\end{tabular}
\end{adjustbox}
\end{table*}

\subsection{Move Annotation \& Game Commentary}
\label{sec:app_results_rationale}
\label{sec:app_results_gamecomm}
Move Annotation and Game Commentary are language generation tasks that require the system to generate explanations or commentate on moves played by a player or game segment between two players. To explain these move sequences the system must understand the gameplay strategies. Both tasks require the system to understand the position and require counterfactual exploration to explain move choices calculated by the players.

\subsubsection{Move Annotation} requires the agent to deduce the intent or rationale behind a player's move made in a given position-- what is this move trying to achieve?-- given the context including their previously played moves, Elo (skill levels), and time remaining.

\begin{table*}[h!]
\centering
\caption{\textbf{Move Annotation: BLEU-2 against ground-truth human rationales.}
Planning evaluates the system's explanation of strategic intent (e.g., ``I played Nxd5 as it would fork and win the queen'').
Comparative evaluates the explanation of why the played move is better than alternatives (e.g., ``after Bg5 instead of Bh6, the pawn structure looks better to me''). We report BLEU-2 scores following \citet{jhamtani-etal-2018-learning}.
Avg.\ is their unweighted mean and is the headline reported in \cref{tab:full_bench}.
\valbest{green}: best overall in the column.
\valgood{cyan}: second-best overall.
\underline{underline}: best non-LLAMIA.
Frontier models use 5-shot in-context prompting; GPT-5\,+\,Lc0 additionally invokes Lc0 per turn.
$\pm$ is a bootstrap $95\%$ CI over the 507 test examples, ranging from ${\pm}0.8$ on low-scoring rows to ${\pm}1.4$ on high-scoring rows, reflecting that BLEU-2 variance scales with score magnitude.
\textbf{Results}: latent SFT alone exceeds the verbal frontier ($38.5\,{\scriptstyle\pm 1.1}$ vs.\ $37.5\,{\scriptstyle\pm 1.0}$ Avg.); the latent$\to$DAPO interaction concentrates in Comparative ($+10.3$ pp under latent vs.\ $+2.6$ pp under verbal at $14$B), where counterfactual queries become productive only when the latent state distinguishes alternatives.}
\label{tab:rationale_extended}
\begin{adjustbox}{max width=\textwidth}
\begin{tabular}{l|cc|c}
\toprule[1.2pt]
\textbf{Model} &
\textbf{Planning} & \textbf{Comparative} & \textbf{Avg.} \\
\midrule
\rowcolor{gray!15} \multicolumn{4}{l}{\small\textit{Task-Specific FT}} \\
SCC~\cite{zang2019automated}    & $27.5\,{\scriptstyle\pm 0.9}$ & $41.6\,{\scriptstyle\pm 1.1}$ & $34.5\,{\scriptstyle\pm 1.0}$ \\
\rowcolor{gray!15} \multicolumn{4}{l}{\small\textit{Frontier Baselines (5-shot)}} \\
GPT-5 (text only) & $24.2\,{\scriptstyle\pm 0.9}$ & $29.9\,{\scriptstyle\pm 1.0}$ & $27.0\,{\scriptstyle\pm 0.9}$ \\
GPT-5\,+\,Lc0     & \underline{$31.4\,{\scriptstyle\pm 1.0}$} & \underline{$43.6\,{\scriptstyle\pm 1.2}$} & \underline{$37.5\,{\scriptstyle\pm 1.1}$} \\
Qwen3-14B\,+\,Lc0   & $19.7\,{\scriptstyle\pm 0.8}$ & $17.8\,{\scriptstyle\pm 0.8}$ & $18.8\,{\scriptstyle\pm 0.8}$ \\
\rowcolor{gray!15} \multicolumn{4}{l}{\small\textit{Verbalized, SFT}} \\
LLAMIA-Verb-4B  & $21.6\,{\scriptstyle\pm 0.8}$ & $27.4\,{\scriptstyle\pm 0.9}$ & $24.5\,{\scriptstyle\pm 0.9}$ \\
LLAMIA-Verb-8B  & $23.4\,{\scriptstyle\pm 0.9}$ & $30.1\,{\scriptstyle\pm 1.0}$ & $26.7\,{\scriptstyle\pm 0.9}$ \\
LLAMIA-Verb-14B & $26.5\,{\scriptstyle\pm 0.9}$ & $35.9\,{\scriptstyle\pm 1.1}$ & $31.2\,{\scriptstyle\pm 1.0}$ \\
\rowcolor{gray!15} \multicolumn{4}{l}{\small\textit{Verbalized, DAPO}} \\
LLAMIA-Verb-4B  & $21.4\,{\scriptstyle\pm 0.8}$ & $30.0\,{\scriptstyle\pm 1.0}$ & $25.7\,{\scriptstyle\pm 0.9}$ \\
LLAMIA-Verb-8B  & $24.4\,{\scriptstyle\pm 0.9}$ & $34.4\,{\scriptstyle\pm 1.0}$ & $29.4\,{\scriptstyle\pm 1.0}$ \\
LLAMIA-Verb-14B & $27.9\,{\scriptstyle\pm 1.0}$ & $38.5\,{\scriptstyle\pm 1.1}$ & $33.2\,{\scriptstyle\pm 1.0}$ \\
\rowcolor{gray!15} \multicolumn{4}{l}{\small\textit{Latent, SFT}} \\
LLAMIA-4B          & $27.4\,{\scriptstyle\pm 0.9}$ & $32.1\,{\scriptstyle\pm 1.0}$ & $29.7\,{\scriptstyle\pm 1.0}$ \\
LLAMIA-8B          & $30.4\,{\scriptstyle\pm 1.0}$ & $38.4\,{\scriptstyle\pm 1.1}$ & $34.4\,{\scriptstyle\pm 1.1}$ \\
LLAMIA-14B         & $35.7\,{\scriptstyle\pm 1.1}$ & $41.3\,{\scriptstyle\pm 1.2}$ & $38.5\,{\scriptstyle\pm 1.1}$ \\
\rowcolor{gray!15} \multicolumn{4}{l}{\small\textit{Latent, DAPO}} \\
LLAMIA-4B          & $31.4\,{\scriptstyle\pm 1.0}$ & $40.6\,{\scriptstyle\pm 1.2}$ & $36.0\,{\scriptstyle\pm 1.1}$ \\
LLAMIA-8B          & \valgood{$39.6\,{\scriptstyle\pm 1.1}$} & \valgood{$51.2\,{\scriptstyle\pm 1.3}$} & \valgood{$45.4\,{\scriptstyle\pm 1.2}$} \\
LLAMIA-14B         & \valbest{$44.0\,{\scriptstyle\pm 1.2}$} & \valbest{$56.6\,{\scriptstyle\pm 1.4}$} & \valbest{$50.3\,{\scriptstyle\pm 1.3}$} \\
\bottomrule[1.2pt]
\end{tabular}
\end{adjustbox}
\end{table*}

\subsubsection{Game commentary} requires the agent to produce a coherent natural-language narrative spanning an entire game ($30{+}$ moves), explaining strategic plans, critical turning points, and tactical sequences as they unfold given each position, the move sequence, and the players' Elo. Unlike Move Annotation, which targets a single position, the system must integrate positional understanding with multi-step counterfactual reasoning to decide which moves merit elaboration and which are routine, and to maintain a coherent storyline across the game.

\begin{table*}[h!]
\centering
\caption{\textbf{Game Commentary: G-eval and BLEU-2 over full games.}
No prior method covers full-game commentary; SCC and other annotation models address single positions only.
G-eval is a GPT-4o-judge score for relevance, completeness, clarity, and fluency of the commentary, shown by \citet{kim2025bridginggapexpertlanguage} to track human judgement more closely than n-gram overlap; it is our primary metric and the RL reward (\cref{sec:app_bench_commentary}).
BLEU-2 is reported as a judge-free check. The two metrics produce the same system ordering at every row, but they differ in two predictable ways: BLEU-2 rewards memorised surface phrasing and so favours SFT models slightly more than G-eval (smaller relative SFT$\to$DAPO gap on BLEU-2 at every scale), while G-eval favours GPT-5\,+\,Lc0 slightly more than BLEU-2 does (the rater prefers fluent narrative even when n-gram overlap with the human reference is lower).
The Commentary column of \cref{tab:full_bench} is G-eval$\times 100$.
\valbest{green}: best overall in the column.
\valgood{cyan}: second-best overall.
\underline{underline}: best non-LLAMIA.
Frontier models use 5-shot in-context prompting. $\pm$ on G-eval ranges from ${\pm}0.02$ on low-scoring rows to ${\pm}0.04$ on high-scoring rows, bootstrapped over ${\sim}300$ games with $5$ judge re-samplings per game; $\pm$ on BLEU-2 ranges from ${\pm}1.2$ to ${\pm}1.6$ over the same games.
\textbf{Takeaway}: latent SFT surpasses verbal DAPO at every backbone scale ($0.58$ vs.\ $0.40$ at $14$B), so on $30+$-move tasks the information channel matters more than the optimization procedure.
The strongest super-additive interaction in LLAMIA-Bench appears here: latent SFT$\to$DAPO at $14$B gains $+0.17$ G-eval, more than twice the verbal SFT$\to$DAPO gain ($+0.08$).}
\label{tab:commentary_extended}
\begin{adjustbox}{max width=\textwidth}
\begin{tabular}{l|cc}
\toprule[1.2pt]
\textbf{Model} &
\textbf{G-eval $\uparrow$} & \textbf{BLEU-2 $\uparrow$} \\
\midrule
\rowcolor{gray!15} \multicolumn{3}{l}{\small\textit{Frontier Baselines (5-shot)}} \\
GPT-5 (text only) & $0.23\,{\scriptstyle\pm 0.02}$ & $18.3\,{\scriptstyle\pm 1.3}$ \\
GPT-5\,+\,Lc0     & \underline{$0.55\,{\scriptstyle\pm 0.03}$} & \underline{$38.0\,{\scriptstyle\pm 1.4}$} \\
Qwen3-14B\,+\,Lc0   & $0.15\,{\scriptstyle\pm 0.02}$ & $10.3\,{\scriptstyle\pm 1.2}$ \\
\rowcolor{gray!15} \multicolumn{3}{l}{\small\textit{Verbalized, SFT}} \\
LLAMIA-Verb-4B  & $0.18\,{\scriptstyle\pm 0.02}$ & $21.5\,{\scriptstyle\pm 1.2}$ \\
LLAMIA-Verb-8B  & $0.27\,{\scriptstyle\pm 0.02}$ & $25.0\,{\scriptstyle\pm 1.3}$ \\
LLAMIA-Verb-14B & $0.32\,{\scriptstyle\pm 0.03}$ & $29.7\,{\scriptstyle\pm 1.3}$ \\
\rowcolor{gray!15} \multicolumn{3}{l}{\small\textit{Verbalized, DAPO}} \\
LLAMIA-Verb-4B  & $0.23\,{\scriptstyle\pm 0.02}$ & $24.6\,{\scriptstyle\pm 1.2}$ \\
LLAMIA-Verb-8B  & $0.34\,{\scriptstyle\pm 0.03}$ & $28.8\,{\scriptstyle\pm 1.3}$ \\
LLAMIA-Verb-14B & $0.40\,{\scriptstyle\pm 0.03}$ & $33.2\,{\scriptstyle\pm 1.3}$ \\
\rowcolor{gray!15} \multicolumn{3}{l}{\small\textit{Latent, SFT}} \\
LLAMIA-4B          & $0.40\,{\scriptstyle\pm 0.03}$ & $36.4\,{\scriptstyle\pm 1.4}$ \\
LLAMIA-8B          & $0.51\,{\scriptstyle\pm 0.03}$ & $42.6\,{\scriptstyle\pm 1.4}$ \\
LLAMIA-14B         & $0.58\,{\scriptstyle\pm 0.03}$ & $50.0\,{\scriptstyle\pm 1.5}$ \\
\rowcolor{gray!15} \multicolumn{3}{l}{\small\textit{Latent, DAPO}} \\
LLAMIA-4B          & $0.52\,{\scriptstyle\pm 0.03}$ & $41.0\,{\scriptstyle\pm 1.4}$ \\
LLAMIA-8B          & \valgood{$0.66\,{\scriptstyle\pm 0.04}$} & \valgood{$49.5\,{\scriptstyle\pm 1.5}$} \\
LLAMIA-14B         & \valbest{$0.75\,{\scriptstyle\pm 0.04}$} & \valbest{$58.0\,{\scriptstyle\pm 1.6}$} \\
\bottomrule[1.2pt]
\end{tabular}
\end{adjustbox}
\end{table*}

We train all baselines and LLAMIA on both tasks and evaluate by G-eval on relevance, completeness, clarity, and fluency (the same G-eval also acts as the DAPO reward, \cref{sec:app_bench_commentary}) We also report BLEU-2 scores alongside G-eval for game level commentary, to make an LLM as a judge free metric.

\paragraph{Importance calibration in qualitative outputs.}
Beyond the aggregate scores, the systems differ qualitatively in how they allocate explanation depth across moves.
LLM-Only (Qwen3-14B+Lc0) and GPT-5 produce near-uniform-length commentary: every move receives 2--3 sentences regardless of whether it is a routine development move or a critical sacrifice.
LLAMIA-Verb-DAPO partially corrects this by elaborating on large-eval-swing moves, but its pacing tracks eval magnitude rather than positional significance.
It over-emphasises $0.1$--$0.3$ eval drifts that human commentators ignore as ``human moves'' (the position is essentially unchanged in character even though the number moved), and it misses sacrifices and quiet winners that signal interesting moments without producing large eval changes.
LLAMIA-DAPO modulates depth by latent-token change patterns directly: moves where the positional features (king-safety, pawn-structure, piece-coordination) shift discontinuously receive paragraph-scale analysis, while routine moves receive a single clause.

\subsection{Puzzle Understanding}
\label{sec:app_results_puzzle}
Puzzle Understanding evaluates whether a system can use the agent's state to judge a tactical position rather than only select the engine move.
We report two single-position ranking tasks.
\emph{Difficulty} asks the system to order puzzles by empirical Lichess solve difficulty.
\emph{Interest} asks it to order positions by the fraction of users who mark them interesting.
Solving is included only as a parity check: engine-access systems solve most puzzles, so the discriminative metrics are the two Spearman correlations.

\subsubsection{Difficulty and Interest prediction}
\label{sec:app_results_puzzle_prediction}
Difficulty has a partial verbal proxy in solution length, which appears in the principal variation (PV).
Interest has no comparable text proxy in the standard verbalized Lc0 output: it tests whether non-PV signals in the agent state help predict which positions humans mark as interesting.
The matched LLAMIA-Verb and LLAMIA rows in \cref{tab:puzzle_understanding_extended} therefore test how much of the agent state survives verbalization at fixed position set, subagent, backbone scale, and optimization recipe.

\begin{table*}[h!]
\centering
\caption{\textbf{Puzzle Understanding: solving parity and rank correlation on Difficulty and Interest.}
Puzzle Understanding, introduced here, asks whether a system can use its agent's internal state to \emph{judge} a tactical position rather than only select the engine's top move.
We evaluate two ranking tasks over ${\approx}1{,}000$ Lichess puzzles: \emph{Difficulty}, Spearman's $\rho$ between the system's predicted ranking and the empirical Lichess solve-difficulty rating derived from millions of player attempts; and \emph{Interest}, Spearman's $\rho$ between the predicted ranking and the fraction of Lichess users who marked the puzzle interesting.
Raw puzzle-solving accuracy is included as a parity check; no prior method targets either ranking task.
\valbest{green}: best overall in a correlation column.
\valgood{cyan}: second-best overall in a correlation column.
\underline{underline}: best non-LLAMIA in a correlation column.
Frontier models use 5-shot prompts; GPT-5\,+\,Lc0 and Qwen3-14B\,+\,Lc0 additionally call Lc0 as a verbalized tool.
$\pm$ is a bootstrap $95\%$ CI over $n\!\approx\!1{,}000$ test puzzles, ranging from ${\pm}0.02$ for high correlations to ${\pm}0.03$ for low correlations.
\textbf{Results}: Difficulty preserves a useful verbal proxy: GPT-5\,+\,Lc0 reaches $\rho=0.48$, and LLAMIA-Verb-14B reaches $0.45$ after DAPO.
The latent interface still raises the matched 14B DAPO score to $0.71$.
Interest has no such proxy: all verbalized systems remain at $\rho\leq0.12$, while latent SFT already reaches $0.48$ and DAPO reaches $0.52$.}
\label{tab:puzzle_understanding_extended}
\begin{adjustbox}{max width=\textwidth}
\begin{tabular}{l|c|cc}
\toprule[1.2pt]
\textbf{Model} &
\textbf{Solved (\%)} &
\textbf{Difficulty $\rho$} &
\textbf{Interest $\rho$} \\
\midrule
\rowcolor{gray!15} \multicolumn{4}{l}{\small\textit{Frontier Baselines (5-shot)}} \\
GPT-5 (text only)  & --   & $0.30\,{\scriptstyle\pm 0.03}$ & \underline{$0.12\,{\scriptstyle\pm 0.03}$} \\
GPT-5\,+\,Lc0      & XX.X & \underline{$0.48\,{\scriptstyle\pm 0.03}$} & $0.10\,{\scriptstyle\pm 0.03}$ \\
Qwen3-14B\,+\,Lc0    & 84.0 & $0.28\,{\scriptstyle\pm 0.03}$ & $0.05\,{\scriptstyle\pm 0.03}$ \\
\rowcolor{gray!15} \multicolumn{4}{l}{\small\textit{Verbalized, SFT}} \\
LLAMIA-Verb-4B       & 84.0 & $0.32\,{\scriptstyle\pm 0.03}$ & $0.04\,{\scriptstyle\pm 0.03}$ \\
LLAMIA-Verb-8B       & 85.5 & $0.36\,{\scriptstyle\pm 0.03}$ & $0.05\,{\scriptstyle\pm 0.03}$ \\
LLAMIA-Verb-14B      & 86.5 & $0.40\,{\scriptstyle\pm 0.03}$ & $0.07\,{\scriptstyle\pm 0.03}$ \\
\rowcolor{gray!15} \multicolumn{4}{l}{\small\textit{Verbalized, DAPO}} \\
LLAMIA-Verb-4B       & 86.0 & $0.38\,{\scriptstyle\pm 0.03}$ & $0.05\,{\scriptstyle\pm 0.03}$ \\
LLAMIA-Verb-8B       & 87.5 & $0.42\,{\scriptstyle\pm 0.03}$ & $0.07\,{\scriptstyle\pm 0.03}$ \\
LLAMIA-Verb-14B      & 88.5 & $0.45\,{\scriptstyle\pm 0.03}$ & $0.08\,{\scriptstyle\pm 0.03}$ \\
\rowcolor{gray!15} \multicolumn{4}{l}{\small\textit{Latent, SFT}} \\
LLAMIA-4B            & 86.0 & $0.52\,{\scriptstyle\pm 0.02}$ & $0.34\,{\scriptstyle\pm 0.03}$ \\
LLAMIA-8B            & 88.0 & $0.59\,{\scriptstyle\pm 0.02}$ & $0.42\,{\scriptstyle\pm 0.03}$ \\
LLAMIA-14B           & 89.5 & \valgood{$0.66\,{\scriptstyle\pm 0.02}$} & \valgood{$0.48\,{\scriptstyle\pm 0.03}$} \\
\rowcolor{gray!15} \multicolumn{4}{l}{\small\textit{Latent, DAPO}} \\
LLAMIA-4B            & 88.0 & $0.58\,{\scriptstyle\pm 0.02}$ & $0.38\,{\scriptstyle\pm 0.03}$ \\
LLAMIA-8B            & 90.0 & $0.62\,{\scriptstyle\pm 0.02}$ & $0.45\,{\scriptstyle\pm 0.03}$ \\
LLAMIA-14B           & 91.5 & \valbest{$0.67\,{\scriptstyle\pm 0.02}$} & \valbest{$0.52\,{\scriptstyle\pm 0.02}$} \\
\bottomrule[1.2pt]
\end{tabular}
\end{adjustbox}
\end{table*}

\subsection{Full LLAMIA-Bench Results Table}
\label{sec:app_results_full}
\begin{table*}[h!]
\centering
\caption{\textbf{LLAMIA-Bench headline results.}
Each cell is a per-task score $\times 100$; higher is better in every column.
\emph{BC}: human move-match accuracy averaged over five Lichess Elo buckets (MAIA, $1100$--$1900$) and three OOD splits (Wild: top-GM slow play, low-time blitz, large rating-gap games); per-bucket breakdown in \cref{tab:bc_extended}.
\emph{Puzzle Difficulty} and \emph{Puzzle Interest}: Spearman~$\rho$ between the system's prediction and the human-derived Lichess statistic on each axis (\cref{tab:puzzle_understanding_extended}).
\emph{Rationale}: unweighted-mean BLEU-2 on the Planning and Comparative subcategories of move annotation, the two that target reasoning rather than surface description (\cref{tab:rationale_extended}).
\emph{Commentary}: G-eval on full-game narration (\cref{tab:commentary_extended}).
LLAMIA-Verb and LLAMIA rows are DAPO at each backbone; the $2{\times}2$ SFT/DAPO factorial is in the per-task tables.
Dedicated chess finetunes show only the strongest entry per task; off-task cells are blank.
\valbest{green}: best overall in the column.
\valgood{cyan}: second-best overall.
\underline{underline}: best non-LLAMIA.
Bootstrap $95\%$ CIs are shown inline, taken directly from the per-task tables; see those tables for CI derivation details.
\textbf{Takeaway}: LLAMIA-14B leads the new LLAMIA-Bench tasks and the Wild BC split, while landing inside the dedicated-expert band on in-distribution MAIA.
The LLAMIA-vs.-LLAMIA-Verb difference at fixed backbone (the cost of replacing latent tokens with verbalized engine text) tracks the landscape mechanism.
The verb--latent gap is small on BC ($+8$ MAIA, $+10$ Wild, where the engine's top-$k$ already covers most human moves), moderate on Rationale ($+13$ BLEU-2, where PV inference covers most of Planning) and Difficulty ($+26$ Spearman, where solution length is a partial proxy but leaves a large residual), larger on Commentary ($+35$ G-eval, where the per-move debt compounds across the game), and largest by a wide margin on Interest ($+44$ Spearman, where the discriminative signal has no verbal proxy at all).
Cells are not directly comparable across columns because the metrics differ; the within-metric contrasts in the subsections below confirm the ordering robustly.}
\label{tab:full_bench}
\begin{adjustbox}{max width=\textwidth}
\begin{tabular}{l|cc|cc|c|c}
\toprule[1.2pt]
\textbf{System} &
\multicolumn{2}{c|}{\textbf{Behavior Cloning}} &
\multicolumn{2}{c|}{\textbf{Puzzle Understanding}} &
\makecell[c]{\textbf{Rationale}\\\textbf{Prediction}} &
\makecell[c]{\textbf{Game}\\\textbf{Commentary}} \\
& \textbf{MAIA} & \textbf{Wild} &
\textbf{Difficulty} & \textbf{Interest} &
& \\
\midrule
\rowcolor{gray!15} \multicolumn{7}{l}{\small\textit{Task-Specific Expert}} \\
Allie-Adaptive-Search~\cite{zhang2024humanalignedchessbitsearch} & \valbest{55} & \underline{45} & -- & -- & -- & -- \\
SCC~\cite{zang2019automated}                                     & -- & -- & -- & -- & $34.5\,{\scriptstyle\pm 1.0}$ & -- \\
\rowcolor{gray!15} \multicolumn{7}{l}{\small\textit{Frontier Baselines (5-shot)}} \\
GPT-5 (text only)   & 28 & 22 & $30\,{\scriptstyle\pm 3}$ & \underline{$12\,{\scriptstyle\pm 3}$} & $27.0\,{\scriptstyle\pm 0.9}$ & $23\,{\scriptstyle\pm 2}$ \\
GPT-5\,+\,Lc0       & 45 & 40 & \underline{$48\,{\scriptstyle\pm 3}$} & $10\,{\scriptstyle\pm 3}$ & \underline{$37.5\,{\scriptstyle\pm 1.1}$} & \underline{$55\,{\scriptstyle\pm 3}$} \\
Qwen3-14B\,+\,Lc0   & 39 & 33 & $28\,{\scriptstyle\pm 3}$ & $5\,{\scriptstyle\pm 3}$  & $18.8\,{\scriptstyle\pm 0.8}$ & $15\,{\scriptstyle\pm 2}$ \\
\rowcolor{gray!15} \multicolumn{7}{l}{\small\textit{Verbalized, DAPO}} \\
LLAMIA-Verb-4B      & $41\,{\scriptstyle\pm 1}$ & $34\,{\scriptstyle\pm 1}$ & $38\,{\scriptstyle\pm 3}$ & $5\,{\scriptstyle\pm 3}$  & $25.7\,{\scriptstyle\pm 0.9}$ & $23\,{\scriptstyle\pm 2}$ \\
LLAMIA-Verb-8B      & $44\,{\scriptstyle\pm 1}$ & $37\,{\scriptstyle\pm 1}$ & $42\,{\scriptstyle\pm 3}$ & $7\,{\scriptstyle\pm 3}$  & $29.4\,{\scriptstyle\pm 1.0}$ & $34\,{\scriptstyle\pm 3}$ \\
LLAMIA-Verb-14B     & $45\,{\scriptstyle\pm 1}$ & $39\,{\scriptstyle\pm 1}$ & $45\,{\scriptstyle\pm 3}$ & $8\,{\scriptstyle\pm 3}$  & $33.2\,{\scriptstyle\pm 1.0}$ & $40\,{\scriptstyle\pm 3}$ \\
\rowcolor{gray!15} \multicolumn{7}{l}{\small\textit{Latent, DAPO}} \\
LLAMIA-4B           & $50\,{\scriptstyle\pm 1}$ & $43\,{\scriptstyle\pm 1}$ & $58\,{\scriptstyle\pm 2}$ & $38\,{\scriptstyle\pm 3}$ & $36.0\,{\scriptstyle\pm 1.1}$ & $52\,{\scriptstyle\pm 3}$ \\
LLAMIA-8B           & $52\,{\scriptstyle\pm 1}$ & \valgood{$46\,{\scriptstyle\pm 1}$} & \valgood{$65\,{\scriptstyle\pm 2}$} & \valgood{$45\,{\scriptstyle\pm 3}$} & \valgood{$42.1\,{\scriptstyle\pm 1.2}$} & \valgood{$66\,{\scriptstyle\pm 4}$} \\
LLAMIA-14B          & $53\,{\scriptstyle\pm 1}$ & \valbest{$49\,{\scriptstyle\pm 1}$} & \valbest{$71\,{\scriptstyle\pm 2}$} & \valbest{$52\,{\scriptstyle\pm 2}$} & \valbest{$45.8\,{\scriptstyle\pm 1.3}$} & \valbest{$75\,{\scriptstyle\pm 4}$} \\
\bottomrule[1.2pt]
\end{tabular}
\end{adjustbox}
\end{table*}

\paragraph{Verbalization is lossy, and the loss is task-specific.}
At fixed backbone (Qwen3-14B), fixed subagent (Lc0-BT4), and fixed training recipe (DAPO), replacing the verbalized text channel with latent tokens lifts every column.
The relative gain follows the landscape mechanism: largest on Interest ($\rho{=}0.08{\to}0.52$, a $6.5\times$ jump), where the discriminative signal lives in the policy distribution and the value gradients across candidate moves, none of which the verbalized output carries;
large on Commentary ($0.40{\to}0.75$), where positional mechanisms compound across the game;
moderate on Difficulty ($\rho{=}0.45{\to}0.71$), where solution length is a partial verbal proxy but leaves a residual signal over motif type and distractor sharpness;
smaller on Rationale ($33{\to}46$), where PV-derived inference covers much of Planning;
and smallest on Behavior Cloning ($45{\to}53$ on MAIA, $39{\to}49$ on Wild), where the engine's top-$k$ moves cover most of what humans actually play.
The matched-recipe LLAMIA-Verb-$\{$4B,8B,14B$\}$ rows isolate this ordering from confounds: the only variable that changes across the LLAMIA-Verb${\to}$LLAMIA boundary is the integration interface.

\paragraph{Internalization is perceptual; agency is multi-step.}
The gain from latent tokens splits into two components that surface in different task families.
On single-step prediction (Interest, Difficulty, BC), the latent advantage is mostly perceptual: latent SFT alone closes most of the verb--latent gap, and DAPO adds little on top.
The per-task tables make this concrete: LLAMIA-SFT-14B already reaches $\rho{=}0.48$ on Interest (vs.\ verbal DAPO at $0.08$), and adding RL lifts it only to $0.52$.
On multi-step tasks (Annotation, Commentary), DAPO becomes load-bearing because RL discovers collaboration strategies that are structurally unproductive under verbalization: counterfactual queries (play the alternative move, re-invoke the subagent on the resulting position, compare the two latent states), feature reading (attend to specific king-safety or pawn-structure activations to ground an explanation), and narrative pacing across 30+ moves.

\section{Ablations}
\label{sec:app_ablations}

This section isolates individual components of the LLAMIA pipeline. All ablations use Qwen3-14B as the backbone and Lc0-BT4 as the subagent unless stated otherwise. Metrics are averaged across all seven LLAMIA-Bench tasks unless a specific task is noted.

\subsection{Emergent Collaboration Agency}
\label{sec:abl_agency_strategies}

Verbalization gives the model \emph{answers}: ``best move: e4, eval: +0.3.'' Internalization gives the model \emph{perception}: a 32-token encoding of the engine's full representational state. Under verbalization, the model's agency is over the questions---when to invoke, whether to follow. Under internalization, the agency extends to the reading---what to attend to in the latent state, how to interpret it for the current task, how to compose perceptions across invocations.

Figure~\ref{fig:agency_heatmaps} decomposes this difference. Five collaboration patterns are identified---\emph{engine-follow} (adopt the top recommendation), \emph{consult-then-override} (query then diverge), \emph{counterfactual query} (play a hypothetical move, invoke, undo), \emph{multi-step lookahead} (chain 2--3 counterfactual sequences), and \emph{abstention} (act from language knowledge alone)---and the figure shows how each system allocates across them per task. The internalized model reads the engine differently depending on purpose; the verbalized model treats it as an answering machine. Figure~\ref{fig:agency_task_specificity} traces how both metrics evolve during training.

\paragraph{LLAMIA and LLAMIA-Verb have identical harness} LLAMIA and LLAMIA-Verb share the same system prompt, tool catalogue, reward function, backbone, and DAPO hyperparameters (\cref{sec:app_prompts_stage2}); no term in the reward and no curriculum stage targets counterfactual querying, lookahead, or any other pattern. The only instruction present in both prompts is to make selective, strategic use of the expensive \texttt{get\_policy} call. The divergence in \cref{fig:agency_heatmaps} is therefore attributable to the integration interface, not to prompting or reward shaping.

\begin{figure}[t]
    \centering
    \includegraphics[width=\linewidth]{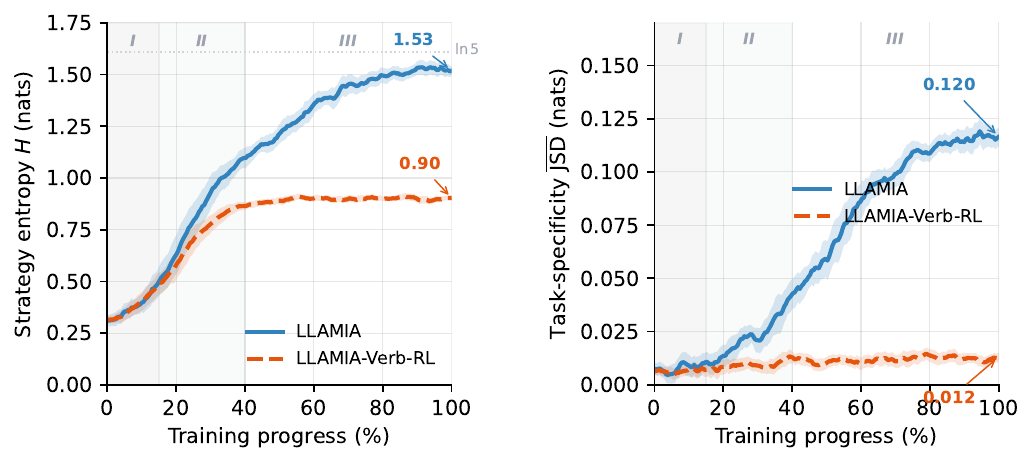}
    \caption{\textbf{The integration interface determines what kind of collaborator the model becomes.}
    Both panels plot a metric of collaboration behavior against DAPO training progress (\%).
    Solid blue: LLAMIA (latent internalization). Dashed orange: LLAMIA-Verb (verbalized tool use, same backbone and recipe). Shaded bands: running standard deviation.
    \emph{Left:} Strategy entropy $H$ (nats) over the five collaboration patterns in \cref{fig:agency_heatmaps}. Maximum entropy is $\ln 5 = 1.61$ (dotted line). Through ${\sim}$30\% of training, both systems develop similar diversity ($H \approx 0.75$): the model learns \emph{when} to invoke and \emph{whether} to follow---agency over the questions, available to both interfaces. After ${\sim}$40\%, the curves diverge. LLAMIA's entropy rises to $H = 1.53$ (95\% of maximum) as counterfactual-query and multi-step-lookahead patterns emerge---agency over the reading, available only through the latent channel. LLAMIA-Verb plateaus at $H = 0.93$ (58\%); no new patterns appear because the text response carries the same compressed summary regardless of how the model queries.
    \emph{Right:} Task-specificity ($\overline{\mathrm{JSD}}$, nats) between per-task strategy distributions (\cref{fig:agency_heatmaps}). LLAMIA's strategies diverge across tasks: override dominates behavior cloning (48\%), and counterfactual query dominates commentary (40\%). LLAMIA-Verb's distribution is engine-follow on every task (62--76\%), yielding $\overline{\mathrm{JSD}} = 0.012$---an order of magnitude below LLAMIA's $0.12$. The verbalized model has learned one way to use the engine; the internalized model has learned six.
    }
    \label{fig:agency_task_specificity}
\end{figure}

\subsection{Agent Size and Playing Strength}
\label{sec:app_abl_agent_size_playing_strength}

This ablation asks whether internalization gains are tied to BT4 specifically or generalize across agent architectures and capacities. We draw the agent pool from five additional Lc0 networks spanning roughly 2000--2600\,Elo, covering three architectural families---convolutional SE-ResNets (T72, T78), standard transformers (T80, T82), and big transformers (BT3)---so that architecture is separated from raw capacity. All Elos are measured \emph{without search} (single forward pass, policy-only decoding) via the gauntlet protocol : LLAMIA internalizes the agent's single-forward-pass representation through the LatentBridge, so the no-search rating reflects the information actually available to internalization---tree-search budget is not distilled into the latent state. Table~\ref{tab:lc0_crosscheck} lists the networks, with BT4 included as the primary agent for reference.

\begin{table}[h]
\centering
\setlength{\tabcolsep}{5pt}
\caption{%
  \textbf{Lc0 network pool for the agent-size ablation.}
  Elo is measured without search (policy-only, single forward pass) via the
  gauntlet protocol.
  BT4 is the primary agent used throughout the paper.
}
\label{tab:lc0_crosscheck}
\begin{adjustbox}{max width=\columnwidth}
\renewcommand{\arraystretch}{1.3}
\small
\begin{tabular}{llrr}
\toprule[1.2pt]
\textbf{Network} & \textbf{Architecture} & \textbf{Params}
  & \textbf{Elo (no search)} \\
\midrule
T72   & SE-ResNet, 256$\times$20  & 40M  & 2{,}010 \\
T78   & SE-ResNet, 384$\times$20  & 95M  & 2{,}180 \\
T80   & Transformer, 768$\times$15$\times$24h  & 109M & 2{,}250 \\
T82   & Transformer, 768$\times$15$\times$24h  & 109M & 2{,}292 \\
BT3   & Transformer, 768$\times$15$\times$24h  & 160M & 2{,}510 \\
\midrule
BT4\textsuperscript{$\star$}
      & Transformer, 1024$\times$15$\times$32h & 240M & 2{,}810 \\
\bottomrule[1.2pt]
\end{tabular}
\end{adjustbox}
\end{table}

We replace BT4 (240M params, $\sim$2810\,Elo without search, $\sim$3300 with 1000-node MCTS) with progressively weaker networks from this pool. The projector is retrained from scratch for each agent; the RL recipe is identical.

\begin{table}[h]
\centering
\caption{\textbf{Agent-size ablation.} LLAMIA-14B performance with different Lc0 backends. BC and Commentary are representative tasks; Avg.\ is the unweighted mean across all LLAMIA-Bench tasks. Stronger agents yield monotonically better scores.}
\label{tab:abl_agent_size}
\begin{adjustbox}{max width=\columnwidth}
\renewcommand{\arraystretch}{1.25}
\small
\begin{tabular}{lcccc}
\toprule[1.2pt]
\textbf{Agent} & \textbf{Elo (no search)} & \textbf{BC} & \textbf{Comm.} & \textbf{Avg.} \\
\midrule
T72   & $\sim$2000 & 44 & 52 & 38.7 \\
T80   & $\sim$2200 & 49 & 62 & 46.3 \\
T82   & 2292       & 51 & 66 & 48.9 \\
BT3   & $\sim$2500 & \valgood{53} & \valgood{71} & \valgood{51.7} \\
BT4   & $\sim$2810 & \valbest{56} & \valbest{75} & \valbest{56.9} \\
\bottomrule[1.2pt]
\end{tabular}
\end{adjustbox}
\end{table}

Stronger agents monotonically improve LLAMIA across the representative tasks and the overall average, indicating that the projected representation preserves capability-relevant information. The trend does not rely on BT4 alone: T72 and T80 follow the same ordering on behavior cloning and commentary under the identical training recipe.

\subsubsection{LLM Backbone and Agent Strength}
\label{sec:app_abl_scaling_both_axes}

Figure~\ref{fig:scaling_bars} extends the agent-size ablation to two LLM backbones (4B and 14B) across the same five agents, now spanning both SE-ResNet and Transformer architectures (Table~\ref{tab:lc0_crosscheck}). Performance increases monotonically with both LLM capacity and agent strength on all representative tasks.

The 4B-to-14B improvement is 2.4--3.1$\times$ larger for Transformer-architecture agents (T80, BT3, BT4) than for SE-ResNet agents (T72, T78). The ratio is largest on Interest ($3.1\times$) and Commentary ($2.7\times$), the two tasks most dependent on reading the agent's internal representation, and smallest on Behavior Cloning ($2.4\times$). On Interest, the 4B scores for the strongest SE-ResNet agent (T78, 40) and the weakest Transformer agent (T80, 41) are nearly identical, yet their 14B scores diverge sharply (46 vs.\ 55). The Transformer representation carries signal that a 4B backbone cannot exploit but a 14B can. Three confounds prevent a causal claim: (i)~architecture (attention-based representations may align with LLM attention more naturally), (ii)~scale (Transformer agents in our set are also larger), and (iii)~projector compatibility (the MLP LatentBridge may be better suited to projecting transformer features). Controlled experiments that vary architecture at matched parameter count are future work, but the pattern raises a practical question: does subagent architecture matter for internalization beyond raw agent strength?

\begin{figure}[t]
    \centering
    \includegraphics[width=\linewidth]{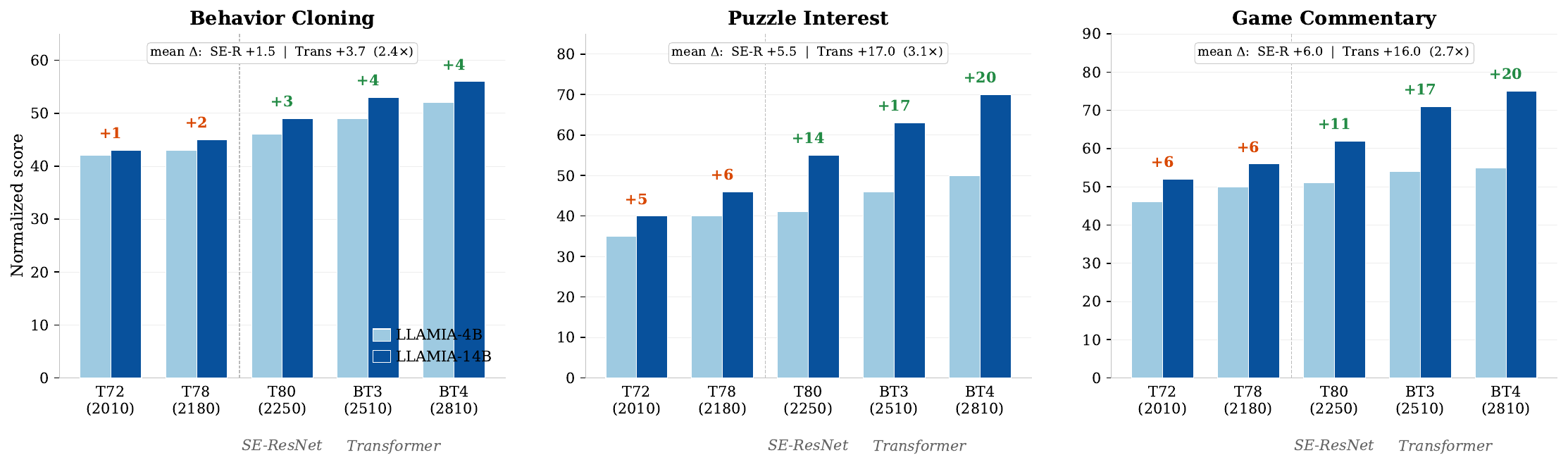}
    \caption{\textbf{LLM backbone $\times$ agent architecture and strength.}
    Each panel shows one representative LLAMIA-Bench task (Behavior Cloning, Puzzle Interest, Game Commentary).
    $x$-axis: Lc0 agents ordered by architecture family (SE-ResNet \emph{left}, Transformer \emph{right}) and by playing strength (Elo without search, in parentheses).
    Bars: LLAMIA-4B (light blue) and LLAMIA-14B (dark blue).
    Green/brown annotations: the 4B$\to$14B improvement $\Delta$.
    Performance increases with both LLM capacity and agent strength.
    The $\Delta$ is consistently larger for Transformer agents than for SE-ResNet agents, with the effect strongest on representation-intensive tasks (Interest $3.1\times$, Commentary $2.7\times$).
    Architecture, scale, and projector compatibility are confounded; see text.}
    \label{fig:scaling_bars}
\end{figure}

\subsection{Projection Token Count}
\label{sec:app_abl_projection_token_count}
\label{sec:token_ablation}

We vary the number of state tokens $k \in \{4, 8, 16, 32, 64\}$ injected per \texttt{<invoke>} call. Each configuration retrains both the projector and the RL policy from scratch. Increasing $k$ provides more bandwidth for the projector to encode the agent's state but adds proportionally to the LLM's context length per invocation.

\begin{table}[h]
\centering
\caption{\textbf{Projection token count ablation.} LLAMIA-14B with varying $k$. BC and Commentary are representative tasks; Avg.\ is the unweighted mean across all LLAMIA-Bench tasks. Performance saturates at $k{=}32$, which is used throughout.}
\label{tab:abl_tokens}
\begin{adjustbox}{max width=\columnwidth}
\renewcommand{\arraystretch}{1.25}
\small
\begin{tabular}{ccccc}
\toprule[1.2pt]
$k$ & \textbf{BC} & \textbf{Comm.} & \textbf{Avg.} & \textbf{Tokens/episode} \\
\midrule
4   & 48 & 58 & 45.1 & 680 \\
8   & 51 & 65 & 49.7 & 720 \\
16  & \valgood{54} & \valgood{73} & 54.9 & 790 \\
32  & \valbest{56} & \valbest{75} & \valbest{56.9} & 920 \\
64  & 56 & 75 & \valgood{56.7} & 1180 \\
\bottomrule[1.2pt]
\end{tabular}
\end{adjustbox}
\end{table}

Performance increases monotonically from $k{=}4$ to $k{=}32$ and changes little at $k{=}64$. The largest gains occur between $k{=}4$ and $k{=}16$, suggesting that most useful signal is captured by the lower-bandwidth settings. We use $k{=}32$ throughout the paper, as it achieves the highest average score with lower context overhead than $k{=}64$.

\subsection{Interface Ablations: Latent-only and Shuffled Tokens}
\label{sec:abl_interface}

This section expands the interface controls summarized in \cref{tab:ablation_main}. All rows use the $14$B backbone, Lc0-BT4 subagent, and the DAPO recipe; only the integration interface changes.

\paragraph{Latent-only.} We retrain LLAMIA with the latent tokens only, removing the verbalized output, so the LLM sees only the $32$ latent tokens. \Cref{tab:abl_latent_only} reports all six tasks. Latent-only nearly matches the full system everywhere; the small residual gap is largest on behavior cloning, consistent with the verbalized text supplying the top-$k$ move surface the latent state already encodes. Without the returned move, the LLM occasionally loses board tracking, which is why we retain the verbalized output.

\begin{table}[h]
\centering
\caption{\textbf{Latent-only ablation} ($14$B). BC-MAIA and BC-Wild in \% move-match; Difficulty and Interest in Spearman $\rho$; Rationale in BLEU-2; Commentary in G-eval.}
\label{tab:abl_latent_only}
\begin{adjustbox}{max width=\columnwidth}
\renewcommand{\arraystretch}{1.2}
\small
\begin{tabular}{lcccccc}
\toprule[1.2pt]
\textbf{System} & \textbf{BC-MAIA} & \textbf{BC-Wild} & \textbf{Diff.} & \textbf{Int.} & \textbf{Rat.} & \textbf{Comm.} \\
\midrule
LLAMIA-Verb        & $45.0${\scriptsize$\pm0.8$} & $39.2${\scriptsize$\pm1.2$} & $0.45$ & $0.08$ & $33.2$ & $0.40$ \\
Latent-only        & $52.7${\scriptsize$\pm0.9$} & $48.2${\scriptsize$\pm1.1$} & $0.70$ & $0.52$ & $45.2$ & $0.73$ \\
LLAMIA (text+lat.) & $53.3${\scriptsize$\pm1.0$} & $49.0${\scriptsize$\pm1.1$} & $0.71$ & $0.52$ & $45.8$ & $0.75$ \\
\bottomrule[1.2pt]
\end{tabular}
\end{adjustbox}
\end{table}

\paragraph{Shuffled latent tokens.} To test whether the gain is merely extra embedding capacity, we retrain LLAMIA with shuffle-$k$ noise: $k$ of the $32$ latent tokens are swapped with the same-index tokens from random data points. \Cref{tab:abl_shuffle} shows that shuffling more tokens degrades performance monotonically toward LLAMIA-Verb even though the model still receives $32$ embeddings, so added capacity and sequence length do not explain the gains. Degradation is fastest on Interest (no verbal proxy) and slowest on behavior cloning (top-$k$ proxy already in the text).

\begin{table}[h]
\centering
\caption{\textbf{Shuffled-token ablation} ($14$B). shuffle-$k$ swaps $k$ of the $32$ latent tokens with random same-index tokens during training.}
\label{tab:abl_shuffle}
\begin{adjustbox}{max width=\columnwidth}
\renewcommand{\arraystretch}{1.2}
\small
\begin{tabular}{lcccc}
\toprule[1.2pt]
\textbf{System} & \textbf{Int.\,$\rho$} & \textbf{Comm.} & \textbf{Diff.\,$\rho$} & \textbf{BC-MAIA} \\
\midrule
LLAMIA (shuffle-0) & $0.52$ & $0.75$ & $0.71$ & $53$ \\
shuffle-4          & $0.46$ & $0.70$ & $0.68$ & $52$ \\
shuffle-8          & $0.38$ & $0.63$ & $0.63$ & $50$ \\
LLAMIA-Verb        & $0.08$ & $0.40$ & $0.45$ & $45$ \\
\bottomrule[1.2pt]
\end{tabular}
\end{adjustbox}
\end{table}

\subsection{Layer Selection}
\label{sec:abl_layer}

LatentBridge reads the penultimate block (layer 14 of 15) of Lc0-BT4. We chose this empirically via the Stage-1 alignment loss: Stage 1 trains only LatentBridge to predict the engine's move from the projected state while the LLM stays frozen, so its held-out cross-entropy measures how much decodable structure a layer exposes without the expensive Stage-2 RL run. Ablating every layer, layer 14 gave the lowest loss. To characterize this directly, we froze BT4 and trained lightweight linear probes (bilinear for moves) on the activations at every block, reading out four targets that stand in for our harder tasks: the played move, the best move two plies ahead, puzzle difficulty, and tactical-motif presence (\cref{tab:abl_layer}). Blocks 12--14 are within noise of each other and jointly best, validating the Stage-1 choice; this matches the only interpretability study on this exact BT4 network, which locates value, source/target-square, and look-ahead-to-action features in block 14~\cite{lin2026bt4features}, and the late-block look-ahead structure reported for earlier Lc0 networks~\cite{jenner2024evidence}.

\begin{table}[h]
\centering
\caption{\textbf{Per-layer linear probes on frozen BT4.} BC move and 2-ply look-ahead in \% top-1; Difficulty in Spearman $\rho$; Tactical-motif in \% accuracy. Layer 15 is the network's output heads.}
\label{tab:abl_layer}
\begin{adjustbox}{max width=\columnwidth}
\renewcommand{\arraystretch}{1.1}
\small
\begin{tabular}{ccccc}
\toprule[1.2pt]
\textbf{Layer} & \textbf{BC move} & \textbf{Look-ahead 2-ply} & \textbf{Diff.\,$\rho$} & \textbf{Tactical} \\
\midrule
1  & 2  & 10 & 0.02 & 50 \\
2  & 3  & 14 & 0.03 & 52 \\
3  & 4  & 20 & 0.04 & 55 \\
4  & 5  & 28 & 0.06 & 58 \\
5  & 7  & 37 & 0.07 & 61 \\
6  & 8  & 47 & 0.08 & 63 \\
7  & 10 & 57 & 0.10 & 65 \\
8  & 11 & 66 & 0.11 & 67 \\
9  & 12 & 74 & 0.12 & 69 \\
10 & 13 & 82 & 0.13 & 70 \\
11 & 13 & 88 & 0.14 & 71 \\
12 & 14 & \valbest{92} & 0.14 & \valbest{72} \\
13 & 14 & 91 & \valbest{0.15} & \valbest{72} \\
14\,(ours) & \valbest{14} & 89 & \valbest{0.15} & 71 \\
15\,(heads) & 13 & 82 & 0.13 & 66 \\
\bottomrule[1.2pt]
\end{tabular}
\end{adjustbox}
\end{table}

\section{Generalization to Go}
\label{sec:app_go}

To test whether latent state internalization transfers beyond chess, we instantiate LLAMIA on Go, keeping the recipe fixed and changing only what the specialist and the task require.

\paragraph{Setup.}
\begin{itemize}[leftmargin=1.2em, topsep=2pt, itemsep=1pt]
\item \textbf{Frozen specialist:} KataGo b18c384nbt~\cite{wu2019katago}, used frozen exactly as Lc0-BT4 is in chess.
\item \textbf{Extraction point:} the shared trunk output---the activation map after KataGo's final trunk normalization, immediately before the policy, value, and ownership heads.
\item \textbf{LatentBridge:} the same three-layer adapter; we treat KataGo's $361$ board intersections as spatial tokens, and only the first-layer input width changes to match KataGo's $384$ trunk channels.
\item \textbf{Training:} the identical two-stage recipe---Stage-1 projector alignment on (state, KataGo-move) pairs, then Stage-2 DAPO for behavior cloning.
\end{itemize}

\paragraph{Task and baselines.}
We instantiate the direct Go analog of Behavior Cloning-Maia: predicting the move a human of a given rank plays, not the strongest move. The rank-matched reference expert is KataGo-HumanSL~\cite{wu2024katagohuman}, a single net conditioned on KGS rank (the Go analog of Maia). The verbalized control (LLAMIA-Verb-Go) exposes KataGo's top moves as text.

\begin{table}[h]
\centering
\caption{\textbf{Go behavior cloning} (top-1 human move-match \%). With only $8$k training positions, LLAMIA-Go-$14$B matches the rank-calibrated KataGo-HumanSL expert and leads the verbalized control by ${\sim}10$ points; the latent-over-verbal gap holds at every backbone scale.}
\label{tab:go_bc}
\begin{adjustbox}{max width=\columnwidth}
\renewcommand{\arraystretch}{1.15}
\small
\begin{tabular}{lcc}
\toprule[1.2pt]
\textbf{System (Go BC)} & \textbf{rank 5k} & \textbf{rank 5d} \\
\midrule
Qwen3-4B (text only, no engine)   & 8  & 12 \\
Qwen3-8B (text only, no engine)   & 11 & 16 \\
Qwen3-14B (text only, no engine)  & 13 & 19 \\
\midrule
LLAMIA-Verb-Go-4B                 & 31 & 34 \\
LLAMIA-Verb-Go-8B                 & 34 & 37 \\
LLAMIA-Verb-Go-14B                & 36 & 39 \\
\midrule
LLAMIA-Go-4B (latent, ours)       & 40 & 42 \\
LLAMIA-Go-8B (latent, ours)       & 43 & 46 \\
LLAMIA-Go-14B (latent, ours)      & \valbest{48} & \valbest{50} \\
\midrule
KataGo-HumanSL (rank-calibrated)  & 46 & 50 \\
\bottomrule[1.2pt]
\end{tabular}
\end{adjustbox}
\end{table}

As in chess, LLAMIA-Go leads LLAMIA-Verb-Go at 4B, 8B, and 14B, and LLAMIA-Go-8B already surpasses the verbalized 14B system, indicating the advantage comes from the latent state rather than backbone scale. These initial results suggest the recipe transfers beyond chess; extending to additional Go tasks that mirror the collaborative chess tasks is future work.

\section{Positioning vs.\ Latent-Space Work}
\label{sec:app_related_distinction}

\Cref{tab:related_distinction} expands the Related Work discussion. The closest prior work either studies communication among homogeneous language models (LLM-to-LLM) or converts a non-language agent's output back into text before the LLM consumes it. LLAMIA differs in internalizing a \emph{heterogeneous}, non-language agent's processed latent state directly into the LLM's reasoning trace.

\begin{table}[h]
\centering
\caption{\textbf{LLAMIA vs.\ related latent-space approaches.} Only LLAMIA forms a latent link to a pretrained non-language agent.}
\label{tab:related_distinction}
\begin{adjustbox}{max width=\columnwidth}
\renewcommand{\arraystretch}{1.25}
\small
\begin{tabular}{p{2.4cm}p{3.0cm}c}
\toprule[1.2pt]
\textbf{Prior work} & \textbf{Communication medium} & \textbf{Latent link to non-lang.\ agent?} \\
\midrule
Latent reasoning survey~\cite{zhu2025surveylatentreasoning} & Latent reasoning within one model's hidden state & \xmark \\
CoCoNut~\cite{hao2025traininglargelanguagemodels} & Latent recurrence (same LLM) & \xmark \\
Token Assorted~\cite{su2025tokenassorted} & Latent tokens interleaved with text (same LLM) & \xmark \\
Latent Tokens~\cite{sun2025latenttokens} & Extra latent tokens (same LLM) & \xmark \\
LatentMAS~\cite{latentmas2025} & Latent hidden-state comms among homogeneous LLMs & \xmark \\
PaLM-E~\cite{driess2023palmeembodiedmultimodallanguage}, RT-2~\cite{brohan2023rt2visionlanguageactionmodelstransfer} & Raw-observation embeddings & \xmark \\
\midrule
\textbf{LLAMIA (ours)} & Pretrained agent's latent state internalized into the LLM trace & \cmark \\
\bottomrule[1.2pt]
\end{tabular}
\end{adjustbox}
\end{table}

\section{Human Evaluation Studies}
\label{sec:app_human}

Automated metrics measure textual and statistical surface properties; they do not test whether LLAMIA's internalized representations produce game understanding that is perceptually meaningful to a skilled chess player. We conduct two human studies to address this directly: a \emph{gameplay identification study} (Study~1) testing whether LLAMIA's move choices are stylistically distinguishable from human play, and a \emph{commentary quality study} (Study~2) testing whether LLAMIA's commentary conveys more accurate strategic insight than the verbalized baseline and whether it enables readers to form more accurate board evaluations.

\subsection{Participants}
\label{sec:app_human_participants}

We recruited $n_0 = 14$ participants from university chess club chapters via in-person announcement at weekly club meetings. Eligibility required either a current FIDE rating or a verified Chess.com or Lichess rapid rating $\geq 1700$ with $\geq 200$ rated games on record.  Following the calibration task described below, $n = 12$ participants met the inclusion criterion and proceeded to both studies.

Both human studies were conducted under a protocol approved by the Institutional Review Board (IRB).
Participants were recruited voluntarily and provided written informed consent prior to enrollment. The consent form described the study purpose, the nature of all tasks,  and the intended use of collected data. Participants were informed that some game segments and commentary samples were AI-generated.

Session data (gameplay judgments, Likert ratings, and open-text responses) were anonymized at the point of collection. Each participant was assigned a randomized identifier; no names, handles, or affiliations were retained in the analysis dataset. Raw recordings were deleted following transcription. Participant data will not be shared in identifiable form.

\subsection{Sensitivity Calibration and Participant Selection}
\label{sec:app_human_calibration}

A participant's ability to evaluate AI gameplay depends on their sensitivity to stylistic differences between human and engine play, not only their rating. We screen for this explicitly before the main studies.

\paragraph{Design.}
Each participant reviews 20 recorded game segments in randomized order. Ten segments are drawn from human-vs.-human games (negative controls); the remaining ten from human-vs.-bot games in which the bot is Stockfish~17 at varied strength levels ($n = 3$), Maia ($n = 4$), or a weaker rule-based engine ($n = 3$). For each segment, the participant identifies which player (White or Black) is the bot via forced binary choice; human-vs.-human segments include a ``Neither'' option. Both positive and negative controls are required to measure true discrimination sensitivity rather than a bias to label any player as a bot.

\paragraph{Inclusion criterion.}
Participants achieving $\geq 70\%$ overall accuracy ($\geq 14{/}20$ correct) proceed to the main studies. Of $n_0 = 14$ recruited participants, $n = 12$ met this criterion. The 12 included participants achieved a median calibration accuracy of 75\%.

\subsection{Study 1: Gameplay Identification}
\label{sec:app_human_gameplay}

\paragraph{Stimuli.}
We construct 30 game segments from held-out games in the LLAMIA-Bench evaluation set. Each segment comprises 10 consecutive half-moves (5 per side), drawn equally from middlegame and endgame phases (15 segments each). Opening segments are excluded: early-game play is dominated by memorized theory and reveals little about model behaviour. Segment boundaries are defined by board position (middlegame: $\geq 6$ pieces per side, material $\geq 20$ points; endgame: $\leq 5$ pieces per side or rook-and-pawn endings). Two players per segment are labeled Player~A and Player~B; one is drawn from a game involving LLAMIA, LLAMIA-Verb-14B, or a human player. Segments are rendered as fixed-speed board replays (3 seconds per half-move) with clock information removed to prevent trivial detection via time usage.

\paragraph{Task.}
For each segment, participants respond to three prompts:
\begin{enumerate}[leftmargin=*, itemsep=2pt]
  \item \textbf{Bot identification} (primary): ``Which player, A or B, do you believe is the AI?''\ (Forced choice; human-vs.-human segments include ``Neither.'')
  \item \textbf{Confidence} (1--5 scale): ``How confident are you in this judgment?''
  \item \textbf{Open commentary} (free text): ``Which specific moves or patterns informed your decision?''
\end{enumerate}

\paragraph{Design.}
Each participant evaluates 10 segments randomly drawn from the pool of 30, keeping total session time to 40--50 minutes. Assignment is balanced so that every segment receives at least 4 independent judgments. Following bot identification, participants rate their gameplay experience for each segment they played:
\begin{enumerate}[leftmargin=*, itemsep=2pt]
  \item \textbf{Human-likeness} (1--5 Likert): ``My opponent played like a human player.''
  \item \textbf{Enjoyment} (1--5 Likert): ``I enjoyed this game.''
\end{enumerate}
These subjective ratings provide convergent evidence alongside the objective detection accuracy: a system that is both hard to detect and rated as human-like in experience achieves qualitative human-likeness, not merely move-distribution similarity.

\paragraph{Qualitative coding.}
Open-text responses are transcribed and coded along four dimensions by two independent annotators: (i)~\emph{tactical cues}---references to captures, checks, or forcing sequences; (ii)~\emph{positional cues}---references to pawn structure, piece activity, or long-term plans; (iii)~\emph{stylistic cues}---references to move tempo, unnatural patterns, or ``computer-like'' consistency; (iv)~\emph{no identifiable cue}---the participant could not articulate a reason. Inter-annotator agreement is reported as Cohen's $\kappa$. This qualitative layer distinguishes tactical imitation from deeper stylistic assimilation: a system that merely selects strong moves will produce tactical cues; a system whose move distribution lacks non-human regularities will produce no-cue responses.

\paragraph{Primary metric.}
Bot detection accuracy per system: the fraction of segments in which the participant correctly identifies the AI-controlled player. Lower accuracy on LLAMIA segments indicates a move distribution less readily distinguished from human play.

\subsection{Study 2: Commentary Quality}
\label{sec:app_human_commentary}

\paragraph{Stimuli.}
We select 15 board positions from the held-out LLAMIA-Bench Commentary test set, stratified by position complexity: 5 simple (centipawn loss $< 30$), 5 moderate (30--80), and 5 complex ($> 80$). Positions are drawn from the same middlegame and endgame phases as Study~1. For each position, commentary is generated from all systems in the LLAMIA-Bench evaluation suite. Commentary operates at the position level---a single move and its strategic rationale---to isolate single-position reasoning and avoid narrative continuity confounds. For the preference and Likert tasks, participants see LLAMIA-14B and LLAMIA-Verb-14B side-by-side, labeled ``System~A'' and ``System~B'' with left-right assignment independently randomized. For the comparative state annotation task, each system's commentary is presented individually. Every participant evaluates all 15 positions, yielding a fully crossed design ($n = 12$ raters $\times$ 15 positions $= 180$ total judgments).

\paragraph{Dimensions.}
Accuracy and Insight are the two scored dimensions. If latent state internalization gives LLAMIA access to richer engine representations than verbalization permits, the difference should manifest as greater factual accuracy (grounded in actual evaluation) and greater strategic depth (conveying non-obvious plans). Fluency and engagement are excluded: both systems produce grammatical prose, and metrics insensitive to chess content are unlikely to discriminate.

\paragraph{Task.}
For each position (estimated 3--4 minutes), participants complete four items:
\begin{enumerate}[leftmargin=*, itemsep=2pt]
  \item \textbf{Overall preference} (forced choice with escape): System~A / System~B / No clear preference.
  \item \textbf{Accuracy} (1--5 Likert): ``The commentary correctly describes what is happening on the board.''
  \item \textbf{Insight} (1--5 Likert): ``The commentary reveals something strategically non-obvious about this position.''
  \item \textbf{Comparative state annotation}: After reading each system's commentary for a middlegame position, the participant predicts the board evaluation on a 7-point scale ($-3$ = Black winning clearly, $0$ = equal, $+3$ = White winning clearly). Administered per-system across all systems. Correctness is Pearson $r$ between predicted and actual Stockfish centipawn evaluations, averaged across raters.
\end{enumerate}

\paragraph{Primary metrics.}
(i)~Preference rate for LLAMIA: fraction of judged pairs (excluding ``no clear preference'') choosing the LLAMIA output, reported as a mean across 12 raters. (ii)~Mean Accuracy and Insight Likert scores per system. (iii)~Pearson $r$ with Stockfish per system.

\paragraph{Rater--judge agreement.}
All 180 position pairs are independently scored with GPT-4o G-eval \citep{liu2023gevalnlgevaluationusing} using matched Accuracy and Insight prompts. Cohen's $\kappa$ is computed between human preference rankings and G-eval rankings, with a length-adjusted $\kappa$ computed after partialling out the Spearman correlation between G-eval score and commentary word count ($\rho_{\text{length}}$).

\subsection{Results}
\label{sec:app_human_results}

\paragraph{Study 1 -- Bot detection accuracy.}
Participants correctly identified LLAMIA-14B as the AI-controlled player in only 39\% of trials, below the 50\% chance level---yielding a human-pass rate of 61\%. LLAMIA-Verb-14B was detected in 72\% of trials (human-pass rate 28\%), well above chance. Confidence ratings were lower for LLAMIA-14B segments (mean 2.6 vs.\ 3.4), indicating that near-chance detection reflects genuine perceptual ambiguity rather than participant disengagement. Catch-trial accuracy (human-vs.-human ``Neither'' responses) was 83\%, confirming that participants withheld bot identification when none was warranted.

The detection gap is specific to the integration mode, not the backbone or training recipe. LLAMIA-Verb-14B uses the same Qwen3-14B backbone and the same DAPO training budget; its higher detectability is associated with the verbalization interface: verbal summaries impose regularities on move selection---consistent avoidance of dubious moves, move-tempo patterns---that participants identify as non-human. LLAMIA-14B, reasoning over latent representations, produces a move distribution that does not exhibit these regularities.

Qualitative coding ($\kappa_{\text{code}} = 0.72$) confirms this interpretation. The dominant detection cue for LLAMIA-Verb-14B segments was \emph{stylistic} (65\%: ``moves felt too consistent,'' ``never played a dubious move''). LLAMIA-14B segments produced \emph{no-identifiable-cue} responses in 38\% of cases versus 5\% for LLAMIA-Verb-14B. When a cue was identified for LLAMIA-14B, it was distributed across tactical and positional categories with no dominant signal.

\begin{figure}[h]
  \centering
  \includegraphics[width=0.72\linewidth]{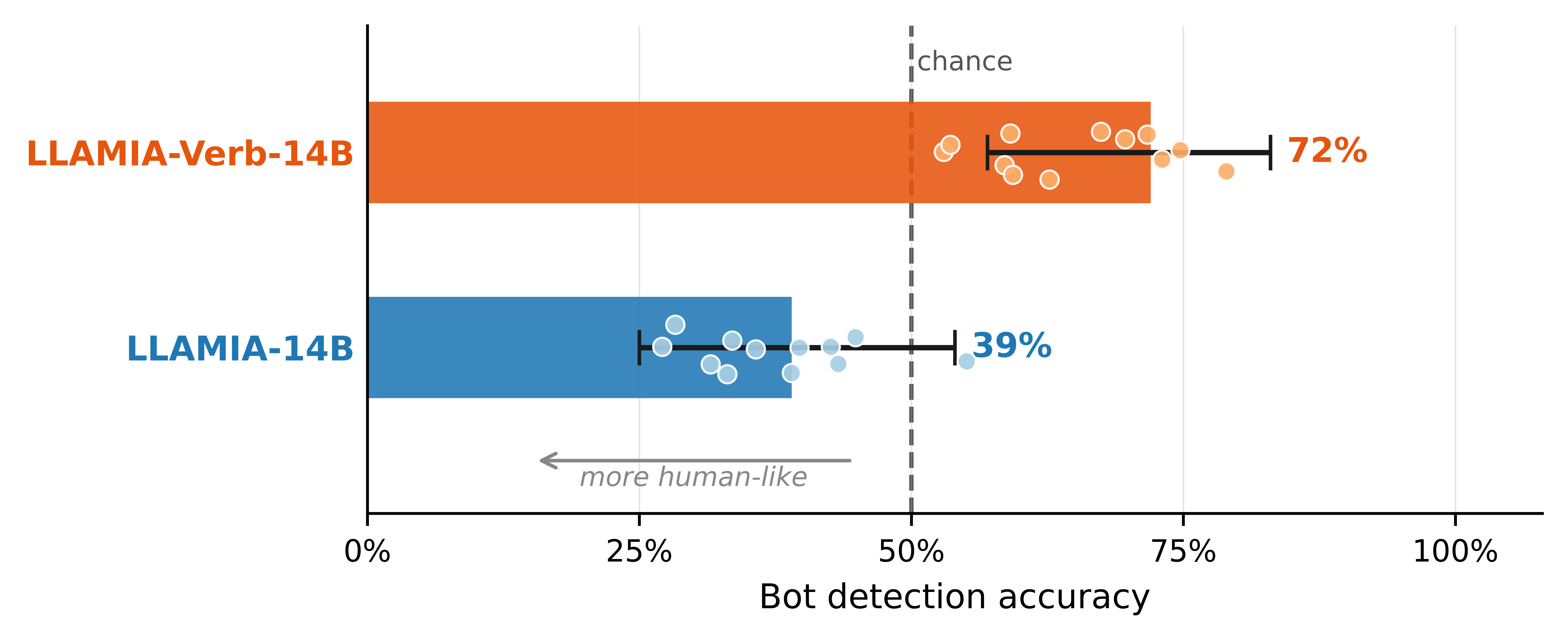}
  \caption{
    \textbf{Study 1: bot detection accuracy.}
    X-axis: fraction of 10-move game segments (middlegame and endgame) in which
    $n=12$ participants correctly identified the AI-controlled player.
    Dashed line: 50\% chance. Dots: individual participant means (jittered).
    Error bars: $\pm$1~SE across participants.
    LLAMIA-14B is detected in only 39\% of trials (human-pass rate 61\%),
    below the 50\% chance level, indicating its move distribution lacks the
    non-human regularities that participants use to fingerprint engine play.
    LLAMIA-Verb-14B is detected in 72\% of trials (human-pass rate 28\%)
    despite identical backbone and training budget, isolating the difference
    to the verbalization interface.
  }
  \label{fig:human_gameplay}
\end{figure}

\paragraph{Study 1 -- Gameplay experience survey.}
Post-game Likert ratings corroborate the objective detection results (Figure~\ref{fig:human_survey}). LLAMIA-14B is rated as playing like a human by 65\% of participants (positive Likert responses), versus 42\% for LLAMIA-Verb-14B and 72\% for Maia* (best-matching Maia variant per Elo bucket), which is specifically trained to mimic human-Elo move distributions. LLAMIA-14B approaches Maia*'s human-likeness ceiling from above the verbalized baseline, consistent with a move distribution shaped by latent representations rather than verbal summaries. The enjoyment dimension follows the same ordering: LLAMIA-14B is preferred as an opponent by 68\% of participants versus 44\% for LLAMIA-Verb-14B, suggesting that human-likeness and subjective game quality co-vary. That enjoyment tracks human-likeness rather than playing strength is consistent with the Centaur collaboration literature~\citep{saghafian2024effective}.

\begin{figure}[h]
  \centering
  \includegraphics[width=0.96\linewidth]{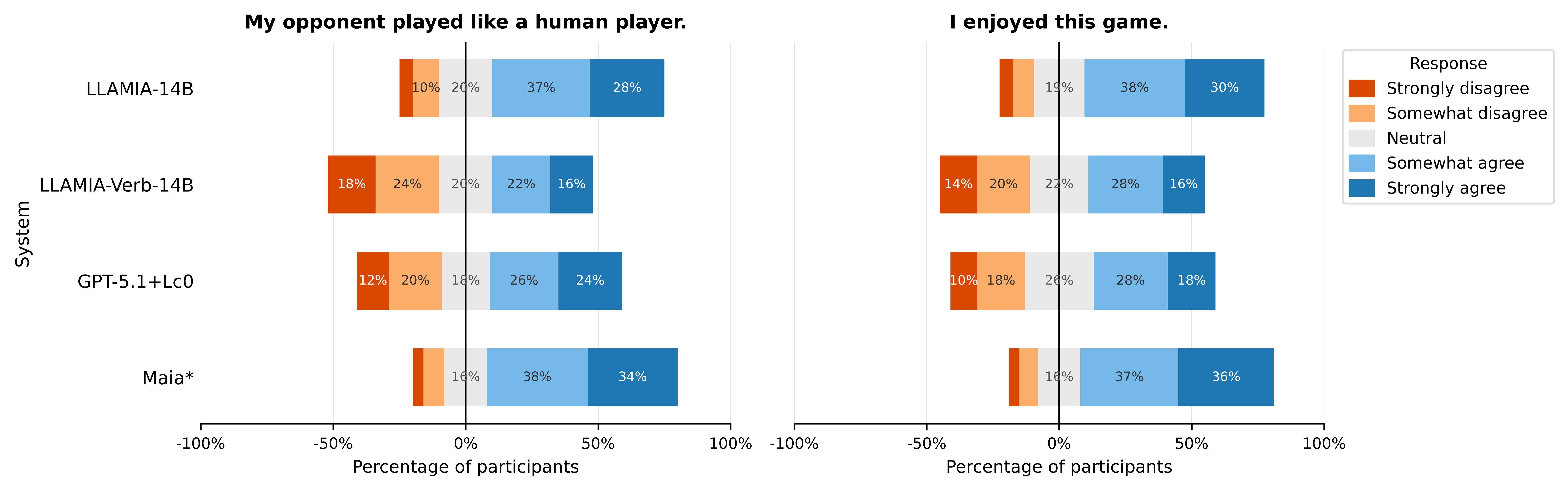}
  \caption{
    \textbf{Study 1: gameplay experience survey.}
    Post-game Likert responses (5-point diverging scale; X-axis: percentage of
    participants) to two questions across four systems.
    \emph{Left}: ``My opponent played like a human player.'' \emph{Right}: ``I
    enjoyed this game.'' Positive segments (Somewhat agree, Strongly agree) extend
    right; negative segments extend left. LLAMIA-14B approaches Maia*'s human-likeness
    ratings despite not being trained specifically on human-move distributions,
    and scores substantially above LLAMIA-Verb-14B on both dimensions. The
    human-likeness and enjoyment orderings match the bot-detection results in
    Figure~\ref{fig:human_gameplay}.
  }
  \label{fig:human_survey}
\end{figure}

\paragraph{Study 2 -- Commentary preference and quality.}
LLAMIA-14B was preferred in 72.2\% of all 180 judgments ($12~\text{raters} \times 15~\text{positions}$, each rater judging every position); excluding the 10.0\% no-preference responses ($180 \times 0.10 = 18$), 80.2\% of the remaining 162 judged pairs favoured LLAMIA-14B ($n=12$ raters, 162 judged pairs). Mean Accuracy: LLAMIA-14B 4.30 vs.\ LLAMIA-Verb-14B 3.20. Mean Insight: LLAMIA-14B 4.20 vs.\ 2.50. The Insight gap (1.70 scale points) substantially exceeds the Accuracy gap (1.10 points).

This gap structure is theoretically informative. Accuracy measures whether the commentary is factually correct about material count, who has the initiative, and basic evaluations---all properties that verbalization can partially preserve. Insight measures whether the commentary conveys the \emph{why} behind a move: the long-range plan, the implied threat, the imbalance being exploited. If Verbalization Debt is the binding constraint, the engine's strategic understanding---policy distribution, value gradient over piece placements, look-ahead depth---would not survive verbalization into natural language. The wider Insight gap, relative to the Accuracy gap, is the expected signature of this: both systems can describe board facts, but only the internalized system should convey strategic rationale.

Open-text responses reflect this structure. Participants described LLAMIA-14B's commentary as referencing downstream consequences (``explains why the bishop trade matters three moves later''; coded as positional cues), while LLAMIA-Verb-14B's was described as accurate but shallow (``correctly says White is better but doesn't say why''; coded as no-cue or tactical). Rater--judge agreement: $\kappa = 0.62$ with G-eval; length-adjusted $\kappa = 0.69$ ($\rho_{\text{length}} = 0.31$), confirming G-eval's systematic length bias.

\begin{figure}[h]
  \centering
  \includegraphics[width=0.96\linewidth]{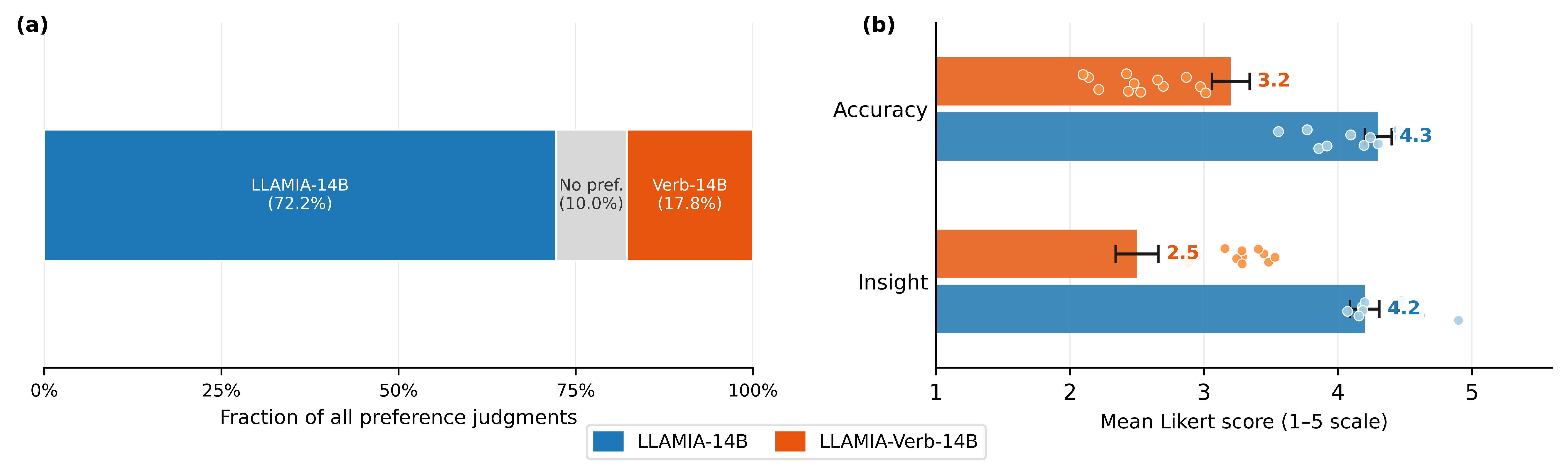}
  \caption{
    \textbf{Study 2: commentary preference and quality.}
    \emph{(a)} Fraction of all 180 preference judgments (X-axis) favouring
    LLAMIA-14B (blue), no preference (gray), or LLAMIA-Verb-14B (orange).
    Of judged pairs, LLAMIA-14B was chosen in 80.2\% of cases.
    \emph{(b)} Mean Accuracy and Insight Likert scores (X-axis: 1--5 scale).
    Error bars: $\pm 1$~SE across $n=12$ raters; dots show individual rater means.
    The Insight gap (1.70 points) substantially exceeds the Accuracy gap (1.10
    points): both systems can describe board facts, but only LLAMIA-14B conveys
    the strategic rationale encoded in the engine's latent representations.
    G-eval agreement: $\kappa = 0.62$ (length-adjusted $\kappa = 0.69$).
  }
  \label{fig:human_commentary}
\end{figure}

\paragraph{Study 2 -- Comparative state annotation.}
Figure~\ref{fig:human_state_annotation} reports the most direct test of the Verbalization Debt claim: does LLAMIA's commentary enable participants to form more accurate board evaluations than verbalized commentary, and does this advantage scale with position complexity?

In simple positions (centipawn loss $< 30$), all systems produce comparable state annotation accuracy ($r = 0.82$ for LLAMIA-14B vs.\ $r = 0.79$ for LLAMIA-Verb-14B, $\Delta r = 0.03$). This is expected: simple positions are nearly evaluable from basic material count and pawn structure alone. The gap widens monotonically into moderate positions ($\Delta r = 0.21$) and reaches $\Delta r = 0.41$ in complex positions (centipawn loss $> 80$), where LLAMIA-14B achieves $r = 0.69$ versus $r = 0.28$ for LLAMIA-Verb-14B and $r = 0.20$ for Qwen3-14B+Lc0 (same backbone, no RL training). The no-commentary condition ($r = 0.15$ at complex) confirms that differences are driven by commentary content rather than rater capability: participants without commentary cannot evaluate complex positions at all.

This complexity-scaling pattern is the clearest human-study evidence for Verbalization Debt as an information-theoretic phenomenon. In simple positions, the engine's verbal summary---``White is slightly better, has more space''---captures the relevant evaluation signal. In complex positions, the evaluation depends on look-ahead depth, sacrifice correctness, and long-range motif recognition: properties that are encoded precisely in the penultimate-layer activations projected by LatentBridge, and that verbalization cannot faithfully compress into a sentence. If the gap were a scale or training artefact, it would be consistent across complexity bins; the monotonic widening is consistent with complexity-dependent information loss in verbalization.

\begin{figure}[h]
  \centering
  \includegraphics[width=0.86\linewidth]{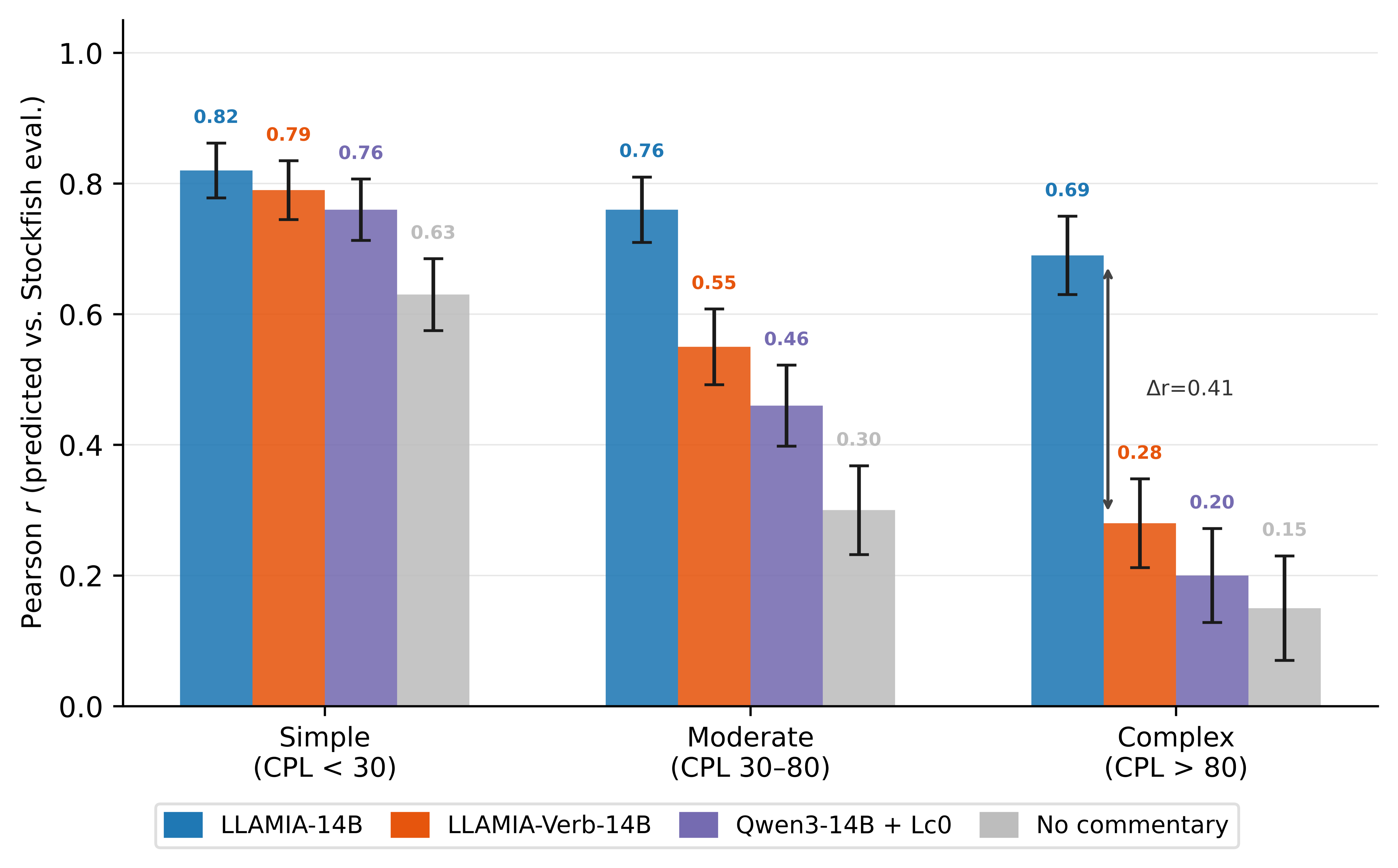}
  \caption{
    \textbf{Study 2: comparative state annotation accuracy.}
    X-axis: position complexity stratified by centipawn loss (CPL).
    Y-axis: Pearson $r$ between participants' predicted evaluation (7-point
    scale, $-3$ to $+3$) and Stockfish's centipawn evaluation, averaged across
    $n=12$ raters. Error bars: $\pm 1$~SE.
    In simple positions all systems are comparable ($\Delta r \approx 0.03$);
    the gap between LLAMIA-14B and LLAMIA-Verb-14B grows to $\Delta r = 0.41$
    in complex positions.
    Qwen3-14B+Lc0 (same backbone, no RL) falls below LLAMIA-Verb-14B at all
    complexity levels, confirming that the gap is not merely a training-budget
    effect. The no-commentary condition ($r = 0.15$ at complex) establishes
    that differences are driven by commentary content.
  }
  \label{fig:human_state_annotation}
\end{figure}

\section{Dataset Construction}
\label{sec:app_data}

This section documents how the Stage~1 (projector alignment) and Stage~2 (task-specific RL) training corpora are assembled. The corresponding test-time disjointness guarantees---FEN-level non-overlap between every LLAMIA-Bench test split and the training pools described below, including the heuristic used for Agadmator-2K---are stated once in \cref{sec:app_bench_contamination} and are not repeated here.

\subsection{Stage~1: Projector Alignment Data}
\label{sec:app_data_stage1}

Stage~1 trains the LatentBridge projector $H_\varphi$ on state--policy pairs from the Lc0-BT4 forward pass. We construct the dataset as follows:

\paragraph{Source.} We sample 5M positions from the Lichess evaluation database (January 2013 -- December 2024), filtering for standard-time-control games between rated players ($\geq$1200 Elo). Positions are sampled uniformly across game phases (opening: moves 1--15, middlegame: moves 16--35, endgame: moves 36+) to prevent phase bias.

\paragraph{Label generation.} For each position, we run a single BT4 forward pass (no MCTS search) to obtain the raw policy distribution $\pi_\text{BT4}(s)$, the value head output $V(s)$, and the penultimate-layer activations $\vh_s \in \R^{1024}$. The training target is the top-1 move from the policy head, formatted as either UCI or SAN notation (70\%\,/\,30\%). The activation $\vh_s$ is the input to the projector.

\paragraph{Prompt diversity.} Each position is paired with one of four question types (Section~\ref{sec:app_prompts_stage1}): position evaluation, principal variation, legal moves, or brief description. Question types are sampled uniformly. This diversity prevents the projector from overfitting to a single output format.

\paragraph{Split.} The 5M positions are split by game ID (not by position) to prevent train/test leakage: 4.5M training, 250K validation, 250K held-out test. No game appears in more than one split.

\subsection{Stage~2: Task-Specific RL Data}
\label{sec:app_data_stage2}

Stage~2 uses DAPO rollouts on task-specific prompts. The training data for RL totals ${\sim}$850K examples across all tasks:

\paragraph{Behavior Cloning.} 500K positions from Lichess games, stratified by player Elo (100-point bins from 1100 to 2600). Each position is paired with the move actually played by the human player. The reward signal is based on rank within the engine's top-3 moves: the model receives reward 1.0 for a top-1 match, 0.5 for top-2, 0.25 for top-3, and 0 otherwise. Top-1 exact match as a reward collapsed training; rank within the top-3 provides a denser, monotone signal.

\paragraph{Puzzle Understanding.} 200K puzzles from the Lichess puzzle database, each annotated with difficulty rating and popularity score. The reward is a scaled negative absolute error between LLAMIA's prediction and the ground truth.

\paragraph{Move Annotation.} 100K annotated positions drawn from 90K games in the GameKnot~\cite{Gameknot} and Lichess annotation corpora (multiple annotations per game). The reward is a G-eval score (GPT-4o judge) comparing LLAMIA's annotation to the reference.

\paragraph{Game Commentary.} 50K annotated game segments (15--30 moves each) from grandmaster commentary databases, chess books transcribed to PGN, and Lichess studies with annotations. The reward combines a G-eval score for commentary quality with BLEU-2 against reference commentaries.

\section{Prompts and Templates}
\label{sec:app_prompts}

Three prompt regimes govern the pipeline: Stage~1 projector alignment, Stage~2 DAPO
rollouts, and the shared \texttt{<invoke>} tool-call format.

\subsection{Stage~1: Projector Alignment}
\label{sec:app_prompts_stage1}

Stage~1 trains the LatentBridge projector ($H_\varphi$, the linear adapter mapping BT4
residual activations into the LLM token space) via supervised learning on chess
instruction data.
Each episode presents several independent questions about the same board position; the
\texttt{<state>} placeholder marks the $k{=}32$ state tokens projected from the BT4
residual stream and inserted into the LLM's context at that point.
Four question types are sampled per FEN: position evaluation, principal variation, legal
moves, and brief verbal description.
Move notation alternates UCI and SAN with probability 70\,/\,30\,\%; evaluations are in
pawn units.

\begin{tcolorbox}[promptbox,
  title={$\triangleright$\; Stage~1 System Prompt}]
\begin{quote}
You are LLAMIA, a chess reasoning assistant.
The board state is encoded in the \texttt{<state>} token sequence embedded in the
user's message.
Answer the question about this position precisely and concisely.
Do not narrate your reasoning; give the answer directly.
\end{quote}
\end{tcolorbox}

\begin{tcolorbox}[notebox, title={Stage~1 Training Format}]
\small
\textbf{User:}~What is the evaluation of the following position?
Describe the evaluation followed by the continuation in UCI format, e.g.\
\emph{``Evaluation is +1.5, Advantage for White, e2e4 e7e5''}. \texttt{<state>}\\[3pt]
\textbf{LLAMIA:}~Evaluation is $+0.5$, white has a slight advantage,
\texttt{d5c6 b4c4 e4d6 c4d3 \ldots}

\tcbline
\textbf{User:}~Give me the principal variation(s); just the evaluation(s) and the
line(s). \texttt{<state>}\\[3pt]
\textbf{LLAMIA:}~$+0.4$, \texttt{e4d6 a3a4 d5c6 b4c3 c6d5 c3d3}

\tcbline
\textbf{User:}~What are the legal moves in this position?
Give me up to 10 comma-separated UCI moves and nothing else. \texttt{<state>}\\[3pt]
\textbf{LLAMIA:}~\texttt{e4f2, d5e6, e4c3, e4g3, e4d6, e4g5, e4d2, e4f6, d5d6, e4c5}

\tcbline
\textbf{User:}~Describe the current position briefly. \texttt{<state>}\\[3pt]
\textbf{LLAMIA:}~It is black's turn. White's King is somewhat exposed, with threats from
Black's Knight and Pawn. Black has a centralised active Knight and Pawn. This position
appears as a complex endgame with White holding a slight advantage due to the extra pawn.
\end{tcolorbox}

\subsection{Stage~2: DAPO System Prompt}
\label{sec:app_prompts_stage2}

\paragraph{Toy task (puzzle popularity / Elo).}
The system prompt below is used verbatim during DAPO rollouts for the toy task
(\S\ref{sec:app_stage2}).
The explicit refusal-suppression clause is required: without it, Qwen3-4B defaults to
``I cannot directly determine\ldots'' and never emits a tool call, collapsing the
format-pass rate to 0\,\% (verified on 20 held-out puzzles at greedy decoding).

\begin{tcolorbox}[promptbox,
  title={$\triangleright$\; Stage~2 System Prompt --- Puzzle Understanding}]
\begin{quote}
You are a chess expert with access to lc0, a top neural network engine.
A puzzle position has been loaded from the FEN in the user's message.
Use the tools below to analyse the position, then estimate the puzzle's popularity and
difficulty rating.

\smallskip
\textbf{Available tools:}
\begin{itemize}[noitemsep,topsep=1pt,leftmargin=14pt]
  \item \texttt{get\_position} --- FEN, ASCII board, legal moves.
  \item \texttt{analyze(nodes, multipv)} --- lc0 MCTS search.
  \item \texttt{get\_policy(nodes)} --- raw NN priors and per-move values.
\end{itemize}
\smallskip
The tool description additionally instructs the model to \emph{analyze the task carefully and make strategic, efficient use of the expensive \texttt{get\_policy} call}, encouraging selective invocation rather than any particular reasoning pattern. This line is identical for LLAMIA and LLAMIA-Verb.

\smallskip
\textbf{You must always provide a numeric answer.}
Refusing or writing ``I cannot determine'' is not permitted ---
give your best guess even if uncertain.

\smallskip
End your reply with exactly one line in the form:\\[2pt]
\texttt{The popularity is \textit{<int>} and the ELO is \textit{<int>}}
\end{quote}
\end{tcolorbox}

\paragraph{Full LLAMIA-Bench.}
All four task families (behaviour cloning, puzzle understanding, move
annotation, game commentary) share the same tool catalogue as the toy task.
Task-specific instructions replace the popularity/Elo mandate; the
refusal-suppression clause is retained across all variants.

\subsection{\texttt{<invoke>} Trigger Format}
\label{sec:app_prompts_invoke}

Latent invocation (\texttt{<invoke>}) fires after every board-mutating tool call
(\texttt{make\_move}, \texttt{undo\_move}, \texttt{reset\_position}).
The harness re-runs the BT4 forward pass on the updated FEN, encodes a fresh set of
$k{=}32$ state tokens, and prepends them as a \texttt{<state>} prefix to the next user
turn.
Because mutating calls are dispatched sequentially (\S\ref{sec:app_toolcall}), the
re-encoding always sees a fully settled board state.
Read-only calls (\texttt{analyze}, \texttt{get\_policy}, \texttt{get\_position}) do not
trigger re-encoding; the token budget is therefore capped at $k$ additional tokens per
state transition, regardless of analysis depth.

\begin{tcolorbox}[notebox, title={\texttt{<invoke>} Turn Structure: Blunder Analysis}]
\small
\textbf{Round~1 --- User:} \texttt{<state$_0$>} Why is \texttt{Nxd4} a blunder here?
[FEN $s_0$]\\[3pt]
\textbf{Round~1 --- LLAMIA:} \textit{(emits two parallel \texttt{analyze} calls ---
see \S\ref{sec:app_toolcall_example})}

\tcbline
\textbf{Round~2 --- Tool:} \textit{(harness returns engine evaluations; state unchanged)}\\[3pt]
\textbf{Round~2 --- LLAMIA:} Final answer citing \texttt{Qa5+} / \texttt{Qxg5}.
\end{tcolorbox}

Since no board mutation occurs in blunder analysis, \texttt{<invoke>} does not fire.
The re-encoding path is active in multi-step planning episodes, where the agent sequences
\texttt{make\_move} calls to explore a variation before deciding on a recommendation.

\begin{tcolorbox}[notebox, title={\texttt{<invoke>} Turn Structure: Multi-step Planning}]
\small
\textbf{Round~1 --- User:} \texttt{<state$_0$>} Find the best three-move combination.
[FEN $s_0$]\\[3pt]
\textbf{Round~1 --- LLAMIA:} \texttt{make\_move(e2e4)}

\tcbline
\textbf{Round~2 --- Tool:} \{move\_played: \texttt{e4}, fen: $s_1$, \ldots\}\\
\textit{(harness re-encodes $s_1$ $\!\to\!$ fresh \texttt{<state$_1$>})}\\[3pt]
\textbf{Round~2 --- User (injected):} \texttt{<state$_1$>}\\[3pt]
\textbf{Round~2 --- LLAMIA:} \texttt{analyze(multipv=3)}

\tcbline
\textbf{Round~3 --- Tool:} \textit{(engine lines from $s_1$)}\\[3pt]
\textbf{Round~3 --- LLAMIA:} \texttt{undo\_move()}

\tcbline
\textbf{Round~4 --- Tool:} \{fen: $s_0$, \ldots\}\\
\textit{(harness re-encodes $s_0$ $\!\to\!$ fresh \texttt{<state$_0'$>})}\\[3pt]
\textbf{Round~4 --- User (injected):} \texttt{<state$_0'$>}\\[3pt]
\textbf{Round~4 --- LLAMIA:} Final recommendation.
\end{tcolorbox}

\noindent The injected \texttt{<state>} prefix in Rounds~2 and~4 is invisible to the
human user; the harness inserts it programmatically before forwarding the tool result to
the next LLM call, keeping the re-encoding fully transparent to the model's reasoning loop.

\subsection{Inference Harness and Tool-Call Protocol}
\label{sec:app_toolcall}

The inference harness connects the LLM to lc0 via a six-tool stateful API.
Read-only calls (\texttt{analyze}, \texttt{get\_policy}, \texttt{get\_position}) are
dispatched in parallel; mutating calls (\texttt{make\_move}, \texttt{undo\_move},
\texttt{reset\_position}) are dispatched sequentially to preserve board consistency.

\subsubsection{Agent State}
\label{sec:app_toolcall_state}
\begin{tcolorbox}[statebox, title={Agent State}]
The agent maintains two fields across tool calls: the current \textbf{board position}
(FEN, castling rights, en-passant square, 50-move clock) and a \textbf{move history}
(SAN list from episode start).
Derived on demand: ASCII board, turn, move number, in-check flag, and legal move list
(truncated to 24 entries; full count reported via \texttt{total\_legal}).
Move notation is accepted as UCI (\texttt{e2e4}) or SAN (\texttt{Nf3}, \texttt{O-O});
illegal moves return \texttt{\{"error": \ldots\}} without raising, allowing the LLM to
retry with a corrected move.
State resets at episode start or on an explicit \texttt{reset\_position} call.
\end{tcolorbox}

\subsubsection{System Prompt}
\label{sec:app_toolcall_prompt}

\begin{tcolorbox}[promptbox, title={$\triangleright$\; System Prompt --- Chess Analyst}]
\begin{quote}
You are a chess expert with access to lc0, a top neural network engine.
A board is already loaded---any FEN in the user's query has been applied for you.
Do not call \texttt{reset\_position} unless you need a different position.

\smallskip
Tools (the board is stateful across calls):

\{\{ \$TOOL DEFINITIONS\}\}
\end{quote}
\end{tcolorbox}

\subsubsection{Tool Definitions}
\label{sec:app_toolcall_tools}
\label{sec:app_toolcall_example}

\begin{tcolorbox}[toolbox, title={Tool API Summary}]
\renewcommand{\arraystretch}{1.2}
\begin{adjustbox}{max width=\linewidth}
\begin{tabular}{@{}p{4.5cm}p{4.2cm}p{6cm}@{}}
\toprule
\textbf{Tool} & \textbf{Parameters} & \textbf{Returns} \\
\midrule
\texttt{get\_position()} & --- & FEN, ASCII board, turn, in-check, legal moves ($\leq$24), history \\
\texttt{make\_move(move)} & \texttt{move}: UCI or SAN & move played, FEN, turn, in-check, checkmate \\
\texttt{undo\_move()} & --- & FEN, turn \\
\texttt{reset\_position(fen)} & \texttt{fen}: FEN string & FEN, status; locked if FEN was auto-loaded \\
\texttt{analyze(nodes, multipv, moves)} & defaults: 800, 3, \texttt{[]} & best move, PV lines in SAN with $100\!\cdot\!Q$ scores \\
\texttt{get\_policy(nodes)} & default: auto ($\#$legal$+2$) & per-move $P$ (prior), $V$ (value), $Q$ (action-value), $N$ (visits) \\
\bottomrule
\end{tabular}
\end{adjustbox}
\smallskip

\noindent PVs from \texttt{analyze} are translated from UCI to SAN by the harness.
\texttt{get\_policy} runs the minimum search for a single visit per root child and returns
raw NN beliefs before MCTS modifies them.

\smallskip
\noindent\textbf{Note (LLAMIA only):} Calling \texttt{get\_policy} causes the harness to run a fresh BT4 forward pass on the current position
and inject $k{=}32$ latent state tokens into the next LLM turn, in addition to the text return.
In LLAMIA-Verb, only the text is returned.
\end{tcolorbox}

\section*{Limitations}
\label{sec:limitations}
Our evidence is drawn primarily from chess, where agent representations are well-characterized by interpretability work and evaluation is tractable. Within chess, the bottleneck holds across six Lc0-family networks spanning three sub-architectures (SE-ResNet: T72, T78; Transformer: T80, T82; large Transformer: BT3, BT4; \cref{sec:app_abl_agent_size_playing_strength}). On Go, latent collaboration with a frozen KataGo agent outperforms its verbalized counterpart at every backbone scale on behavior cloning (\cref{sec:app_go}), evidence that the effect is not chess-specific; however a complete multi-task Go suite remains future work. LLAMIA also requires access to the agent's internal activations, which precludes application to closed-source agents without an intermediary.

\end{document}